\documentclass[sigconf, screen]{acmart}

\usepackage[utf8]{inputenc} % allow utf-8 input
\usepackage[T1]{fontenc}    % use 8-bit T1 fonts
\usepackage{hyperref}       % hyperlinks
\usepackage{url}            % simple URL typesetting
\usepackage{booktabs}       % professional-quality tables
\usepackage{amsfonts}       % blackboard math symbols
\usepackage{nicefrac}       % compact symbols for 1/2, etc.
\usepackage{microtype}      % microtypography
\usepackage{graphicx}
\usepackage{graphics}
\usepackage{color,soul}
\usepackage{comment}
\usepackage{cite}
\usepackage{algorithmic}
\usepackage{textcomp}
\usepackage{xcolor}
\usepackage{hyperref}
\usepackage{wrapfig}
\usepackage{booktabs}
\usepackage{epstopdf} 
\usepackage{multirow}
\usepackage{subfigure}
\usepackage{xspace}
\usepackage{url}
\usepackage{adjustbox}
\usepackage{appendix}

\usepackage{dsfont}
\usepackage{relsize}

\let\origNabla\nabla
\renewcommand*\nabla{\mathlarger\origNabla}

\DeclareMathOperator{\E}{\mathbb{E}}

\AtBeginDocument{%
  \providecommand\BibTeX{{%
    \normalfont B\kern-0.5em{\scshape i\kern-0.25em b}\kern-0.8em\TeX}}}

\setcopyright{acmcopyright}
\copyrightyear{2021}
\acmYear{2021}
\setcopyright{acmcopyright}\acmConference[KDD '21]{Proceedings of the 27th ACM SIGKDD Conference on Knowledge Discovery and Data Mining}{August 14--18, 2021}{Virtual Event, Singapore}
\acmBooktitle{Proceedings of the 27th ACM SIGKDD Conference on Knowledge Discovery and Data Mining (KDD '21), August 14--18, 2021, Virtual Event, Singapore}
\acmPrice{15.00}
\acmDOI{10.1145/3447548.3467449}
\acmISBN{978-1-4503-8332-5/21/08}

\begin{document}
\fancyhead{}

\title{PID-GAN: A GAN Framework based on a Physics-informed Discriminator for Uncertainty Quantification with Physics}

% \affiliation{%

% \affiliation{%

% \affiliation{%

\author{Arka Daw}
\authornote{Both authors contributed equally to this research.}
\affiliation{%
 \institution{Virginia Tech \\ Dept. of Computer Science}
%  \city{Blacksburg}
%  \state{VA}
 \country{}
 }
\email{darka@vt.edu}

\author{M. Maruf}
\authornotemark[1]
\affiliation{%
 \institution{Virginia Tech \\ Dept. of Computer Science}
%  \city{Blacksburg}
%  \state{VA}
 \country{}
 }
\email{marufm@vt.edu}
 
\author{Anuj Karpatne}
\affiliation{%
 \institution{Virginia Tech \\ Dept. of Computer Science}
%  \city{Blacksburg}
%  \state{VA}
 \country{}
 }
\email{karpatne@vt.edu}

% \affiliation{%

% \affiliation{%

\renewcommand{\shortauthors}{Daw, et al.}

\begin{abstract}
  As applications of deep learning (DL) continue to seep into critical scientific use-cases, the importance of performing uncertainty quantification (UQ) with DL has become more pressing than ever before. In scientific applications, it is also important to inform the learning of DL models with knowledge of physics of the problem to produce physically consistent and generalized solutions. This is referred to as the emerging field of physics-informed deep learning (PIDL). We consider the problem of developing PIDL formulations that can also perform UQ. To this end, we propose a novel physics-informed GAN architecture, termed PID-GAN, where the knowledge of physics is used to inform the learning of both the generator and discriminator models, making ample use of unlabeled data instances. We show that our proposed PID-GAN framework does not suffer from imbalance of generator gradients from multiple loss terms as compared to state-of-the-art. We also empirically demonstrate the efficacy of our proposed framework on a variety of case studies involving benchmark physics-based PDEs as well as imperfect physics. All the code and datasets used in this study have been made available on this link \footnote{\url{https://github.com/arkadaw9/PID-GAN}}.
\end{abstract}







\maketitle

\section{Introduction}
% \begin{itemize}
%     \item addressing novelty : This extension although simple, has a lot of positive impact over the baselines. (implication of our method)
%     \item PI-UQ motivating example in Introduction
% \end{itemize}

As applications of deep learning (DL) continue to seep into critical scientific and engineering use-cases such as climate science, medical imaging, and autonomous vehicles, the importance of performing uncertainty quantification (UQ) with deep learning has become more pressing than ever before. UQ in DL is especially important to ensure trust or confidence in deep learning predictions by end-users of DL frameworks such as scientists and real-world practitioners.

% . While there is no way we can guarantee an ideal model, more information regarding the certainty of a models' predictions could prevent catastrophic failures. As a result, models developed using uncertainty quantification (UQ) are far more reliable and trustworthy. 

Another aspect of DL formulations that is highly relevant while solving real-world problems in scientific and engineering domains is their ability to incorporate knowledge of the governing physics  of the problem in the design and training of DL models. Indeed, there is a rapidly growing body of work in the emerging field of \emph{physics-informed deep learning} \citep{karpatne2017theory,willard2020integrating}, where the primary goal is to use physics as another form of supervision for learning \textit{generalizable} DL solutions, even when the number of training labels is small (a problem  encountered in many real-world settings). This is commonly achieved by adding \textit{physics-based loss} functions in the training objective of DL models, for capturing the consistency of DL predictions with physics-equations on unlabeled instances (which are plentifully available in many applications).

\begin{figure}[t]
    \centering
    \includegraphics[scale=0.42]{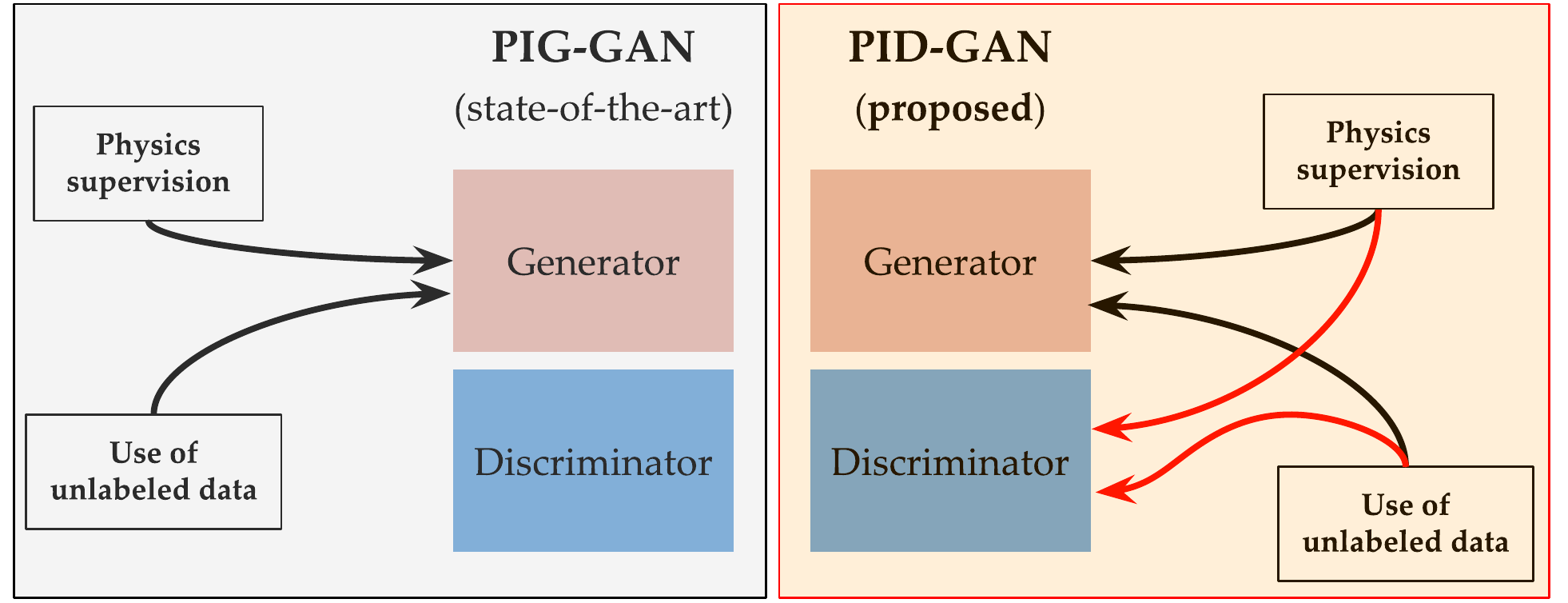}
    \caption{Visualizing differences between our proposed PID-GAN innovations and the state-of-the-art, PIG-GAN \citep{yang2019adversarial}}
    \label{fig:intro_fig}
\vspace{-4ex}
\end{figure}

In this paper, we consider the problem of performing UQ while making use of both physics and data supervision for learning DL models. This is relevant in any scientific application where it is important to generate distributions of output predictions instead of point estimates, while ensuring the learning of physically consistent and generalizable DL solutions. For example, in the domain of climate science, it is important to generate distributions of projected climate variables such as temperature in the future, while ensuring that our predictions comply with the laws of physics such as conservation of mass and energy. A state-of-the-art formulation for this category of problems involves a recent work by Yang et al. \citep{yang2019adversarial}, where the training of conditional Generative Adversarial Network (cGAN) models were informed using physics-based loss functions. We refer to this formulation as Physics-Informed Generator (PIG)-GAN, to reflect the fact that the physics-supervision was explicitly used to inform the generator (but not the discriminator). 

Since GAN-based frameworks are naturally amenable to generating output distributions by varying the built-in input noise vector, PIG-GAN serves as a pivotal study in incorporating physics knowledge in GAN frameworks for performing UQ. However, PIG-GAN fails to exploit the full potential of the adversarial optimization procedure inherent to GAN-based frameworks for minimizing complex physics-based loss functions (e.g., those encountered in real-world problems involving physics-based PDEs). This is because the discriminator of PIG-GAN is still uninformed by physics and hence, does not leverage the physics-supervision on unlabeled instances. Further, the generator of PIG-GAN suffers from the imbalance of gradient dynamics of different loss terms, as demonstrated theoretically and empirically later in this work.

% , which are able to automatically learn an arbitrary loss function required to model the true data distribution. 
% Further, only the generator leverages the rich information available in the vast unlabeled data into the training dynamics, while the discriminator fails to exploits this. Additionally, Wang et. al.\citep{wang2020understanding} identifies that a common mode of failure that arises in physics-informed loss formulations, where imposing the physics as a soft constraint can lead to an imbalance in the gradient flow dynamics of neural networks, which would also translate to GANs. 

% Another more natural way of obtaining uncertainty estimates is using generative adversarial networks(GANs)\citep{goodfellow2014generative}, where a distribution on the outputs can be obtained by varying the built-in noise vector. 

% The ideology of \emph{physics-informed deep learning} has already been integrated into the GAN framework \citep{yang2019adversarial, zhang2019quantifying} by simply imposing the physics supervision on the outputs of the generator. We refer to this as Physics-informed generator-GAN (PIG-GAN) (See Figure \ref{fig:intro_fig}). 

To address the challenges in state-of-the-art for physics-informed UQ, we propose a novel GAN architecture, termed as \textbf{Physics Informed Discriminator (PID)-GAN}, where physics-supervision is directly injected into the adversarial optimization framework to inform the learning of both the generator and discriminator models with physics. 
% Our formulation exploits the adversarial framework to minimize the predictive errors as well as the physical violations of our predictions. 
PID-GAN allows the use of unlabeled data instances for training both the generator and discriminator models, whereas PIG-GAN uses unlabeled data only for  training the generator but not the discriminator.
% Additionally, PID-GAN allows the use of unlabeled data instances for training both the generator and discriminator models, thus addressing the paucity of labeled data for training physics-informed GAN frameworks. 
Figure \ref{fig:intro_fig} illustrates the major differences between our proposed PID-GAN framework and the state-of-the-art framework, PIG-GAN. These differences impart several advantages to our proposed PID-GAN framework. We theoretically show that PID-GAN does not suffer from the imbalance of Generator gradients, in contrast to PIG-GAN. We also present an extension of PID-GAN that can work in situations even when the physics-supervision is imperfect (either due to incomplete knowledge or observational noise). On a variety of case studies (three problems involving physics-based PDEs and two problems involving imperfect physics), we empirically show the efficacy of our framework in terms of prediction accuracy, physical consistency, and validity of uncertainty estimates, compared to baselines.

The remainder of the paper is organized as follows. Section 2 presents background and related work on physics-informed UQ. Section 3 describes the proposed framework. Section 4 describes experimental results while Section 5 provides concluding remarks.

\vspace{-2ex}
\section{Background and Related Work}

\subsection{Uncertainty Quantification with DL}
% The ability of a model to perform uncertainty quantification(UQ) is often considered as the first step in the direction towards robust and interpretable machine learning. 
% Predictions without uncertainty estimates are usually not trust-worthy and could be misleading at times. 
Uncertainty quantification (UQ) is an important end-goal in several real-world scientific applications where it is vital to produce distributions of output predictions as opposed to point estimates, allowing for meaningful analyses of the confidence in our predictions. In the context of deep learning, a number of techniques have been developed for UQ, including the use of Bayesian approximations
 \citep{wang2018adversarial, jospin2020hands, wang2016towards} and ensemble-based methods \citep{hu2019mbpep, mcdermott2019deep, pearce2020uncertainty, zhang2018learning}. 
% Traditionally, Gaussian Mixture Models (GMMs) and deep Gaussian process \citep{damianou2013deep,borovykh2018gaussian, salimbeni2017doubly} have been widely used for uncertainty quantification. These methods assume that the underlying distribution of the outputs are Gaussian and estimates the uncertainty by encoding the similarity between samples using a kernel function. 
% Among other Bayesian methods listed above, a noteworthy UQ technique is Bayes by Backprop \citep{blundell2015weight}, which learns a probability distribution by assuming Gaussian priors on the weights of the neural networks. One of the greatest challenges in UQ is to tractably compute the true posterior distribution. 
A simple approach for performing UQ given a trained DL model is to apply Monte Carlo (MC)-Dropout on the DL weights during testing \citep{gal2016dropout}. While MC-Dropout is easy to implement, they are quite sensitive to the choice of dropout rate, which can be difficult to tune \citep{daw2020physics}.

Another line of work for performing UQ in DL is to use generative models like variational autoencoders (VAEs) \citep{kingma2013auto, bohm2019uncertainty} and cGANs \citep{yang2019adversarial, zhang2019quantifying}. In a cGAN setting, the generator $G_{\boldsymbol{\theta}}$ learns the mapping from input vector $\mathbf{x}$ and some random noise vector $\mathbf{z}$ to $\mathbf{y}$, $G : (\mathbf{x},\mathbf{z}) \xrightarrow[]{} \mathbf{y}$. The generator is trained such that its predictions $\mathbf{\hat{y}} =  G(\mathbf{x},\mathbf{z})$ cannot be distinguished from ``real'' outputs by an adversarially trained discriminator $D_{\boldsymbol{\phi}}$. The discriminator $D$, on the other hand, is trained to detect the generator's ``fake'' predictions. The built-in noise vector $z$ in cGANs can be varied to obtain a distribution on the output predictions $\mathbf{\hat{y}}$. The learning objective of such a cGAN can be written as the following mini-max game:
\vspace{-1ex}
\begin{equation}
    \min_{G_{\boldsymbol{\theta}}} \max_{D_{\boldsymbol{\phi}}} \E_{x,z}\Big[\log D(G(x,z),x)\Big] + \E_{x,y}\Big[\log\big(1-D(y,x)\big)\Big]
\end{equation}
% \begin{equation}
%     \mathcal{L}_{G}(\boldsymbol{\theta}) = \E_{\mathbf{x}, \mathbf{z}}\Big[D(\mathbf{x}, \mathbf{\hat{y}})\Big]
% \end{equation}
% \begin{equation}
%     \mathcal{L}_{D}(\boldsymbol{\phi}) = -\E_{\mathbf{x},\mathbf{z}}\Big[\log \big(D(\mathbf{x},\mathbf{\hat{y}})\big)\Big] -\E_{\mathbf{x},\mathbf{y}}\Big[\log \big(1 -  D(\mathbf{x},\mathbf{y})\big)\Big]
% \end{equation}
where $\boldsymbol{\theta}$ and $\boldsymbol{\phi}$ are the parameters of $G$ and $D$, respectively. In practice, it is common to optimize $\E_{x,z}\big[D(G(x,z),x)\big]$ instead of $\E_{x,z}\big[\log D(G(x,z),x)\big]$ while training the generator \citep{gulrajani2017improved}.

\vspace{-1ex}
\subsection{Physics-Informed Deep Learning (PIDL)}
% Traditionally, generative adversarial networks(GANs) have been introduced as a methods of approximating the intractable data distribution specially in high dimensional spaces. Some of the more prominent applications include image synthesis, image-to-image translation, (give more examples), etc. [citations]. 

% Deep Neural Networks show astounding results across a diverse range of predictive tasks using highly over-parameterized black-box models. Although such models yield remarkable performance in data-rich domains, their power is hugely impacted under sparse-data conditions. Moreover, these black-box models fail to exploit any prior knowledge available in the form of constraints involving symmetries, conservation laws, or other domain knowledge. To address this problem, one needs to embed the domain knowledge into the black-box models. 

There is a growing volume of work on informing deep learning methods with supervision available in the form of physics knowledge, referred to as the field of physics-informed deep learning (PIDL) \citep{karpatne2017theory,willard2020integrating}. 
One of the primary objectives of PIDL is to learn deep learning solutions that are consistent with known physics and generalize better on novel testing scenarios, especially when the supervision contained in the labeled data is small.  A promising direction of research in PIDL is to explicitly add \textit{physics-based loss functions} in the the deep learning objective, that captures the consistency of neural network predictions on unlabeled instances w.r.t. known physics (e.g., physics-based PDEs \citep{raissi2019physics}, monotonic constraints 
\citep{karpatne2017physics,jia2019physics}, and kinematic equations \citep{stewart2017label}). 

Formally, given a labeled set $\{(\mathbf{x_{u_i}}, \mathbf{y_{u_i}})\}^{N_{u}}_{i = 1}$ where $(\mathbf{x}, \mathbf{y})$ denotes an input-output pair, we are interested in learning a neural network model, $\mathbf{{\hat{y}}} = f_{\boldsymbol{\theta}}(\mathbf{x})$, such that along with reducing prediction errors of $\mathbf{{\hat{y}}}$ on the labeled set, we also want to ensure that $\mathbf{{\hat{y}}}$ satisfies $K$ physics-based equations, $\{\mathcal{R}^{(k)}(\mathbf{x},\mathbf{y}) = 0\}_{k=1}^K$ on a larger set of unlabeled instances $\{\mathbf{x_{f_j}}\}^{N_{f}}_{j = 1}$ where $N_f >> N_u$. This can be achieved by adding a physics-based loss in the learning objective that captures the residuals of $\mathbf{{\hat{y}}}$ w.r.t each physics-equation $\mathcal{R}^{(k)}$ as:

\vspace{-3ex}
\begin{align}
    \mathcal{L}(\boldsymbol{\theta}) = \frac{1}{N_{u}} \sum_{i=1}^{N_{u}} \big[ \mathbf{y_{u_{i}}} - \mathbf{{\hat{y}}_{u_{i}}} \big]^{2} + \frac{\lambda}{N_f} \sum_{j=1}^{N_f} \sum_{k=1}^{K} \mathcal{R}^{(k)}(\mathbf{x_{f_{j}}}, \mathbf{{\hat{y}}_{f_{j}}})^2, \label{eq:pinn}
\end{align}
where $\lambda$ is a trade-off hyper-parameter for balancing physics-based loss with prediction errors. Notice that the physics-based loss only depends on model predictions and not ground-truth values, and hence enables the use of unlabeled data to provide an additional form of physics-supervision, along with the supervision contained in the labeled data. A seminal work in optimizing Eq.~\ref{eq:pinn} is the framework of \emph{physics-informed neural networks} (PINN) \citep{raissi2019physics}, that was developed for the specific use-case where the physics-equations $\mathcal{R}^{(k)}$ are PDEs. In this paper, we generically refer to all formulations that optimize Eq.~\ref{eq:pinn} as PINN, regardless of the form of $\mathcal{R}^{(k)}$.

While PINN provides a powerful strategy for incorporating physics-supervision in the learning of neural networks, a fundamental challenge in PINN is to balance the gradient flow dynamics of physics loss and prediction errors at different stages (or epochs) of training with a constant $\lambda$. To address this challenge, a variant of PINN, referred to as adaptive-PINN (APINN) \citep{wang2020understanding}, has recently been developed to adaptively tune $\lambda$ at different stages of training for balancing the gradients of different loss terms. 
\vspace{-2ex}
\subsection{Physics-Informed UQ}

The body of work in PIDL that can perform UQ appears closest to the focus of this paper. Specifically, we are interested in generating uncertainty estimates while ensuring that our predictions lie on a solution manifold that is consistent with known physics. One simple way to achieve this is to implement MC-Dropout for the PINN framework and its variants, so to produce distributions of output predictions. However, as demonstrated in a recent study \citep{daw2020physics}, a major limitation with this approach is that the minor perturbations introduced by MC-Dropout during testing can easily throw off a neural network to become physically inconsistent, even if it was informed using physics during training.
We refer to the MC-Dropout versions of PINN and APINN as PINN-Drop and APINN-Drop, respectively, which are used as baselines in our work.

Another line of research in PIDL for UQ involves incorporating physics-based loss functions in the learning objective of cGAN models, which are inherently capable of generating distributions of output predictions.
% To incorporate prior physical knowledge into the GAN framework, one can impose it as a soft constraint on the generator loss. This would motivate the generator to produce physically consistent outputs while trying to match the true data distribution using the adversarial loss. In this paper, we refer to this framework as Physics-Informed Generator - GAN (PIG-GAN).
In particular, Yang et al. \citep{yang2019adversarial} recently developed a physics-informed GAN formulation with the following objective functions of generator ($G$) and discriminator ($D$), respectively:
\vspace{-2ex}
\begin{align}
\label{eq:PIG_gloss}
    \mathcal{L}_{G}(\boldsymbol{\theta}) &= \frac{1}{N_u}\sum_{i=1}^{N_u} D(\mathbf{x_{u_i}}, \mathbf{\hat{y}_{u_i}})
    \: + \frac{\lambda}{N_f} \sum_{j=1}^{N_f} \sum_{k=1}^{K}  \big[\mathcal{R}^{(k)}(\mathbf{x_{f_j}}, \mathbf{\hat{y}_{f_j}})^2\big], \\
    \label{eq:PIG_dloss}
     \mathcal{L}_{D}(\boldsymbol{\phi}) &= - \frac{1}{N_u}\sum_{i=1}^{N_u} \log \big(D(\mathbf{x_{u_i}}, \mathbf{\hat{y}_{u_i}})\big) - \frac{1}{N_u}\sum_{i=1}^{N_u} \log \big(1 - D(\mathbf{x_{u_i}}, \mathbf{y_{u_i}})\big),
     \vspace{-3ex}
\end{align}
% \vspace{-1ex}
where $\mathbf{\hat{y}_{u_i}} = G(\mathbf{x_{u_i}}, \mathbf{z_{u_i}})$ and $\mathbf{\hat{y}_{f_j}} = G(\mathbf{x_{f_j}}, \mathbf{z_{f_j}})$ are the generator predictions on labeled and unlabeled points, respectively. Notice that in this formulation, the physics-supervision (in terms of $\mathcal{R}^{(k)}$) only appears in the generator objective, while the discriminator only focuses on distinguishing between ``real'' and ``fake'' samples on the labeled set (similar to a cGAN discriminator). Hence, we refer to this formulation as Physics-Informed Generator (PIG)-GAN to  reflect the fact that only the generator is physics-informed.

While PIG-GAN provides a valid approach for physics-informed UQ, it is easy to observe that it suffers from several deficiencies. First, it makes ineffective use of physics-supervision to only inform the generator, while the discriminator is kept unchanged. As a result, it is unable to use the full power of adversarial optimization procedures inherent to GAN frameworks for jointly minimizing physics-based loss functions along with prediction errors. Instead, it under-utilizes the discriminator to only focus on prediction errors on the labeled set. Second, by design, the discriminator of PIG-GAN is only trained on the small set of labeled instances, thus missing out on the vastly large number of unlabeled instances available in many real-world applications. Third, the use of an explicit trade-off parameter $\lambda$ to balance physics loss and prediction errors results in a similar problem of gradient flow imbalance as faced by PINN. As we empirically demonstrate later in Section \ref{sec:pdes}, this leads to inferior generalization performance compared to our proposed approach.

% could create an imbalance in the backpropagated gradients of the generator, ultimately hindering the gradient flow dynamics during optimization. This is a similar to the common mode of failure described in [] for PINNs. 
 
% Labelled vs unlabelled set (R can be computed on unlabelled set)
\section{Proposed Framework of PID-GAN}
\label{sec:methods}
\begin{figure}[t]
    \centering
    \includegraphics[scale=0.21]{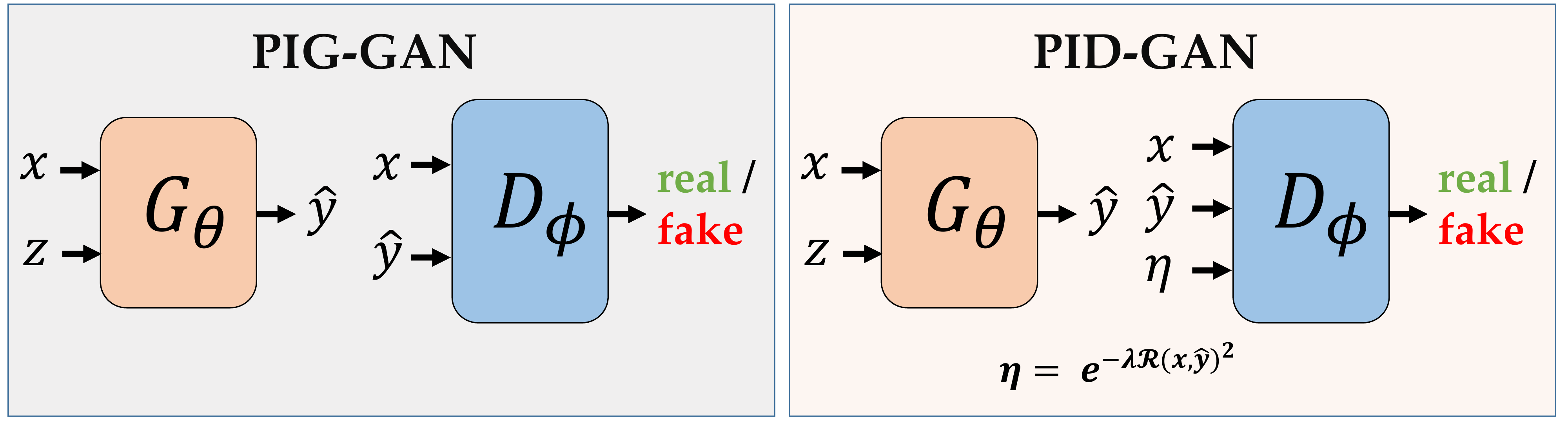}
    % \vspace{-1ex}
    \caption{Architecture of PIG-GAN and PID-GAN}
    \label{fig:method_fig}
\vspace{-3ex}
\end{figure}

We propose a novel Physics-Informed Discriminator (PID)-GAN formulation that embeds physics-supervision directly into the adversarial learning framework, as shown in Figure \ref{fig:method_fig}. Specifically, we utilize the physics residuals to compute a physics consistency score ($\eta$) for each prediction, indicating the likelihood of the prediction being physically consistent. These physics consistency scores are fed into the discriminator as additional inputs, such that the discriminator not only distinguishes between real and fake samples by learning from the underlying distribution of labeled points but also using the additional physics-supervision. 
% In  the following, we describe our approach for computing physics consistency scores and training objective. 
% This facilitates utilization of the adversarial framework to simultaneously optimize both the predictive errors and the physics loss. 
% As a result, both the generator and the discriminator of PID-GAN can be trained on physics-supervision from the unlabeled instances, resulting in better generalization capabilities. 
% We further demonstrate that our design leads to a balanced back-propagated generator gradients for the labeled and unlabeled components its constituent loss terms.

\par \noindent \textbf{Estimating Physics Consistency Scores:}
Formally, we compute the physics consistency score of a prediction $\mathbf{\hat{y}}$ w.r.t. the $k^{\text{th}}$ physical constraint using the following equation:
\begin{equation}
    \eta_{k} = e^{-\lambda \mathcal{R}^{(k)}(\mathbf{x}, \mathbf{\hat{y}})}
\end{equation}
 Larger the value of $\eta_{k}$, it is more likely the prediction $\mathbf{\hat{y}}$ obeys the $k^{\text{th}}$ physical constraint, i.e., $\mathcal{R}^{(k)}$ is smaller. The vector notation of would lead to the following:  $\boldsymbol{\eta}$, such that $\boldsymbol{\eta} = [\eta_{1}, \eta_{2}, \cdots, \eta_{K}]$. 
% \vspace{-1ex} 

\par \noindent \textbf{Training Objective:}
Our physics-informed discriminator 
% is a binary classifier that learns the decision boundary between real and fake samples from $\mathbf{x}$, $\mathbf{y}$, and $\boldsymbol{\eta}$. In other words, we 
learns the mapping $D : (\mathbf{x},\mathbf{y},\boldsymbol{\eta}) \xrightarrow[]{} \Omega \in [0, 1]$, where $\Omega$ represents the probability of a sample being ``fake''. 
% A discriminator score close to 0 means that the sample is more likely to be ``real'' and vice versa.
The objective function of the generator and the discriminator of PID-GAN can then be denoted as:
\begin{equation}
\vspace{-2ex}
\label{eq:PID_gloss}
    \mathcal{L}_{G}(\boldsymbol{\theta}) = \frac{1}{N_u} \sum_{i=1}^{N_u} D(\mathbf{x_{u_i}, \hat{y}_{u_i}, \boldsymbol{\eta}_{u_i}}) + \frac{1}{N_f} \sum_{j=1}^{N_f} D(\mathbf{x_{f_j}, \hat{y}_{f_j}, \boldsymbol{\eta}_{f_j}})
\end{equation}

\vspace{-2ex}
\begin{equation}
\label{eq:PID_dloss}
\begin{split}
    \mathcal{L}_{D}(\boldsymbol{\phi}) & =  -\frac{1}{N_u} \sum_{i=1}^{N_u} \log \big(D(\mathbf{x_{u_i}, \hat{y}_{u_i}, \boldsymbol{\eta}_{u_i}})\big) \\
     & -\frac{1}{N_u} \sum_{i=1}^{N_u} \log \big(1 - D(\mathbf{x_{u_i}, y_{u_i}, 1})\big) \\
    & - \frac{1}{N_f} \sum_{j=1}^{N_f} \log \big(D(\mathbf{x_{f_j}, \hat{y}_{f_j}, \boldsymbol{\eta}_{f_j}})\big) \\
    & -\frac{1}{N_f} \sum_{j=1}^{N_f} \log \big(1 - D(\mathbf{x_{f_j}, \hat{y}_{f_j}, 1})\big)  
\end{split}
\end{equation}
\vspace{-3ex}

% \begin{align}
%      \mathcal{L}_{D}(\boldsymbol{\phi}) = 
%   & -\E_{\mathbf{x_{u}}, \mathbf{z_{u}}}\big[\log D(\mathbf{x_{u}}, \mathbf{\hat{y}_{u}},\boldsymbol{\eta}_{\mathbf{u}})\big]  \label{eq:PID_dloss}\\ 
%      & - \E_{\mathbf{x_{u}},\mathbf{z_{u}}}\big[\log(1- D(\mathbf{x_{u}}, \mathbf{y_{u}}, \textbf{1}_{\mathbf{u}}))\big]  \nonumber\\ 
%      & -\E_{\mathbf{x_{f}}, \mathbf{z_{f}}}\big[\log D(\mathbf{x_{f}}, \mathbf{\hat{y}_{f}}, \boldsymbol{\eta}_{\mathbf{f}})\big]  \nonumber\\ 
%      & - \E_{\mathbf{x_{f}},\mathbf{z_{f}}}\big[\log(1- D(\mathbf{x_{f}}, \mathbf{\hat{y}_{f}}, \textbf{1}_{\mathbf{f}}))\big] \nonumber
% \end{align}

where, $\mathbf{\hat{y}_{u_i}} = G(\mathbf{x_{u_i}}, \mathbf{z_{u_i}})$, $\mathbf{\hat{y}_{f_j}} = G(\mathbf{x_{f_j}}, \mathbf{z_{f_j}})$.
To interpret generator loss function (Equation \ref{eq:PID_gloss}), let us inspect its two constituent terms. The first and the second term represents the score of the physics-informed discriminator $D$ for the predictions on labeled and unlabeled points, respectively. The generator attempts to minimize both of these scores in order to fool the discriminator into thinking that these generated ``fake'' predictions $\mathbf{\hat{y}_{u}}$ and $\mathbf{\hat{y}_{f}}$ are ``real''.

On the other hand, the loss function for the discriminator (Equation \ref{eq:PID_dloss}) is comprised of four terms. The first two terms denote the training objective of a binary classifier which distinguishes between the inputs $(\mathbf{x_{u_i}}, \mathbf{\hat{y}_{u_i}},\boldsymbol{\eta}_{\mathbf{u_i}})$ and $(\mathbf{x_{u_i}}, \mathbf{y_{u_i}}, \textbf{1})$. 
% In other words, it distinguishes between the predictions $\mathbf{\hat{y}_{u_i}}$ with a likelihood score of $\boldsymbol{\eta}_{\mathbf{u_i}}$ and the ground truth labels $\mathbf{y_{u_i}}$ with a likelihood score of \textbf{1} on the labeled set $\mathbf{x_{u}}$. 
Note that the consistency score is \textbf{1} only when $\mathcal{R}^{(k)}(\mathbf{x_{u_i}},\mathbf{{y}_{u_i}})=0$ for all $k \in \{1,2, .., K\}$, i.e., the ground truth labels obey the given physics-equations. 
Similarly, we can interpret the last two terms of equation \ref{eq:PID_dloss} as the training objective of a binary classifier which tries to distinguish between the inputs $(\mathbf{x_{f_j}}, \mathbf{\hat{y}_{f_j}},\boldsymbol{\eta}_{\mathbf{f_j}})$ and $(\mathbf{x_{f_j}}, \mathbf{\hat{y}_{f_j}}, \textbf{1})$. Since we don't have labels on the unlabeled set, we use $\mathbf{\hat{y}_{f_j}}$ as a proxy to the ground truth values. This encourages the generator to increase the physics-consistency score $\boldsymbol{\eta}_{\mathbf{f_j}}$ in order to fool the discriminator on the unlabeled set. 

% Thus, the physics-informed discriminator $D$ is equivalent to a binary classifier which is able to detect the ``fakes'' on both labeled and unlabeled set.

%   In section [fill], we demonstrate that without the 4th term in \ref{eq:PID_dloss}, the generator is not able to optimise the likelihood score $\boldsymbol{\eta}_{\mathbf{f}}$ on the unlabeled set, given that the labeled and the unlabeled sets come from two different distributions.
\vspace{-1ex}
\subsection{Analysis of Generator Gradients}
\subsubsection*{Gradient Analysis for PIG-GAN}
We assume that the both the inputs \{$\mathbf{x_u}$, $\mathbf{x_f}$\} and the outputs \{$\mathbf{\hat{y}_u}$, $\mathbf{\hat{y}_f}$\} follow same distribution. 
The back-propagated gradients of the generator can be computed using the chain-rule of differentiation as follows:
\begin{equation}
\begin{split}
    \nabla_{\theta} \mathcal{L}_{G} & = \sum_{i} \nabla_{\mathbf{\hat{y}_{u_i}}}\mathcal{L}_{G}.\nabla_{\theta}\:\mathbf{\hat{y}_{u_i}} + \sum_{j} \nabla_{\mathbf{\hat{y}_{f_j}}}\mathcal{L}_{G}.\nabla_{\theta}\:\mathbf{\hat{y}_{f_j}} \\ 
    & = \sum_{i} \mathbf{C_{u_i}}.\nabla_{\theta}\:\mathbf{\hat{y}_{u_i}} + \sum_{j} \mathbf{C_{f_j}}.\nabla_{\theta}\:\mathbf{\hat{y}_{f_j}}
\vspace{-2ex}
\end{split}
\end{equation}

% \textbf{Definition 1:}  
Let us define $\mathbf{C_{u_i}}$ as the contribution of the $i^{\text{th}}$ instance of the labeled set to the backpropagated generator gradients.

\begin{equation}
\label{eq:cu_pig}
    \mathbf{C_{u_i}} = \nabla_{\mathbf{\hat{y}_{u_i}}}\mathcal{L}_{G} = \frac{1}{N_u} \nabla_{\mathbf{\hat{y}_{u_i}}} D_{\mathbf{u_i}}
\end{equation}

where, $D_{\mathbf{u_i}} = D(\mathbf{x_{u_i}, \hat{y}_{u_i}})$. Similarly, we can define $\mathbf{C_{f_j}}$ as the contribution of the $j^{\text{th}}$ instance of the unlabeled set to the backpropagated generator gradients.
\vspace{-2ex}
\begin{equation}
\label{eq:cf_pig}
\begin{split}
    \mathbf{C_{f_j}} & = \nabla_{\mathbf{\hat{y}_{f_j}}}\mathcal{L}_{G} = \frac{\lambda}{N_f} \sum_{k=1}^{K} \nabla_{\mathbf{\hat{y}_{f_j}}} \big(\mathbf{\mathcal{R}^{(k)}_{f_j}}\big)^2 \\
    & = \frac{2\lambda}{N_f} \sum_{k=1}^{K} \mathbf{\mathcal{R}^{(k)}_{f_j}}\nabla_{\mathbf{\hat{y}_{f_j}}} \mathbf{\mathcal{R}^{(k)}_{f_j}}
\end{split}
\end{equation}
where $\mathbf{\mathcal{R}^{(k)}_{f_j} = \mathcal{R}^{(k)}\big(x_{f_j}, \hat{y}_{f_j}\big)}$.
Since we assume that $\mathbf{\hat{y}_{u_i}}$ and $\mathbf{\hat{y}_{f_j}}$ follow similar distributions, $\nabla_{\theta}\mathbf{\hat{y}_{u_i}}$ and $\nabla_{\theta}\mathbf{\hat{y}_{f_j}}$ would also follow similar distributions. Hence, the overall gradient dynamics would be mostly controlled by the contribution terms $\mathbf{C_{u_i}}$ and $\mathbf{C_{f_j}}$.

\noindent \textbf{Remarks:} It is easy to see that $\mathbf{C_{u_i}}$ and $\mathbf{C_{f_j}}$ have widely different functional forms in PIG-GAN. $\mathbf{C_{u_i}}$ depends on  $\nabla_{\mathbf{\hat{y}_{u_i}}} D_{\mathbf{u_i}}$, i.e., it changes as the weights of the discriminator are updated during training. On the other hand, $\mathbf{C_{f_j}}$ depends on $\nabla_{\mathbf{\hat{y}_{f_j}}} \mathbf{\mathcal{R}_{f_j}^{(k)}}$, which is specific to the set of physics-equations, and does not change as the discriminator is updated. 
% This suggests that gradients of the backpropagated components on the unlabeled set would rely on the type of physical constraint being enforced. 
While we can try to balance $\mathbf{C_{u_i}}$ and $\mathbf{C_{f_j}}$ with the help of $\lambda$, choosing a constant $\lambda$ is difficult as  $\mathbf{\mathcal{R}_{f_j}^{(k)}}$ changes across epochs and across $j$. 

\subsubsection*{Gradient Analysis for PID-GAN}
The physics consistency score on the labeled and unlabeled sets can be computed as $\mathbf{\boldsymbol{\eta}_{u_i}^{(k)}} = e^{-\lambda\boldsymbol{\mathcal{R}}_{\mathbf{u_i}}^{\mathbf{(k)}}}$ and $\mathbf{\boldsymbol{\eta}_{f_j}^{(k)}} = e^{-\lambda\boldsymbol{\mathcal{R}}_{\mathbf{f_j}}^{\mathbf{(k)}}}$, respectively, resulting in the following values of $\mathbf{C_{u_i}}$:
% \vspace{-10ex}
% It must be noted that $\boldsymbol{\mathcal{R}}_{\mathbf{u}}^{(i)}$ is a vector of size $K$. Hence, $\mathbf{\boldsymbol{\eta}_u^{(i)}}$ would also be a vector of the same size.
\vspace{-2ex}
\begin{equation}
\vspace{-4ex}
\label{eq:cu_pid}
    \begin{split}
        \mathbf{C_{u_i}} & = \frac{1}{N_u}\nabla_{\mathbf{\hat{y}_{u_i}}}\mathbf{D_{u_i}} + \frac{1}{N_u} \Big(\nabla_{\boldsymbol{\eta}_{\mathbf{u_i}}}\mathbf{D_{u_i}}\Big)\Big(\nabla_{\mathbf{\hat{y}_{u_i}}}\mathbf{\boldsymbol{\eta}_{u_i}}\Big) \\ 
        &= \frac{1}{N_u}\nabla_{\mathbf{\hat{y}_{u_i}}}\mathbf{D_{u_i}} - \frac{2\lambda}{N_u} \sum_{k=1}^{K}  \big(\nabla_{\mathbf{\boldsymbol{\eta}_{u_i}}}\mathbf{D_{u_i}}\big)^{(\mathbf{k})}\mathbf{\boldsymbol{\eta}_{u_i}}^{(\mathbf{k})} \mathbf{\mathcal{R}_{u_i}^{(k)}} \nabla_{\mathbf{\hat{y}_{u_i}}}\mathcal{R}_{\mathbf{u_i}}^{\mathbf{(k)}},
    \end{split}
\end{equation}
where $\big(\nabla_{\mathbf{\boldsymbol{\eta}_{u_i}}}\mathbf{D_{u_i}}\big)^{(\mathbf{k})}$ and $\mathbf{\boldsymbol{\eta}_{u_i}}^{(\mathbf{k})}$ denotes the $k^{\text{th}}$ term of the gradient $\nabla_{\mathbf{\boldsymbol{\eta}_{u_i}}}\mathbf{D_{u_i}}$ and  the vector $\mathbf{\boldsymbol{\eta}_{u_i}}$, respectively.
Similarly,
\begin{equation}
\label{eq:cf_pid}
    \mathbf{C_{f_j}} = \frac{1}{N_u}\nabla_{\mathbf{\hat{y}_{f_j}}}\mathbf{D_{f_j}} - \frac{2\lambda}{N_u} \sum_{k=1}^{K}  \big(\nabla_{\mathbf{\boldsymbol{\eta}_{f_j}}}\mathbf{D_{f_j}}\big)^{(\mathbf{k})}\mathbf{\boldsymbol{\eta}_{f_j}}^{(\mathbf{k})} \mathbf{\mathcal{R}_{f_j}^{(k)}} \nabla_{\mathbf{\hat{y}_{f_j}}}\mathcal{R}_{\mathbf{f_j}}^{\mathbf{(k)}}
\end{equation}

\noindent \textbf{Remarks:} Observe that the functional forms of the two contribution terms $\mathbf{C_{u_i}}$ and $\mathbf{C_{f_j}}$ of PID-GANs have similar functional forms. This is not surprising since the generator's objective function is symmetric by formulation, i.e., $\mathbf{x_u}$ and $\mathbf{x_f}$ are interchangeable. If we use the same assumptions on \{$\mathbf{x_u}$, $\mathbf{x_f}$\} and the outputs \{$\mathbf{\hat{y}_u}$, $\mathbf{\hat{y}_f}$\} as made in the case of PIG-GAN, we would see that $\boldsymbol{\eta}_{\mathbf{u_i}}$ and $\boldsymbol{\eta}_{\mathbf{f_j}}$ would have the same distribution. Similarly, it can be shown that $\mathcal{R}_{\mathbf{u_i}}^{(\mathbf{k})}$ and $\mathcal{R}_{\mathbf{f_j}}^{(\mathbf{k})}$ have the same distributions, and $\nabla_{\mathbf{\hat{y}_{u_i}}}\mathcal{R}_{\mathbf{u_i}}^{(\mathbf{k})}$ and $\nabla_{\mathbf{\hat{y}_{f_j}}}\mathcal{R}_{\mathbf{f_j}}^{(\mathbf{k})}$ have the same distributions. Hence, each of the individual components of $\mathbf{C_{u_i}}$ and $\mathbf{C_{f_j}}$ would follow the same distribution, and we can thus expect that the magnitudes of the gradients for the labeled and unlabeled components to be similar. 

We can also notice the similarities in Equations \ref{eq:cf_pig}, \ref{eq:cu_pid}, and \ref{eq:cf_pid} to remark that the contribution terms of PID-GAN automatically learn adaptive weights for the physics-based loss that change across training epochs. In particular, all of these contribution formulae have terms involving physics residuals. For the contribution $\mathbf{C_{u_i}}$ in PID-GAN, we can observe that the second term has  multiplicative factors, $\big(\nabla_{\mathbf{\boldsymbol{\eta}_{u_i}}}\mathbf{D_{u_i}}\big)^{\mathbf{(k)}}$ and $\mathbf{\boldsymbol{\eta}_{u_i}}^{\mathbf{(k)}}$. While $\mathbf{\boldsymbol{\eta}_{u_i}}^{\mathbf{(k)}}$ would change as the  $k^{\text{th}}$ residual varies during training, $\big(\nabla_{\mathbf{\boldsymbol{\eta}_{u_i}}}\mathbf{D_{u_i}}\big)^{\mathbf{(k)}}$ depends on the current state of the discriminator and is optimized after every discriminator update. These changing multiplicative factors applied to the physics residual gradients can be viewed as automatically learning adaptive weights for each individual physics residual, as opposed to the use of a constant trade-off parameter $\lambda$ in PIG-GAN.

\subsection{Extension for Imperfect Physics}
In some applications, the physics-equations available to us during training may be derived using simplistic assumptions of complex real-world phenomena. For example, while predicting the velocities of two particles just after collision, we usually assume that the energy and the momentum of the system would be conserved. However, such assumptions are only valid in ideal conditions and could easily be misleading when working with noisy real-world labels. To account for situations with imperfect physics, , i.e., when $\mathcal{R}^{(k)}(\mathbf{x,y}) \not= 0$ for atleast one $k \in \{1, 2, \cdots , K\}$, we provide the following extension of our proposed approach. 

The training objective of generator of the PID-GAN under imperfect physics conditions would remain the same (as shown in equation \ref{eq:PID_gloss}. However, the objective for the discriminator would be modified as follows: 
\vspace{-1ex}
\begin{equation}
\label{eq:PID_dloss_limited}
\begin{split}
    \mathcal{L}_{D}(\boldsymbol{\phi}) = & -\frac{1}{N_u} \sum_{i=1}^{N_u} \log \big(D(\mathbf{x_{u_i}, \hat{y}_{u_i}, \boldsymbol{\eta}_{u_i}})\big) \\
    & -\frac{1}{N_u} \sum_{i=1}^{N_u} \log \big(1 - D(\mathbf{x_{u_i}, y_{u_i}, \boldsymbol{\eta}'_{\mathbf{u_i}}})\big) \\ 
    & - \frac{1}{N_f} \sum_{j=1}^{N_f} \log \big(D(\mathbf{x_{f_j}, \hat{y}_{f_j}, \boldsymbol{\eta}_{f_j}})\big)
\end{split}
\end{equation}
\vspace{-1.5ex}
\begin{equation}
    \boldsymbol{\eta}'_{\mathbf{u_i}} = [e^{-\lambda \mathcal{R}^{(1)}(\mathbf{x_{u_i},y_{u_i}})^2}, e^{-\lambda \mathcal{R}^{(2)}(\mathbf{x_{u_i},y_{u_i}})^2}, \cdots , e^{-\lambda \mathcal{R}^{(K)}(\mathbf{x_{u_i},y_{u_i}})^2}]
\end{equation}

$\boldsymbol{\eta}'_{\mathbf{u}}$ denotes the physics consistency score of the ground truth satisfying the physical constraints on the labeled set. This formulation prevents the generator from predicting samples which blindly satisfies the imperfect physical constraints. Instead, it would learn to mimic the distribution of physics consistency scores of ground truth samples.
It must also be noted that the discriminator loss in Equation \ref{eq:PID_dloss_limited} does not contain the fourth term from the Equation \ref{eq:PID_dloss}, since we cannot compute the physics consistency score of the predictions on the unlabeled set. This introduces an assumption into the model, that the labeled set and the unlabeled set must come from the same distribution. Otherwise, we might end up learning a trivial discriminator $D$ that would look at the input distributions of $\mathbf{x_u}$ and $\mathbf{x_f}$ and classify the unlabeled set as fake.

\subsection{Mitigating Mode Collapse}
GANs are notoriously difficult to train, and it has been observed that sometimes the generator only learns to generate samples from few modes of the data distribution in spite of wide abundance of samples available from the missing modes. This phenomena is commonly known as \emph{mode collapse} \citep{salimans2016improved}. In order to address this problem, Li et. al. \citep{li2018learning} provides an information theoretic point of view of using an additional inference model $Q_{\boldsymbol{\zeta}}$ to learn the mapping $Q_{\boldsymbol{\zeta}}:\{ \mathbf{x},\mathbf{\hat{y}} \}\xrightarrow{} \mathbf{z}$. This inference model delivers stability to the training procedure by mitigating mode collapse while providing a variational approximation to the true intractable posterior over the latent variables $\mathbf{z}$. We utilize this additional network to improve all of our GAN-based formulations.

\section{Experimental Results}
\par \noindent \textbf{Case Studies:} We evaluate the performance of comparative models on two real-world datasets involving idealized (and imperfect) physics, and three benchmark datasets involving state-of-the-art physics-based PDEs. 

\par \noindent \textbf{Evaluation Metrics:}  We estimate the deviation of the ground truth values from the mean of our model predictions on every test point using the Root Mean Square Error (\textbf{RMSE}). For PDE problems, we use relative $L^2$-error instead of RMSE, following previous literature \citep{raissi2019physics,wang2020understanding}. The physical inconsistencies of our predictions are quantified using the absolute \textbf{residual} errors w.r.t. the physics-equations. Further, to assess the quality of our generated distributions, we utilize both the \textbf{standard deviation} of samples as well as the fraction of ground-truth values that fall within 95\% confidence intervals  of the sample distributions (referred to as \textbf{95\% C.I.} fractions). A lower standard deviation with higher 95\% C.I. is desired. 
% We perform 5 random runs of each model and report mean and standard deviations of evaluation metrics. 

\par \noindent \textbf{Baselines:} We compare our proposed PID-GAN with the following baselines: PIG-GAN, cGAN, PINN-Drop, and APINN-Drop. 

In this section, we briefly describe our results in each case study. Full details of experiments in each case study and additional results focusing on reproducibility are provided in Appendix \ref{sec-implementation}.

\begin{table}[t]
    \caption{Summary of model performances for collision speed estimation and tossing trajectory prediction in terms of the mean and standard deviation for 10 random runs. For collision speed estimation, the ground truth datapoints have mean residual of 93.96, while for tossing trajectory prediction, the ground truth datapoints have mean residual of 0.59.}
    \vspace{-2ex}
    \label{tab:non-ideal}
    \centering
    \begin{adjustbox}{width=1.1\columnwidth,center}
    \begin{tabular}{l|lllll}
    \toprule
    
    Models          & PINN-Drop                         & APINN-Drop             & cGAN               & PIG-GAN             & \textbf{PID-GAN}            \\ \hline
    \multicolumn{6}{l}{\textbf{Collision Speed Estimation}} \\ \hline
    
    RMSE            & $2.34 \pm 0.09$              & $2.35 \pm 0.06$   & $0.92 \pm 0.03$    & $1.65 \pm 0.39$     & $\mathbf{0.73 \pm 0.05}$  \\
    
    Residual        & $24.20 \pm 2.10$              & $22.80 \pm 2.60$   & $96.20 \pm 4.50$    & $45.50 \pm 23.00$     & $\mathbf{92.40 \pm 0.90}$  \\
    
    Std. Dev.        & $1.60 \pm 0.02$ & $1.60 \pm 0.02$ & $0.56 \pm 0.02$ & $0.03 \pm 0.01$  & $0.49 \pm 0.04$\\
    
    95\% C. I.      & $79.80 \pm 2.80$              & $79.60 \pm 0.60$   & $80.60 \pm 4.80$     & $1.70 \pm 1.10$     & $86.40 \pm 1.60$    \\ \hline
    
    \multicolumn{6}{l}{\textbf{Tossing Trajectory Prediction}} \\ \hline
    RMSE            & $0.77 \pm 0.01$   &  $0.74 \pm 0.06$       & $0.81 \pm 0.10$    & $0.50 \pm 0.05$    & $\mathbf{0.32 \pm 0.04}$  \\
    Residual       & $0.48 \pm 0.03$   &  $\mathbf{0.62 \pm 0.09}$       & $1.68 \pm 0.21$   &  $0.44 \pm 0.08$  & $0.64 \pm 0.02$ \\
    
    Std. Dev.        & $1.67 \pm 0.02$ & $1.67 \pm 0.01$ & $0.45 \pm 0.12$ & $\mathbf{0.09 \pm 0.01}$ & $0.09 \pm 0.04$ \\
    
    95\% C. I.      & $\mathbf{99.90 \pm 0.03}$   &  $99.90 \pm 0.11$       & $71.39 \pm 17.10$  &  $29.50 \pm 4.34$  & $40.80 \pm 11.90$ \\ \hline
    \bottomrule
    \end{tabular}
    \end{adjustbox}
\end{table}
% \footnotetext{For non-ideal physics dataset, the best model has the residual error closest to the residual error of the ground truth points.}

\subsection{Case Study: Collision Speed Estimation}
% We use two simulation datasets to evaluate our model performance on conditions where the prior physics knowledge is incomplete or the physics has simplistic assumptions. 

% \subsubsection{Collision Speed Estimation:}
In this real-world problem, we are given the initial speeds $\{v_{a1}, v_{b1}\}$ of two objects $\{a, b\}$ with their respective masses $\{m_{a}, m_{b}\}$ and the initial distance $d$ between them as inputs, $X = \{v_{a1}, v_{b1}, m_{a}, m_{b}, d\}$. The goal is to estimate their speed after collision $\{v_{af}, v_{bf}\}$. For a perfectly elastic collision, both objects should conserve momentum and energy after the collision, i.e., there is no loss in energy and momentum of the entire system, represented as follows. 
\begin{align*}
    m_{a}v_{a_{1}} + m_{b}v_{b_{1}} &= m_{a}v_{a_{f}} + m_{b}v_{b_{f}}\\
    \frac{1}{2}m_{a}v_{a_{1}}^2 + \frac{1}{2}m_{b}v_{b_{1}}^2 &= \frac{1}{2}m_{a}v_{a_{f}}^2 + \frac{1}{2}m_{b}v_{b_{f}}^2
\end{align*}
However, due to the presence of sliding friction during the simulation \citep{ba2019blending}, these ideal physical constraints are violated. 
Table \ref{tab:non-ideal} shows the performance of comparative models on this dataset, where we can see that PID-GAN outperforms all other baselines by a significant margin. At a first glance, it might seem strange that the conservation equations  worsen the performance of the models since the cGAN model is the closest competitor to the PID-GAN. However, due to the presence of sliding friction, these equations are violated for ground-truth predictions, and hence models that blindly use these imperfect physics equations (all physics-informed approaches except PID-GAN) show larger test errors. The mean of physics residuals on  ground-truth  is in fact equal to 93.96. We can see that the residuals of PID-GAN is closest to that of ground-truth. With a lower standard deviation and significantly higher 95\% C.I, PID-GAN also produces better uncertainty estimates.

\begin{figure}[b]
\centering
\includegraphics[scale=0.30]{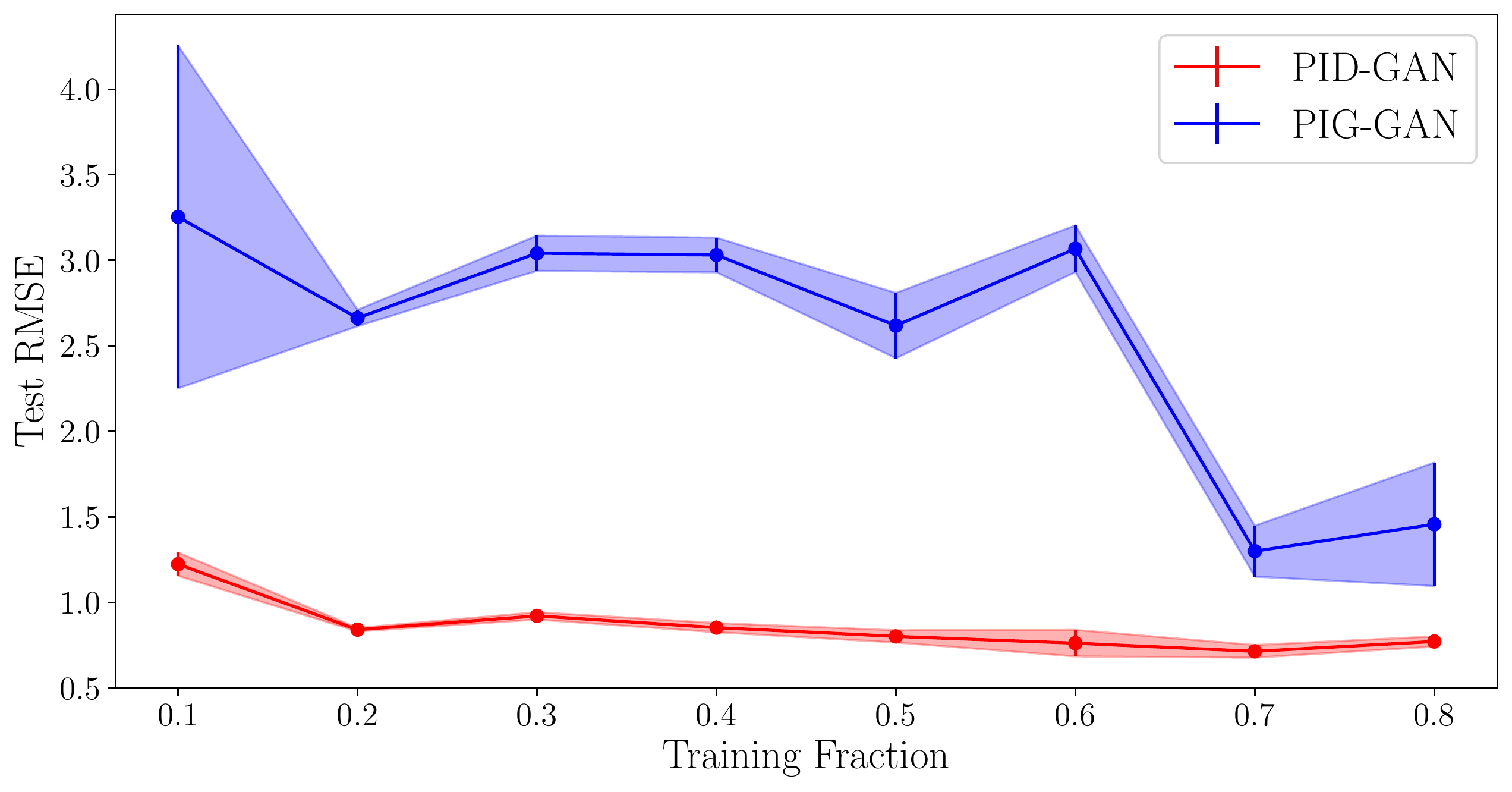}
\caption{Effect of training size on Collision Dataset}
\label{fig:collision_trn_frac}
\end{figure}

\par \noindent \textbf{Effect of Varying Training Size:} 
To demonstrate the ability of PID-GAN for showing better generalizability even in the paucity of labeled samples, Figure \ref{fig:collision_trn_frac} shows the performance of PID-GAN and PIG-GAN over different training fractions for the collision dataset. We can see that PID-GAN results do not degrade as drastically as PIG-GAN as we reduce the training size. This is because PIG-GAN only uses labeled instances for training the discriminator, which can be susceptible to learning spurious solutions when the labeled set is small. In contrast, the discriminator of PID-GAN uses both labeled and unlabeled instances and hence leads to the learning of more generalizable solutions.

\begin{figure}[t]
\centering
% \subfigure[\textbf{Epoch 0}]{\label{fig:collision_0} \includegraphics[scale=0.31]{figures/colision/PID_collision_Epoch_0.pdf}} 
\subfigure[\textbf{PIG-GAN}]{\label{fig:pig_collision_5000} \includegraphics[scale=0.31]{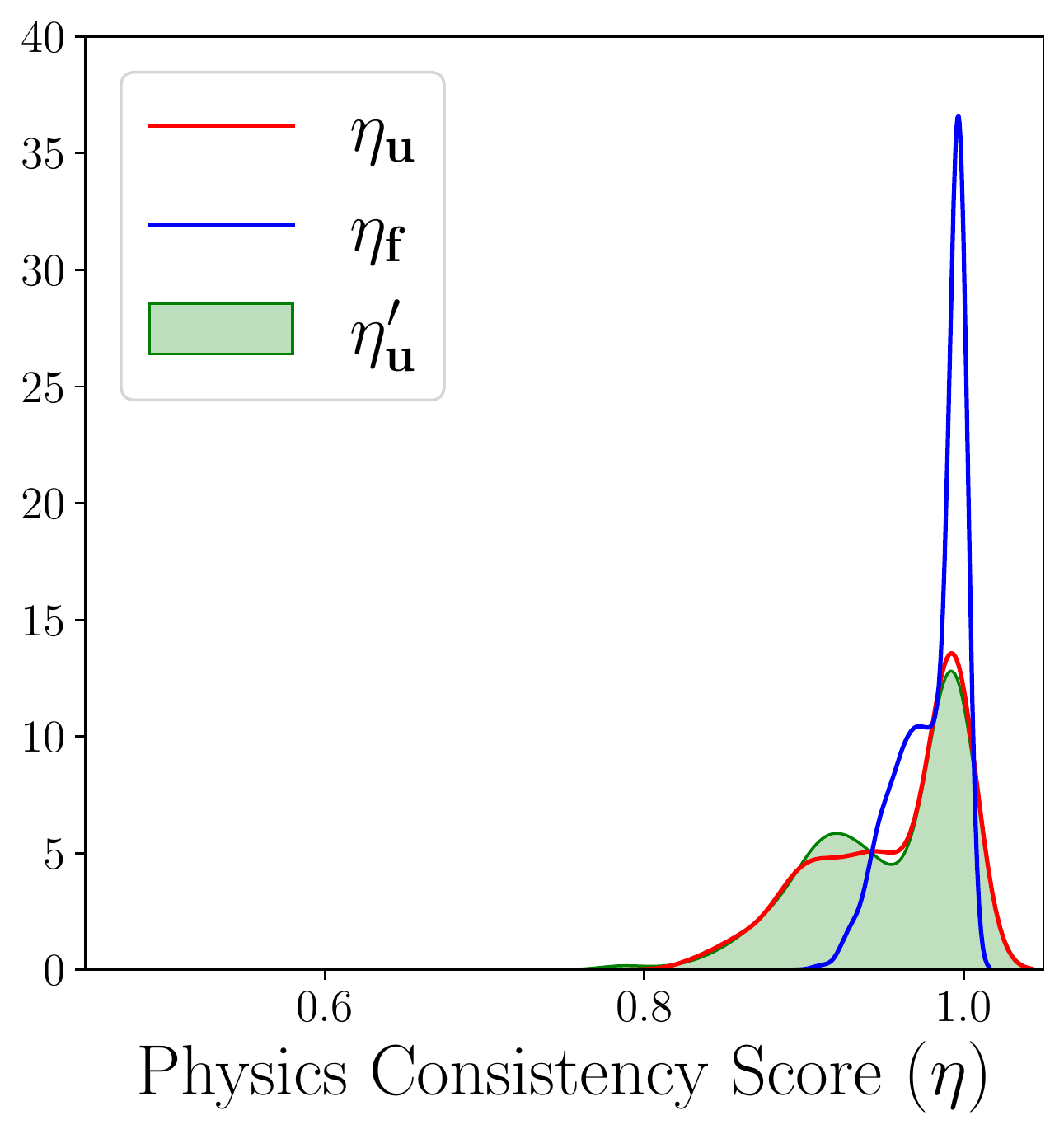}}
\subfigure[\textbf{PID-GAN}]{\label{fig:pid_collision_5000} \includegraphics[scale=0.31]{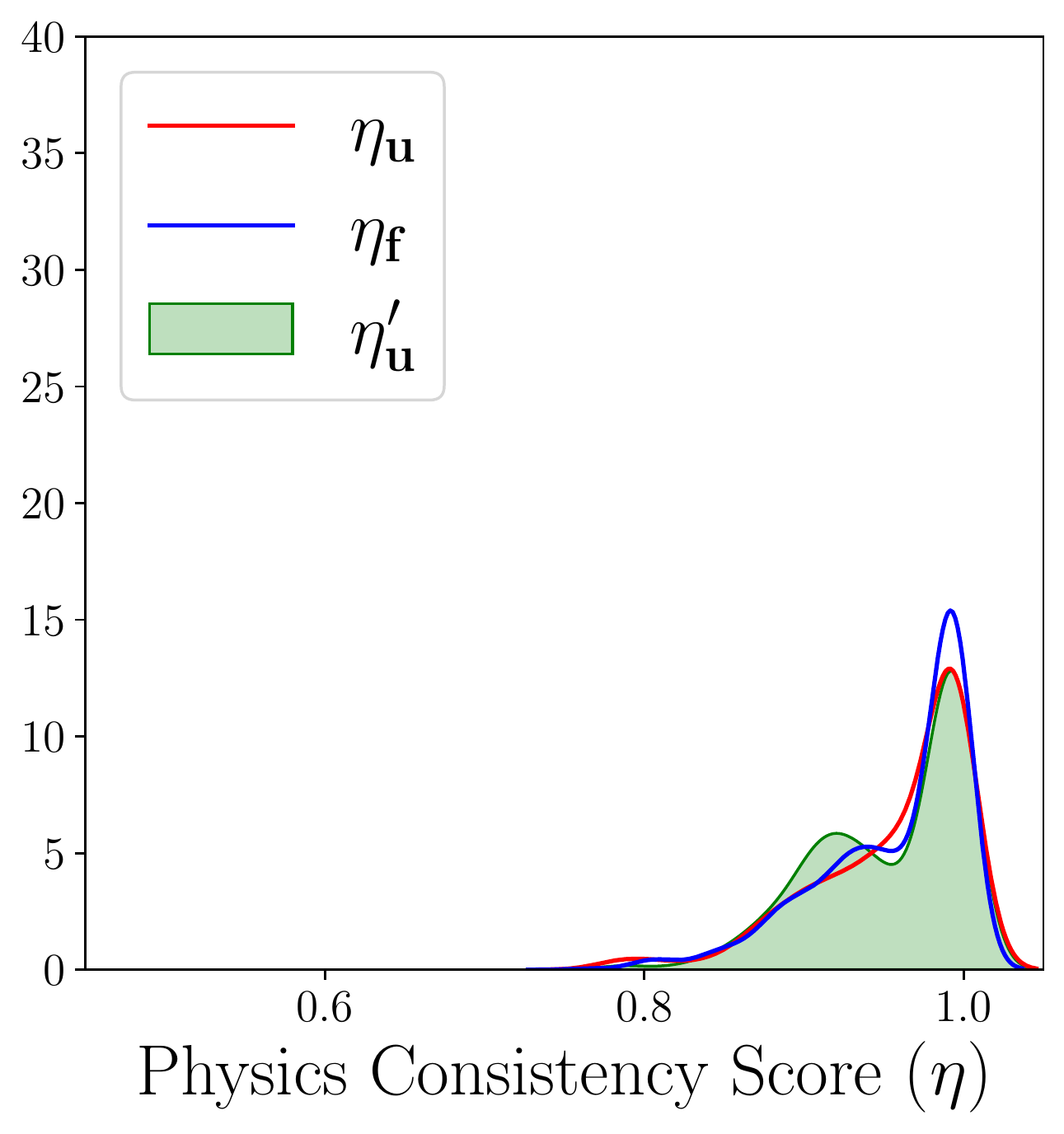}} 
\vspace{-3ex}
\caption{Physics Consistency Score for Collision Dataset}
\label{fig:collision_score}
\end{figure}

\par \noindent \textbf{Analysis of Physics Consistency Scores:}
Figure \ref{fig:pig_collision_5000} shows the distribution of the physics consistency score at the last epoch for both PID-GAN and PIG-GAN. Since this problem involves imperfect physics, we can see that physics consistency scores on ground truth labels, $\eta_{\mathbf{u}}'$ (green), is not always exactly equal to 1. While the consistency scores of PIG-GAN predictions on labeled points, $\eta_{\mathbf{u}}$ (red), matches the distribution on ground truth $\eta_{\mathbf{u}}'$, it blindly maximizes the physics consistency scores on the unlabeled (test) points,  $\eta_{\mathbf{f}}$ (blue). This is because the discriminator of PIG-GAN is only trained on labeled points and thus is unable to model the physics imperfections on the unlabeled points. On the other hand, the physics consistency scores of PID-GAN predictions accurately match the distribution of consistency scores on ground-truth, on both labeled as well as unlabeled points. This demonstrates the fact that PID-GAN makes effective use of label  and physics supervision for training both the generator and discriminator models, thus capturing the distributions of physics consistency scores accurately on both labeled and unlabeled sets.
% This behavior of minimizing the residuals is not observed on the labeled set due to the fact that adversarial loss forces the predictions of the PIG-GAN to be similar to the ground truth predictions on the labeled set. On the other hand, Figure \ref{fig:pid_collision_5000} illustrates that the PID-GAN imitates the true distribution of $\eta'_{\mathbf{u}}$ on both labeled and unlabeled set. This leads to a better overall generalization of the predictions.

\subsection{Case Study: Tossing Trajectory Prediction}
In this problem, we are given the initial three positions of an object as inputs, $X = \{l_{1}, l_{2}, l_{3}\}$, and we want to predict the position of the object for the next 15 time-stamps represented as $\{l_{4}, l_{5}, \cdots, l_{15}\}$. In a two-dimensional system, where the position of an object at time $i$ can be represented as $l_{i} = (l_{x}, l_{y})$, we can adopt the following elementary kinematics equations to model the free-fall physics as additional constraints. 
\begin{align*}
    l_{x_{i}} &= l_{x_{1}} + v_{x}t_{i} \\
    l_{y_{i}} &= l_{y_{1}} + v_{y}t_{i} - \frac{1}{2}gt_{i}^{2}
\end{align*}
where, $l_{x_{i}}$ and $l_{y_{i}}$ are the object location at time $t_{i}$, $v_{x}$ and $v_{y}$ are the horizontal and vertical component of the initial velocity, and g is the gravitational acceleration $9.8 ms^{-2}$. To introduce imperfect physics scenarios, random accelerations as winds and additional damping factor to imitate air resistance were introduced in the dataset \citep{ba2019blending}.
% \subsubsection{Results for simulation datasets}
We can see from Table \ref{tab:non-ideal} that PID-GAN outperforms all other baselines for this problem by a significant margin in terms of RMSE. APINN-Drop and PID-GAN perform similarly in terms of residual error (close to the ground truth residual of 0.59); however, APINN-Drop has a much larger RMSE, which makes it worse than PID-GAN. Although the C.I scores of PINN-Drop and APINN-Drop are almost perfect (close to 100), their standard deviation are significantly high, indicating that they are producing under-confident predictions.

\vspace{-1ex}
\subsection{Case Study: Solving Physics-based PDEs}
\label{sec:pdes}
\begin{table*}[t]
    \caption{Summary of model performances for solving Burgers', Darcy's and Schr\"{o}dinger equations in terms of the mean and standard deviation for 5 random runs. $95\%$ C.I. corresponds to empirical coverage of $95\%$ predictive intervals.}
    \label{tab:pdes}
    \centering
    \begin{adjustbox}{width=2.2\columnwidth,center}
    \begin{tabular}{lllllllll}
\toprule                                                                                                                                
\multicolumn{1}{l|}{Condition}  & \multicolumn{4}{c|}{Noise-free}                                                                                                       & \multicolumn{4}{c}{Noisy}                                                                                                 \\ \hline
\multicolumn{1}{l|}{Models}     & \multicolumn{1}{l}{PINN-Drop} & \multicolumn{1}{l}{APINN-Drop} & \multicolumn{1}{l}{PIG-GAN} & \multicolumn{1}{l|}{\textbf{PID-GAN}}            & \multicolumn{1}{l}{PINN-Drop} & \multicolumn{1}{l}{APINN-Drop} & \multicolumn{1}{l}{PIG-GAN} & \multicolumn{1}{l}{\textbf{PID-GAN}} \\ \hline
\multicolumn{9}{l}{\textbf{Burgers' Equation}}  \\ \hline
\multicolumn{1}{l|}{Error-u}    & $0.380 \pm 0.027$         &  $0.260 \pm 0.019$         & $0.215 \pm 0.205$           & \multicolumn{1}{l|}{$\mathbf{0.100 \pm 0.008}$}    & $0.370 \pm 0.020$          &  $0.250 \pm 0.040$          & $0.150 \pm 0.125$            & $\mathbf{0.116 \pm 0.028}$           \\
\multicolumn{1}{l|}{Residual}   & $0.0040 \pm 0.0003$       &  $0.0330 \pm 0.0040$        & $0.1570 \pm 0.2790$           & \multicolumn{1}{l|}{$\mathbf{0.0010 \pm 0.0004}$} & $0.0040 \pm 0.0005$       &  $0.0360 \pm 0.0100$         & $0.0300 \pm 0.0404$           & $\mathbf{0.0020 \pm 0.0003}$          \\
\multicolumn{1}{l|}{Std. Dev.}   &  $0.05 \pm 0.00$  &  $0.06 \pm 0.00$  &   $0.04 \pm 0.01$  & \multicolumn{1}{l|}{$0.04 \pm 0.00$}  &   $0.05 \pm 0.01$   &   $0.06 \pm 0.00$  &   $0.04 \pm 0.01$ &     $0.04 \pm 0.01$ \\
\multicolumn{1}{l|}{95\% C. I. (\%)} & $55.89 \pm 0.58$    &  $68.69 \pm 1.10$          & $63.48 \pm 35.64$           & \multicolumn{1}{l|}{$81.16 \pm 8.53$}   & $56.65 \pm 0.88$         &  $69.02 \pm 4.27$         & $75.54 \pm 33.15$           & $80.14 \pm 14.63$           \\ \hline

\multicolumn{9}{l}{\textbf{Schr\"{o}dinger Equation}}       \\ \hline
\multicolumn{1}{l|}{Error-h}    & $0.427 \pm 0.001$       &  $0.402 \pm 0.001$         & $0.112 \pm 0.014$           & \multicolumn{1}{l|}{$\mathbf{0.045 \pm 0.012}$}           &  $0.428 \pm 0.001$       & $0.407 \pm 0.002$           & $\mathbf{0.079 \pm 0.008}$           &  $0.081 \pm 0.014$                   \\

\multicolumn{1}{l|}{Residual}   & $0.0100 \pm 0.0002$     &   $0.0770 \pm 0.0002$      & $0.0100 \pm 0.0019$         & \multicolumn{1}{l|}{$\mathbf{0.0013 \pm 0.0002}$}         &   $0.0096 +\pm 0.0004$   & $0.0740 \pm 0.002$           & $0.0339 \pm 0.0058$         &  $\mathbf{0.0007 \pm 0.0003}$                 \\
\multicolumn{1}{l|}{Std. Dev.}   & $0.04 \pm 0.00$  &  $0.04 \pm 0.00$  &  $0.03 \pm 0.01$  & \multicolumn{1}{l|}{$0.01 \pm 0.01$}  &  $0.04 \pm 0.00$  &  $0.04 \pm 0.00$  &    $0.02 \pm 0.00$  &   $0.08 \pm 0.02$\\
\multicolumn{1}{l|}{95\% C. I. (\%)} & $69.90 \pm 0.36$    &   $40.80 \pm 1.57$          & $68.72 \pm 19.34$           & \multicolumn{1}{l|}{$47.49 \pm 26.75$}           &   $64.20 \pm 1.11$        & $35.90 \pm 1.70$             & $49.67 \pm 12.96$           &  $79.82 \pm 11.77$                   \\ \hline

\multicolumn{9}{l}{\textbf{Darcy's Equation}}           \\ \hline
\multicolumn{1}{l|}{Error-u}    & $0.010 \pm 0.002$         &  $\mathbf{0.009 \pm 0.002}$ & $0.013 \pm 0.012$         & \multicolumn{1}{l|}{$\mathbf{0.009 \pm 0.006}$}           & $0.012 \pm 0.002$        &  $0.012 \pm 0.001$         & $0.022 \pm 0.008$           &  $\mathbf{0.011 \pm 0.003}$          \\
\multicolumn{1}{l|}{Error-k}    & $0.450 \pm 0.011$         &  $0.478 \pm 0.013$        & $0.102 \pm 0.039$           & \multicolumn{1}{l|}{$\mathbf{0.082 \pm 0.043}$}  & $0.462 \pm 0.016$        &  $0.540 \pm 0.015$         & $0.197 \pm 0.045$           &  $\mathbf{0.159 \pm 0.020}$           \\
\multicolumn{1}{l|}{Residual}   & $0.0080 \pm 0.0005$       &  $0.0110 \pm 0.0003$       & $0.0005 \pm 0.0003$         & \multicolumn{1}{l|}{$\mathbf{0.0002 \pm 0.0001}$}& $0.0080 \pm 0.0004$       &  $0.0140 \pm 0.0007$       & $0.0030 \pm 0.0020$         &  $\mathbf{0.0020 \pm 0.0006}$         \\
\multicolumn{1}{l|}{Std. Dev.}   & $0.36 \pm 0.01$  & $0.34 \pm 0.01$  &  $0.11 \pm 0.02$  & \multicolumn{1}{l|}{$0.08 \pm 0.01$}  & $0.36 \pm 0.00$   &   $0.35 \pm 0.02$   &  $0.15 \pm 0.05$  &  $0.09 \pm 0.02$ \\
\multicolumn{1}{l|}{95\% C. I. (\%)} & $100.0 \pm 0.00$     &  $100.0 \pm 0.00$ & $84.22 \pm 31.56$           & \multicolumn{1}{l|}{$89.42 \pm 13.46$}           & $100.0 \pm 0.00$          &  $100.0 \pm 0.00$          & $88.70 \pm 13.69$            &  $98.97 \pm 0.65$           \\  \bottomrule
\end{tabular}
\end{adjustbox}
    
\end{table*}

We test the efficacy of our approach on the problem of solving three physics-based PDEs, which have been used as benchmark in existing literature on PINN \citep{yang2019adversarial,raissi2019physics}. with applications spanning multiple domains. In particular, we study the Burgers', Schr\"odinger, and Darcy's Equations, briefly described in the following.

\par \noindent \textbf{Burgers' Equation:}
 We represent the nonlinear time ($t$) dependent Burgers' equation in one spatial dimension ($x$) as:
\vspace{-1ex}
\begin{align*}
\vspace{-2ex}
    &u_{t} + uu_{x} + \nu u_{xx} = 0, \quad  x \in [-1, 1], \: t \in [0, 1], \\
    &u(0, x) = -\sin(\pi x), \quad u(t, -1) = u(t, 1) = 0.
\end{align*}
The goal is to predict $u(t,x)$ as output given spatio-temporal coordinates, $x$ and $t$, as inputs.

\begin{figure*}[ht]
\centering
\subfigure[Burgers PIG-GAN]{\label{fig:Noisy_Burgers_Layer4_pig} \includegraphics[scale=0.20]{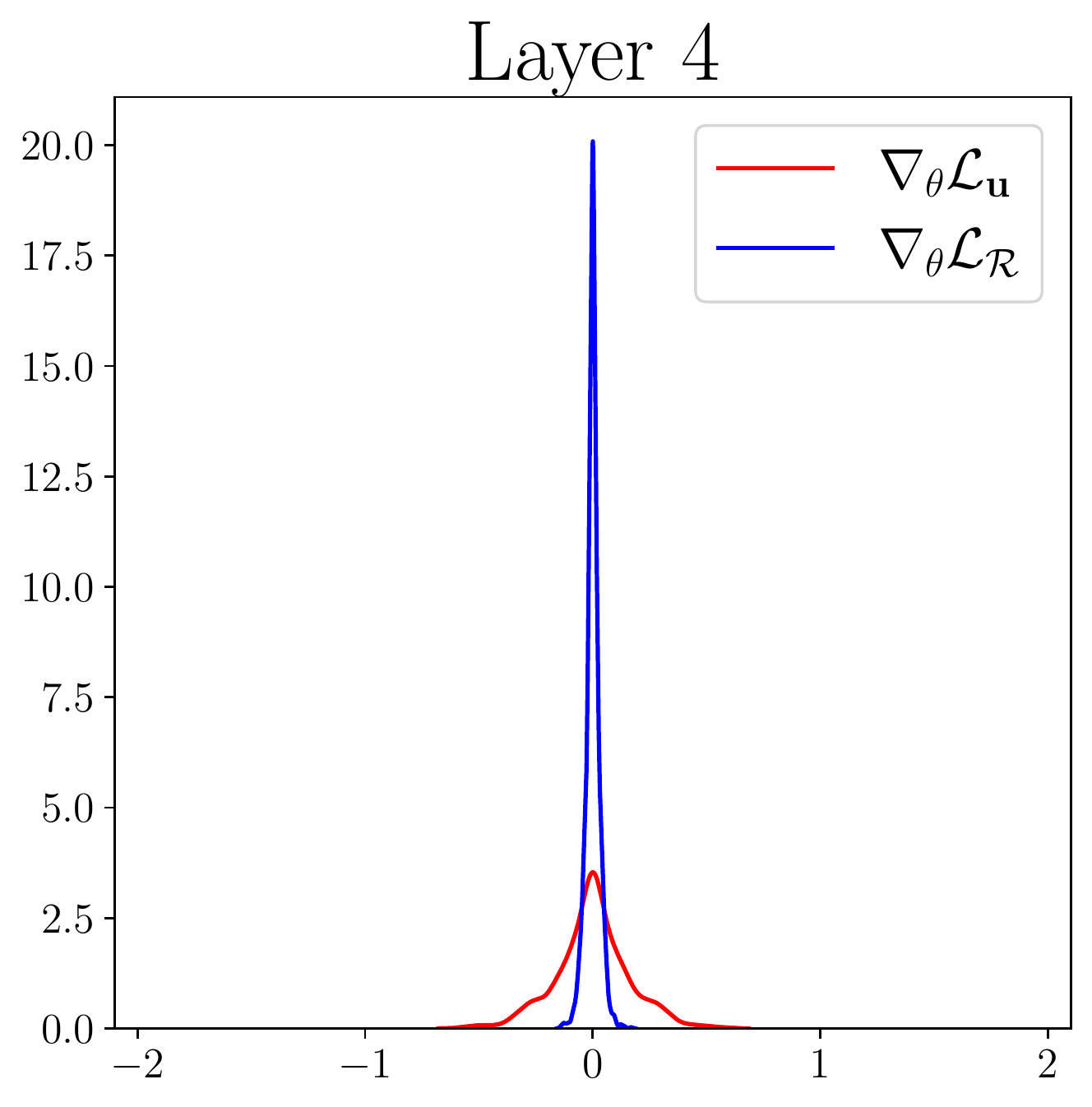}
% } 
% \subfigure[Burgers PIG-GAN]{\label{fig:Noisy_Burgers_Layer5_pig}
\includegraphics[scale=0.20]{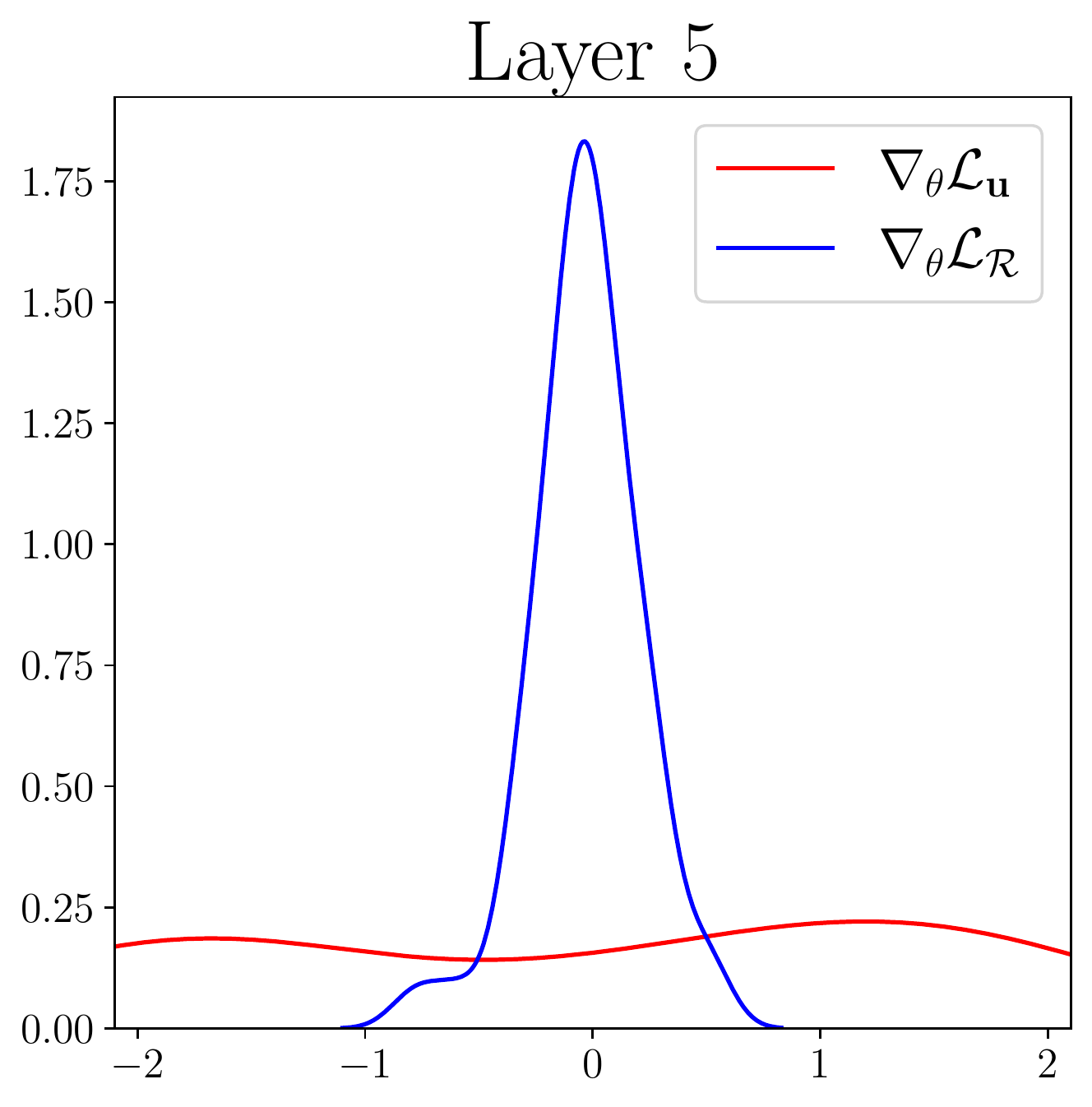}}
\subfigure[Schrodinger PIG-GAN]{\label{fig:Noisy_Schrodinger_Layer4_pig} \includegraphics[scale=0.20]{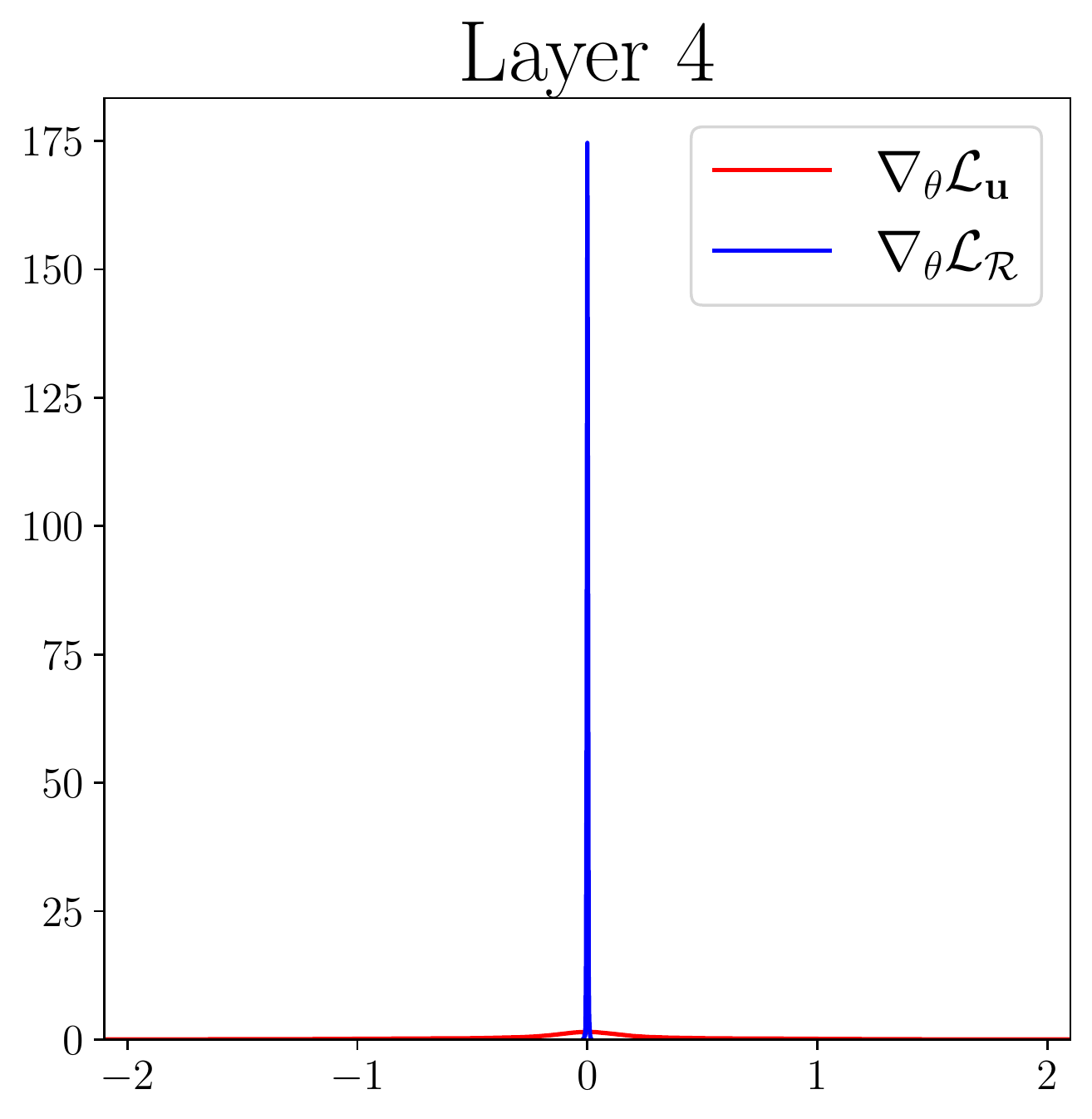}
% }
% \subfigure[Schrodinger PIG-GAN]{\label{fig:Noisy_Schrodinger_Layer5_pig} 
\includegraphics[scale=0.20]{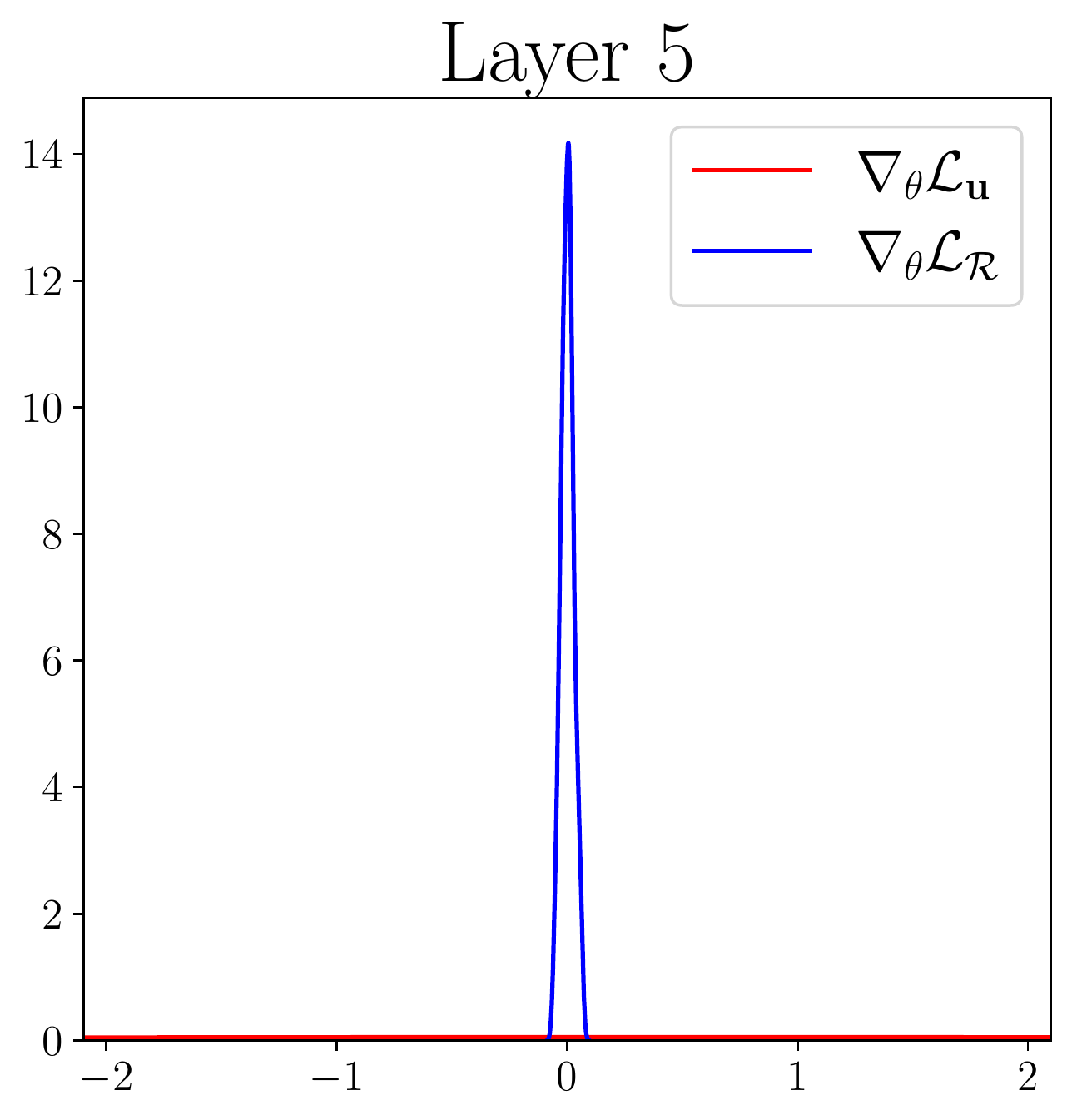}}
\subfigure[Darcy PIG-GAN]{\label{fig:Noisy_Darcy_Layer3_pig} \includegraphics[scale=0.20]{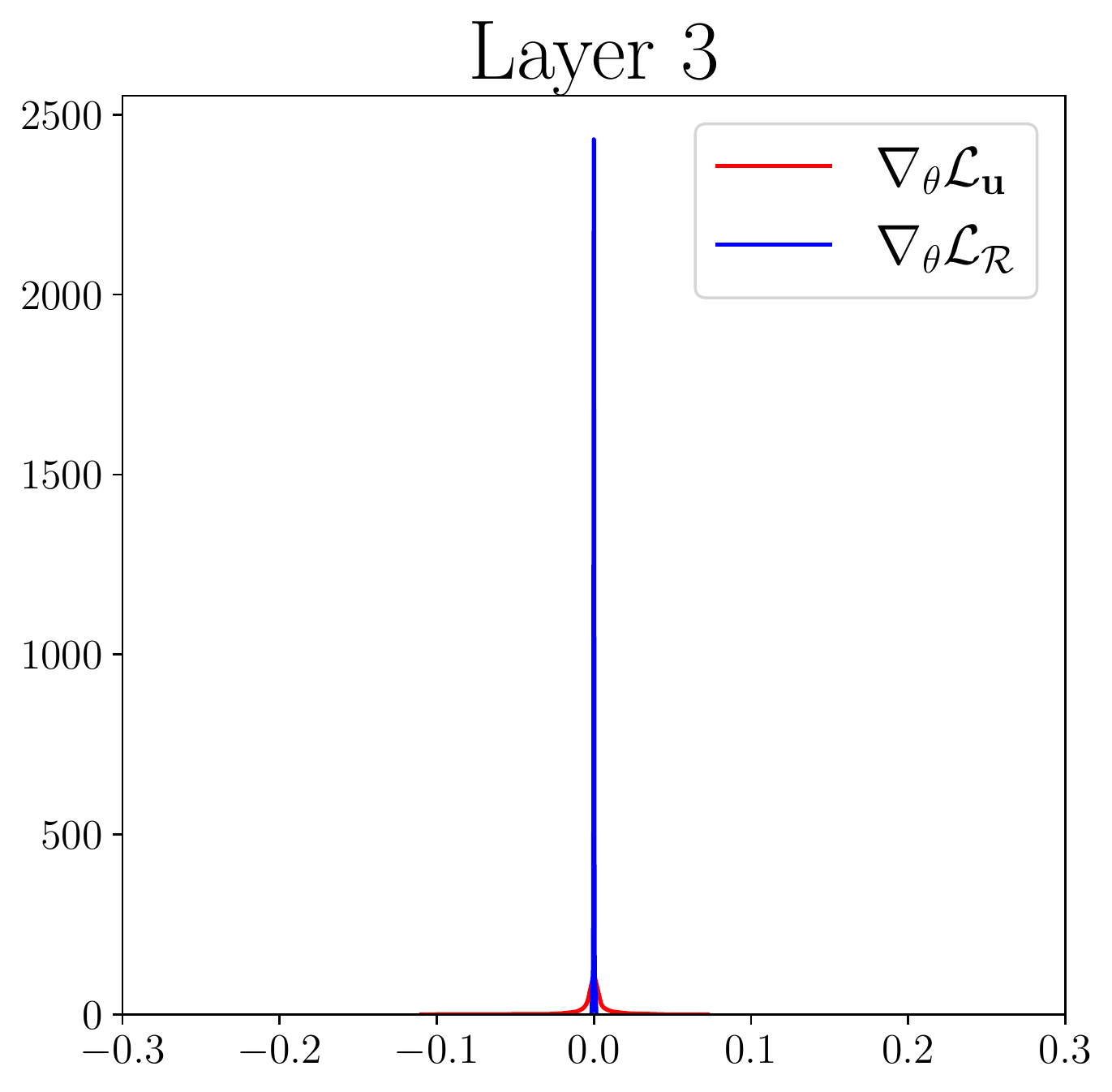}
% }
% \subfigure[Darcy PIG-GAN]{\label{fig:Noisy_Darcy_Layer4_pig}
\includegraphics[scale=0.20]{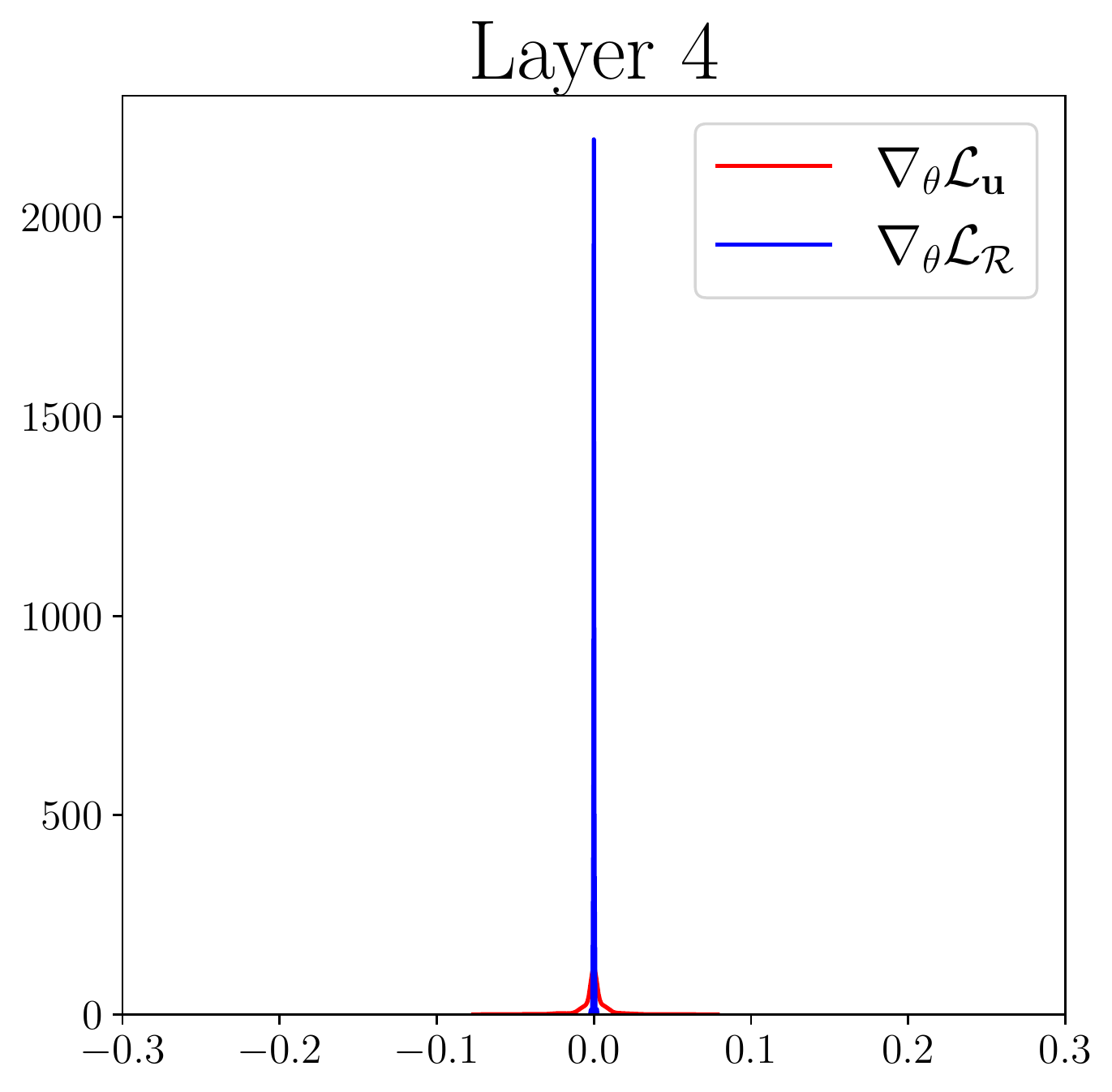}}

\subfigure[Burgers PID-GAN]{\label{fig:Noisy_Burgers_Layer4_pid} \includegraphics[scale=0.20]{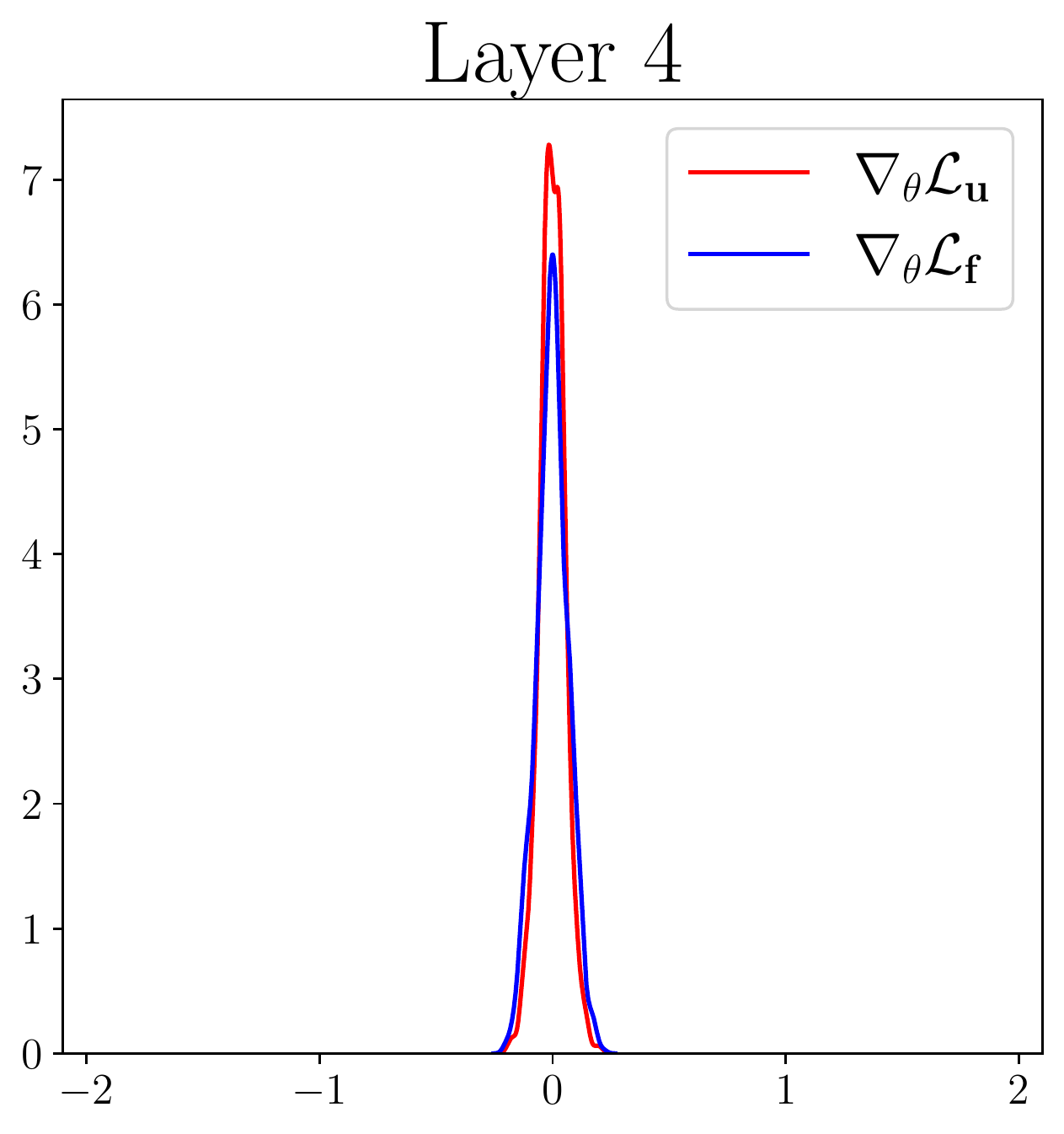}
% } 
% \subfigure[Burgers PID-GAN]{\label{fig:Noisy_Burgers_Layer5_pid}
\includegraphics[scale=0.20]{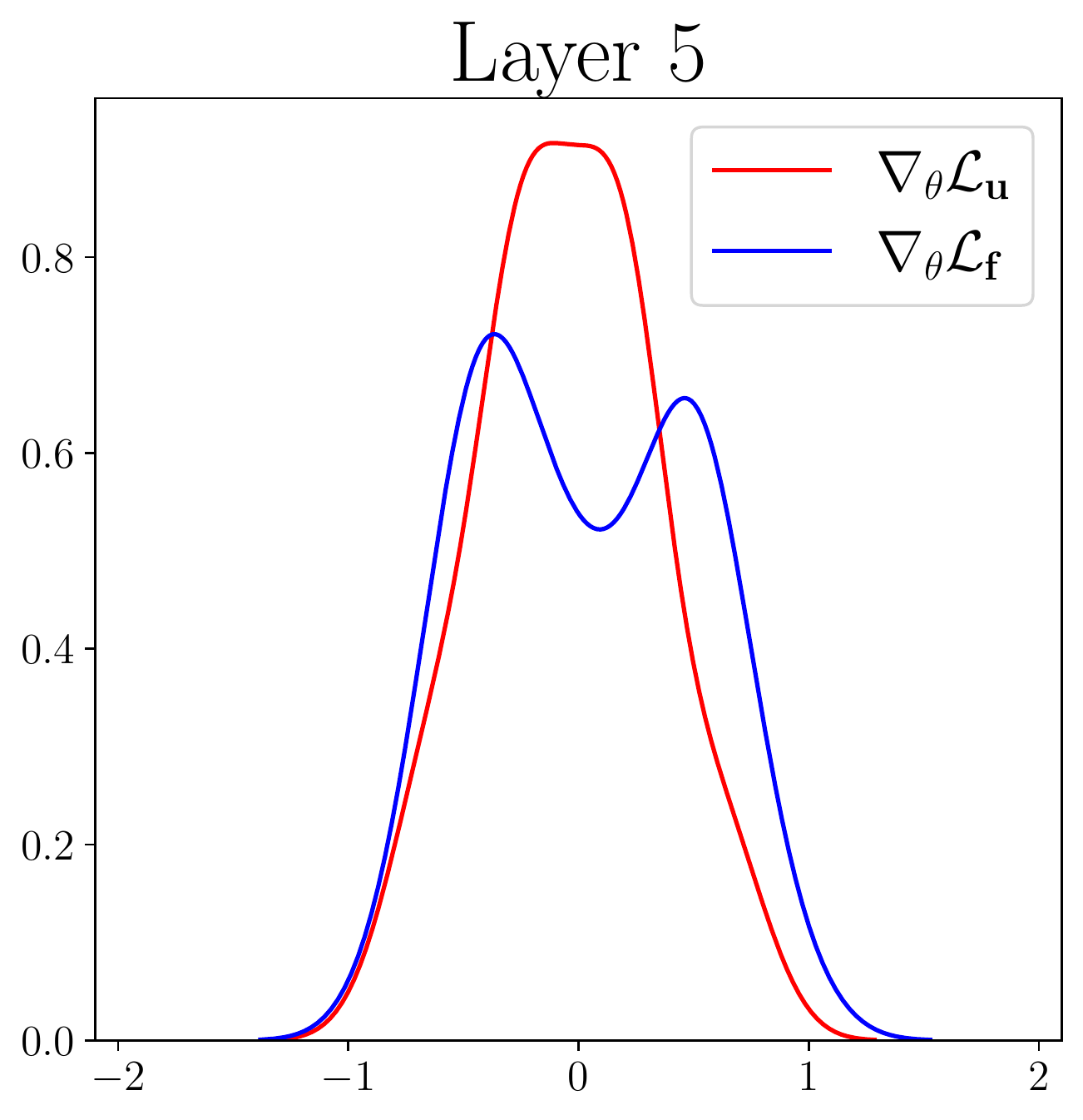}} 
\subfigure[Schrodinger PID-GAN]{\label{fig:Noisy_Schrodinger_Layer4_pid} \includegraphics[scale=0.20]{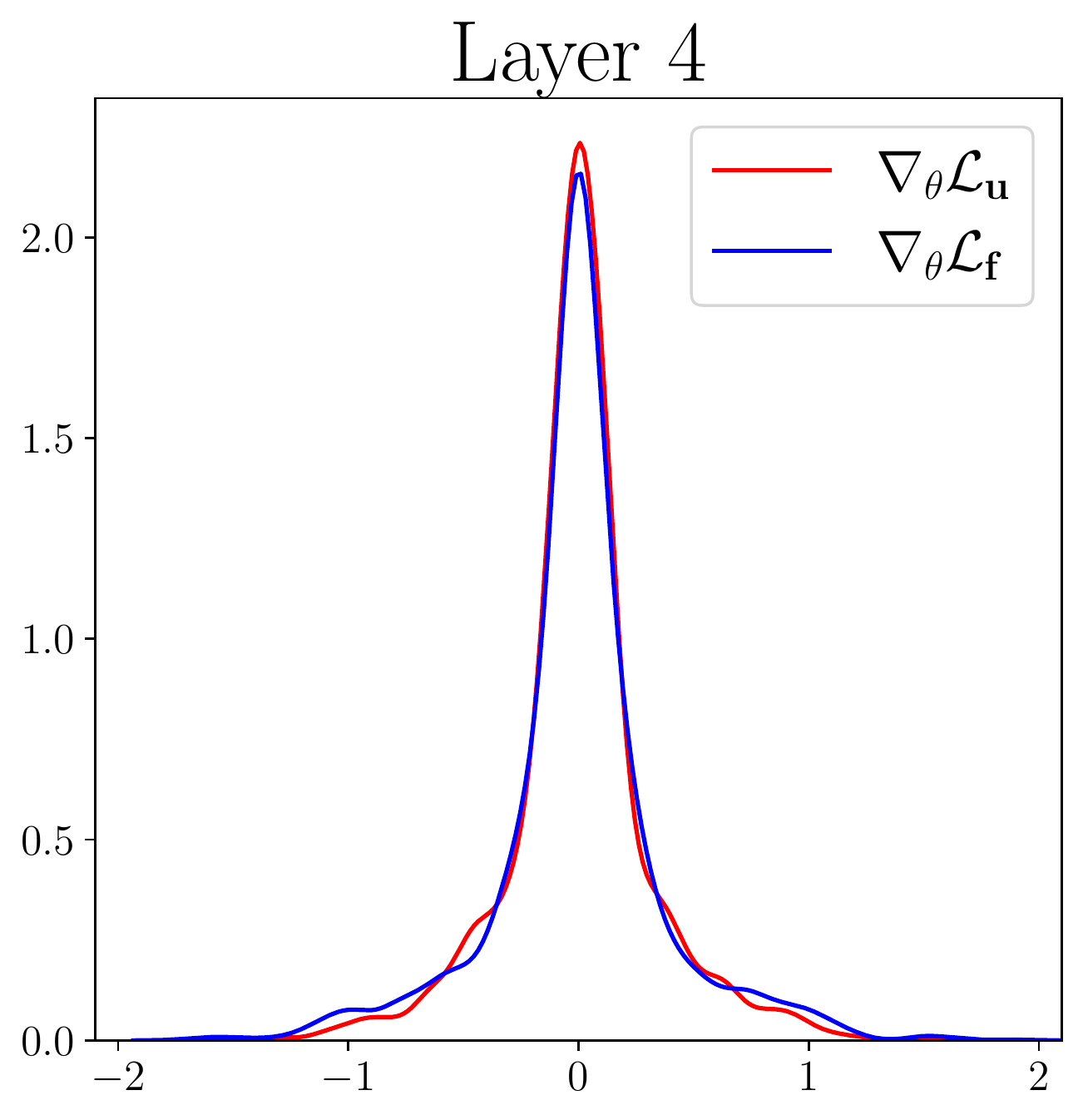}
% }
% \subfigure[Schrodinger PID-GAN]{\label{fig:Noisy_Schrodinger_Layer5_pid}
\includegraphics[scale=0.20]{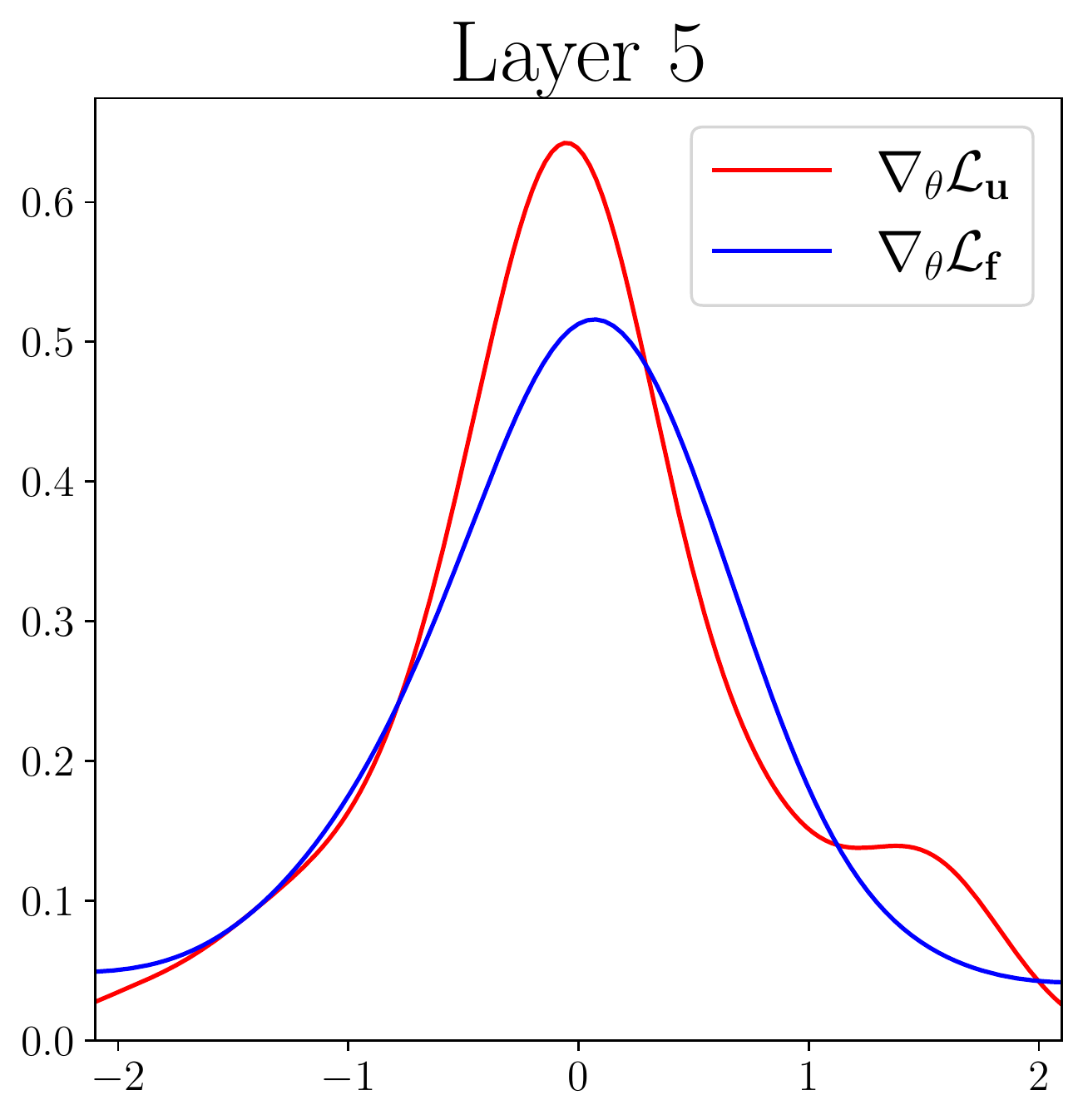}}
\subfigure[Darcy PID-GAN]{\label{fig:Noisy_Darcy_Layer3_pid} \includegraphics[scale=0.20]{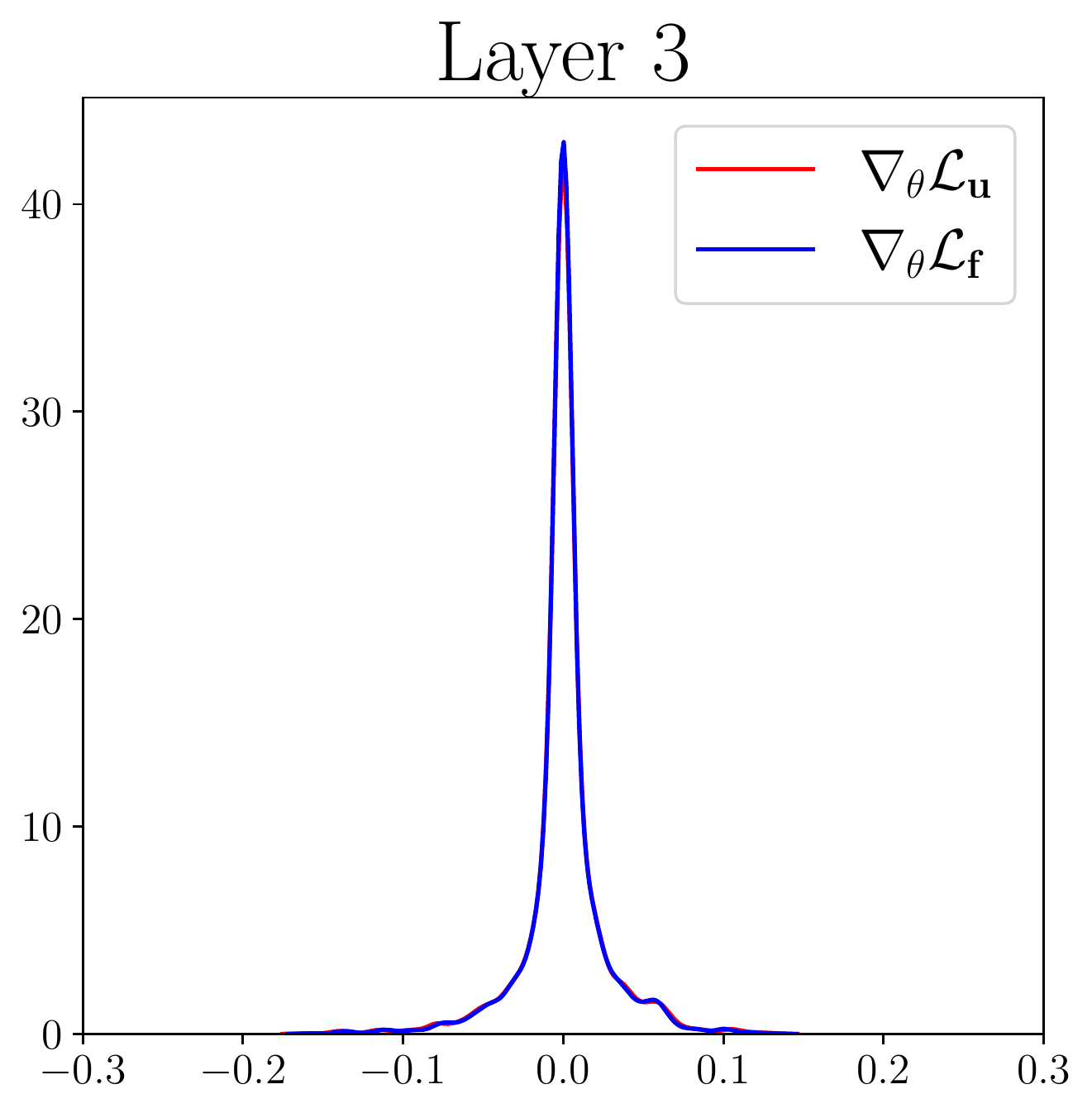}
% }
% \subfigure[Darcy PID-GAN]{\label{fig:Noisy_Darcy_Layer4_pid}
\includegraphics[scale=0.20]{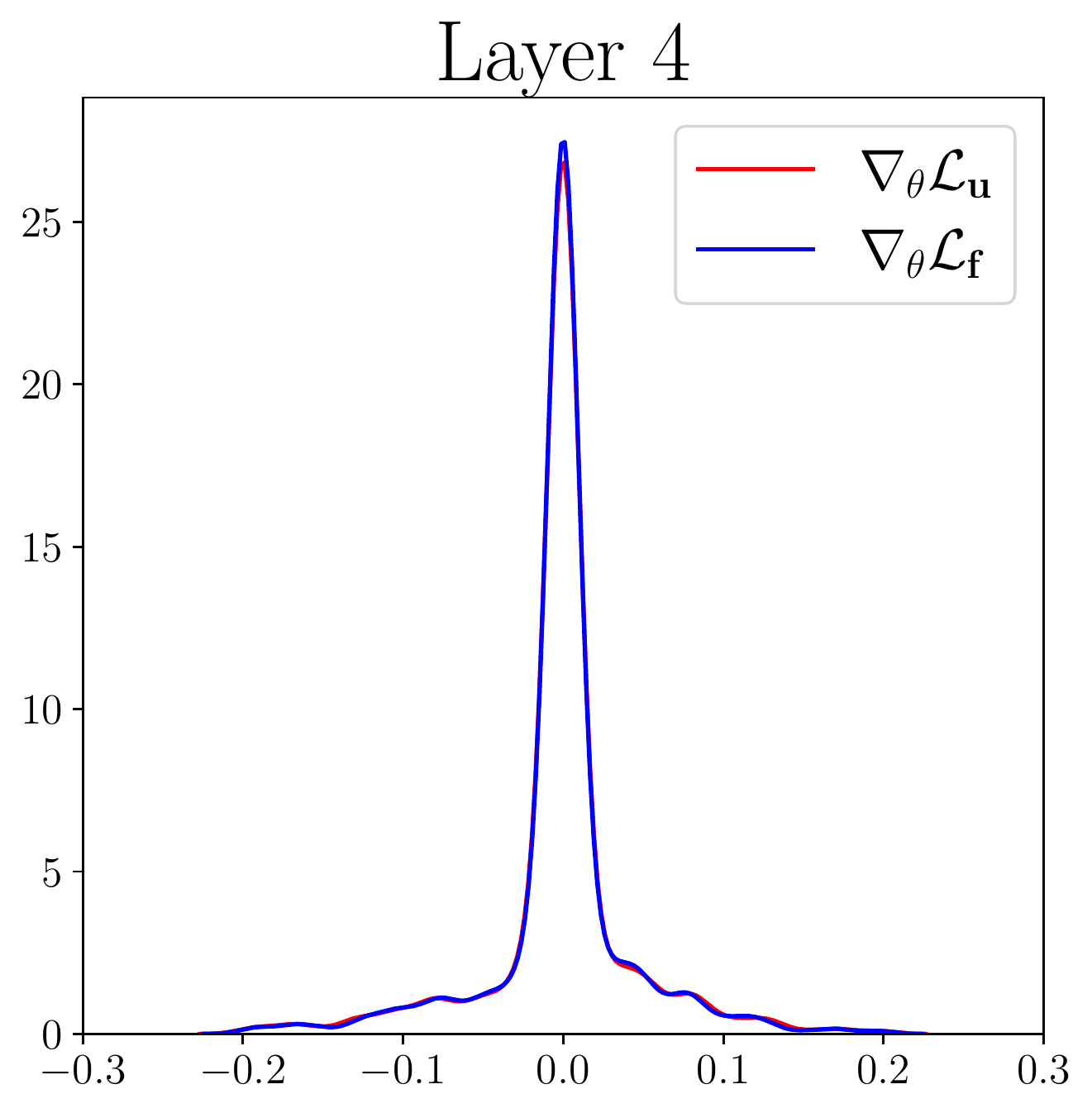}}

\vspace{-3ex}
\caption{Analyzing the Imbalance of Generator Gradients for PIG-GAN and PID-GAN on the three benchmark PDE problems}
\label{fig:generator_gradients}
\end{figure*}

\par \noindent \textbf{Schr\"{o}dinger Equation:}
We consider the problem of solving the one-dimensional nonlinear Schr\"{o}- dinger equation involving periodic boundary conditions and complex-valued solutions as follows: 
\begin{align*}
    &ih_{t} + 0.5h_{xx} + {|h|}^{2}h = 0, \quad x \in [-5, 5], \: t \in [0, \pi/2],  \\
    &h(x, 0) = 2 \: \text{sech}(x), \quad h(-5, t) = h(5, t), \quad h_{x}(-5, t) = h_{x}(5, t),
\end{align*}
where  $h(x, t) = u(x, t) + i v(x, t)$ is the complex-valued solution of the equation with $u(x, t)$ as the real part and $v(x, t)$ as the imaginary part. We predict the real and imaginary parts of $h$ as outputs, given $x$ and $t$ as inputs.  

\par \noindent \textbf{Darcy's Equation:}
We consider the problem of solving Darcy's equation, which is a two-dimensional nonlinear diffusion equation with an unknown state-dependent diffusion coefficient $k$.
\vspace{-1ex}
\begin{align*}
    &\nabla_{\mathbf{x}} \cdot \: [k(u) \nabla_{\mathbf{x}} u(\mathbf{x})] = 0, \quad \mathbf{x} =(x_{1}, x_{2}), \quad u(\mathbf{x}) = u_{0}, \quad x_{1} = L_{1} \\
    &-k(u) \frac{\partial u(\mathbf{x})}{\partial x_{1}} = q, \quad x_{1} = 0 , \quad \frac{\partial u(\mathbf{x})}{\partial x_{2}} = 0, \quad x_{2} = \{0, L_{2}\}
\end{align*}
The goal here is predict $u$ given $x_1$ and $x_2$ as inputs. Further, the labels for the diffusion coefficient $k$ are not provided during training but is expected to be learned by directly solving the PDE.

\par \noindent \textbf{Result Comparison on PDEs:}
For each PDE, we evaluate under two different conditions: deterministic (noise-free) and random boundary conditions (noisy). For noisy conditions, we add 10\% Gaussian uncorrelated noise to the ground-truth labels which augments the inherent epistemic uncertainty of our predictive model due to the randomness in the boundary conditions. 

Table \ref{tab:pdes} provides a summary of the comparison of different baselines across the three benchmark PDEs for noise-free and noisy conditions. For \textbf{Burger's equation}, we see that PID-GAN shows significant improvement in the relative $L^2$-error of the predictions $u$ in both noisy and noise-free settings. Additionally, the variance of the PID-GAN across 5 random runs is the least, suggesting that it is more robust to random initializations. The PID-GAN also achieves the lowest PDE residual of 1x$10^{-3}$ and 2x$10^{-3}$ in noisy and noise-free setting. This displays the ability of the PID-GAN to generate generalizable and physically consistent solutions. For Burgers' equation, PID-GAN also provides an average 81.16\% and 80.14\% empirical coverage of the 95\% confidence intervals for the two settings, i.e., approximately 80\% of the times the ground truth lies between  two standard deviations of the predictions. In general, it is preferred to have a higher 95\% C.I. with lower standard deviations. On Burger's equation, PID-GAN satisfies both these criteria. 

For \textbf{Schr\"odinger Equation}, in terms of relative $L^2$-error of the predictions $h$, the PID-GAN performs the best in noise-free setting while PIG-GAN performs slightly better than PID-GAN in the noisy setting (although the difference is not statistically significant and does not carry over for the other PDE experiments). On the other hand, PINN-Drop and  APINN-Drop performs significantly worse in both noisy and noise-free conditions. Further, we observe a significantly lower PDE residual for the PID-GAN. The 95\% C.I. for the PID-GAN is the best in the noisy setting with lowest std. dev., but in the noise-free setting, PINN-Drop shows highest 95\% C.I although with a higher std. dev., and a significantly higher $L^2$-error. 

For \textbf{Darcy's Equation}, all the baselines show quite similar results w.r.t. the $L^2$-error on the predictions $u$, with PID-GAN having a slight edge in both settings. However, when it comes to predicting $k$ (for which no ground truth values were provided during training), PINN-Drop and APINN-Drop performs significantly poor in terms of $L^2$-errors, with significantly larger standard deviations of samples than other baselines. As a result, even though their 95\% C.I. appear perfect, their predictions are not very useful. The GAN based models on the other hand are able to quite precisely estimate the value of $k$. Also, the PDE residuals are slightly better for the PID-GAN. 
% In the noise-free setting, PINN-Drop and APINN-Drop obtain a perfect 95\% C.I. with the other models slightly trailing behind. 
For the noisy setting, PID-GAN is able to get almost perfect 95\% C.I. Additional analyses of the results comparing PID-GAN and APINN-Drop have been provided in Appendix \ref{sec-additionalresult}.

\par \noindent \textbf{Analyzing Imbalance in Generator Gradients:}
Figure \ref{fig:generator_gradients} provides a comparison of the gradients obtained from the last two layers of the generator of PIG-GAN and PID-GAN for the three benchmark PDEs at the last epoch. 
For the Burgers' equation, we can observe that the gradients of PIG-GAN on the labeled set ($\nabla_{\theta}\mathcal{L}_{\mathbf{u}}$) (i.e., the initial and the boundary points) has a much wider distribution than the gradients of the PDE residuals ($\nabla_{\theta}\mathcal{L}_{\mathbf{f}}$) computed on the unlabeled set. This imbalance between the gradients indirectly demonstrates the imbalance between the gradient contributions of labeled and unlabeled points in PIG-GAN, supporting the theoretical claims made in Section 3. 
This problem of PIG-GAN aggravates for the Schr\"odinger's equation, where we can see that $\nabla_{\theta}\mathcal{L}_{\mathbf{u}}$ has an extremely large variance while the variance of $\nabla_{\theta}\mathcal{L}_{\mathbf{f}}$ is ultra-low. However, for the Darcy's Equation, the imbalance between the gradients of PIG-GAN is not very visible, indicating that the problem of gradient imbalance in PIG-GAN is use-case dependent. 
On the other hand, for PID-GAN, the generator gradients for both the labeled and unlabeled set follow the same distribution across all three PDEs, showing no signs of gradient imbalance. 
% . This is not the case for PID-GAN, which by design would always give balanced gradients thus leading to a more stable training dynamics.
% We demonstrate some empirical results to support the claims made in Section \ref{sec:methods}. 

It must be noted that we only visualize the effect of gradient imbalance in the last two layers of the generator. The magnitude of these gradients decreases as we back-propagate deeper into the network. This same phenomena leads to the common problem of vanishing gradients. Thus, the effect of imbalance in the last few layers would be more prominent than its effect in the earlier layers.

\begin{figure}[t]
\centering
\subfigure[Burgers' PIG-GAN]{\label{fig:Burgers_pig_logits} \includegraphics[scale=0.3]{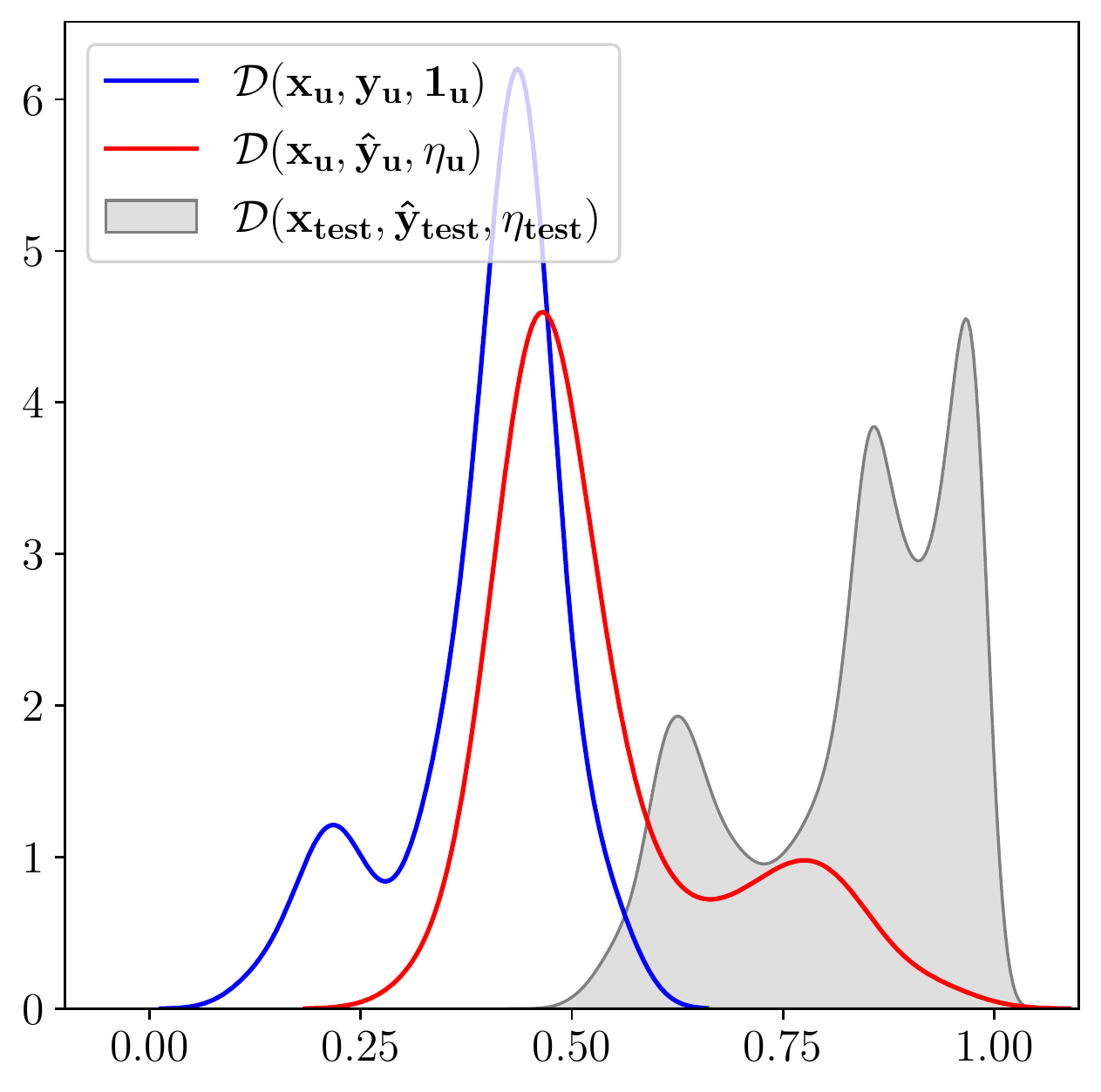}} 
\subfigure[Burgers' PID-GAN]{\label{fig:Burgers_discriminator_logits} \includegraphics[scale=0.3]{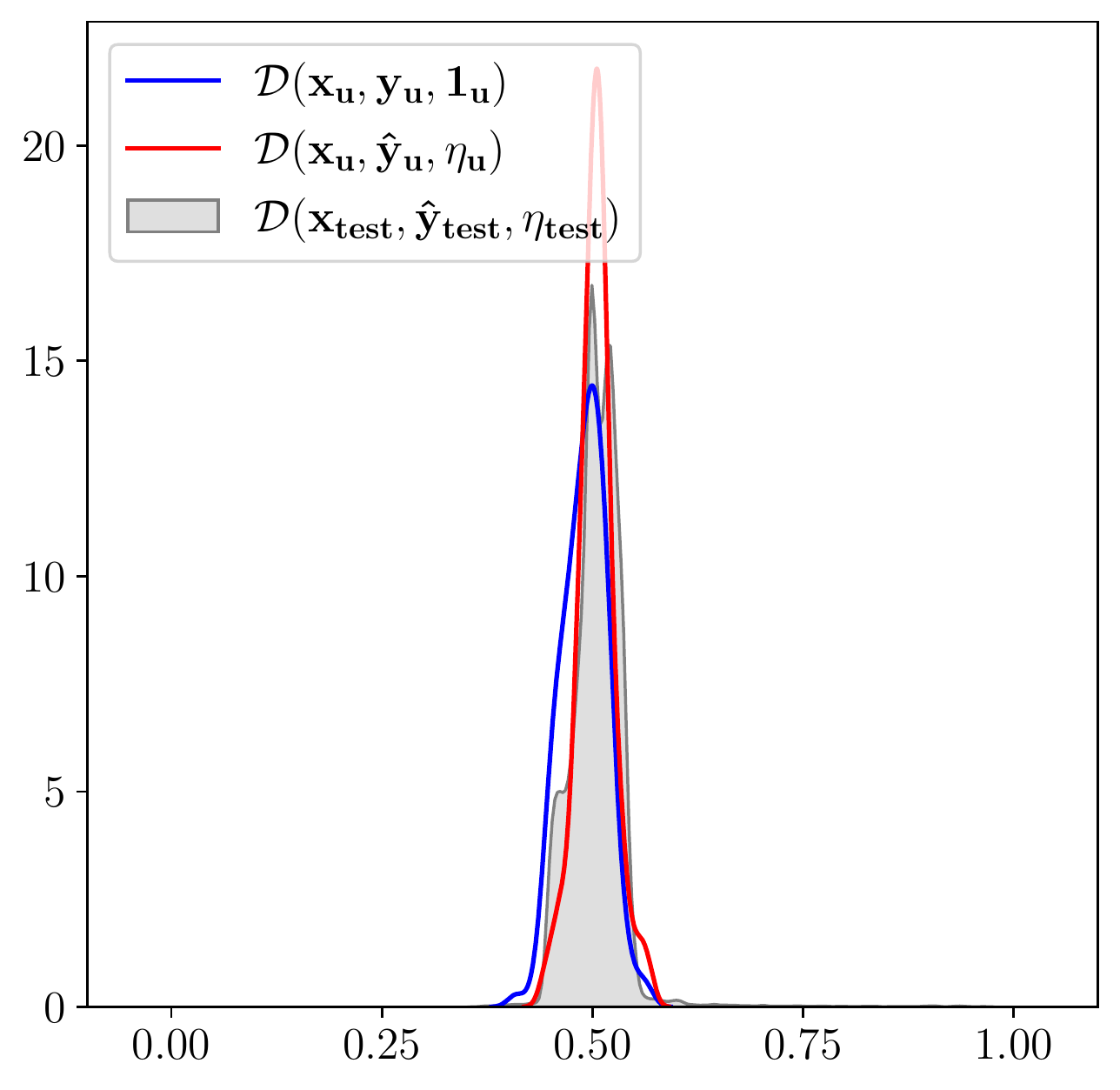}}

\subfigure[Schrodinger PIG-GAN]{\label{fig:Schrodinger_pig_logits} \includegraphics[scale=0.3]{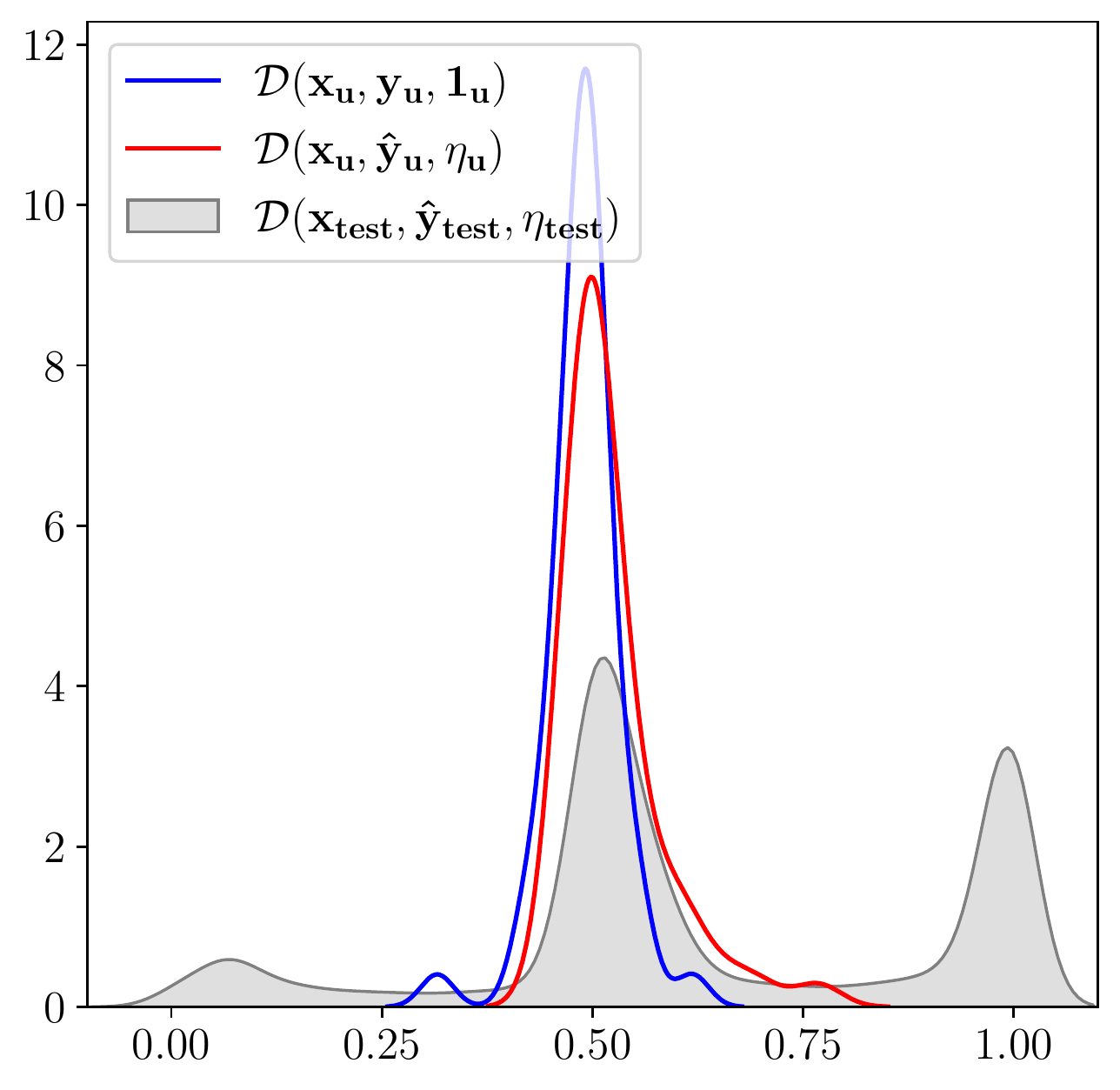}} 
\subfigure[Schrodinger PID-GAN]{\label{fig:Schrodinger_discriminator_logits} \includegraphics[scale=0.3]{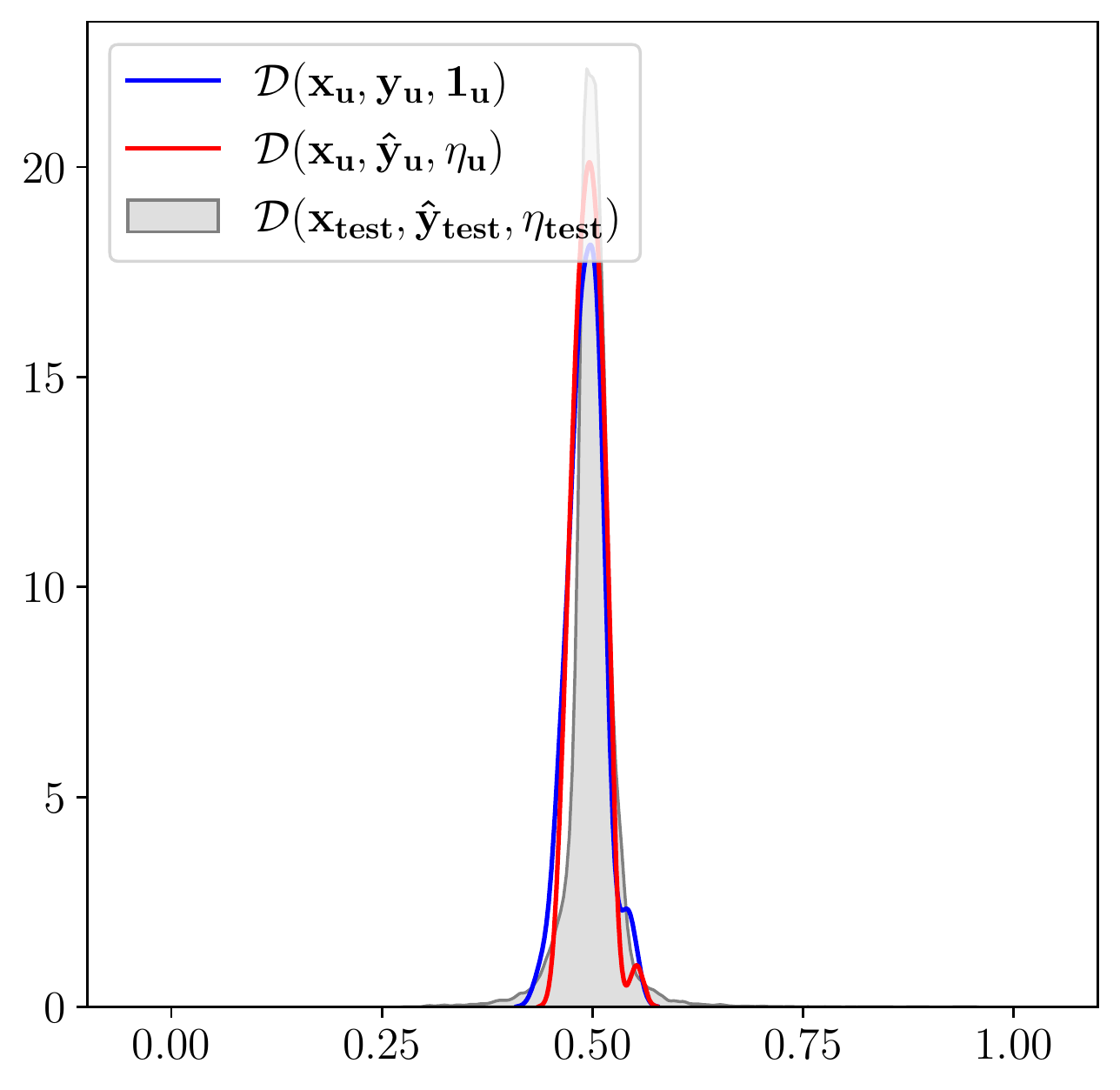}}

\vspace{-3ex}
\caption{Discriminator Scores of PIG-GAN and PID-GAN}
\label{fig:D_logits}
\vspace{-4ex}
\end{figure}

\par \noindent \textbf{Analyzing Discriminator Scores:}
Figure \ref{fig:D_logits} compares the discriminator outputs of PID-GAN and PIG-GAN on the labeled and unlabeled (test) points after training. For PIG-GAN, we evaluate the fully trained discriminator on three types of inputs: $(\mathbf{x_u, y_u})$, $(\mathbf{x_u, \hat{y}_u})$, and $(\mathbf{x_{test}, \hat{y}_{test}})$. (Note that we evaluate the performance of our model on $\mathbf{x_{test}}$ for PDEs, which is different from the unlabeled set $\mathbf{x_f}$ used during training.) On the other hand, for the PID-GAN, we evaluate the discriminator on three types of inputs: $(\mathbf{x_u, y_u, \textbf{1}_u})$, $(\mathbf{x_u, \hat{y}_u}, \boldsymbol{\eta}_{\mathbf{u}})$, and $(\mathbf{x_{test}, \hat{y}_{test}}, \boldsymbol{\eta}_{\mathbf{test}})$. 

On the Burgers' equation for PIG-GAN, we can observe that the discrminator is not able to distinguish between ``real'' samples $(\mathbf{x_u, y_u})$ and the generated ``fake'' samples $(\mathbf{x_u, \hat{y}_u})$ since the distribution of these two are similar and centered around 0.5. However, by analyzing its score on $(\mathbf{x_{test}, \hat{y}_{test}})$, we can see  that it always tends to predict the samples from the test set as ``fake'' by scoring them greater than 0.5 on average. This behavior is quite expected since the discriminator of PIG-GAN only learns the decision boundary between the ``real'' and ``fake'' samples on the labeled set. However, when we provide the discriminator of PID-GAN with similar inputs, we notice that it has adequately learned the decision boundary between the ``real'' and generated ``fake'' samples on both labeled and unlabeled set. Thus, the distribution of discriminator scores for $(\mathbf{x_{test}, \hat{y}_{test}}, \boldsymbol{\eta}_{\mathbf{test}})$ for PID-GAN is centered at 0.5. Again, for Schr\"odinger equation, we observe similar behavior for the discriminator scores of PIG-GAN and PID-GAN.  
% We can conclude that the discriminator of the PID-GAN is well trained on both labeled and unlabeled set and thus can be utilized for other purposes such as rejection sampling.

\par
\begin{figure*}[ht]
\centering
\subfigure[Exact]{\label{fig:B_Exact} \includegraphics[scale=0.22]{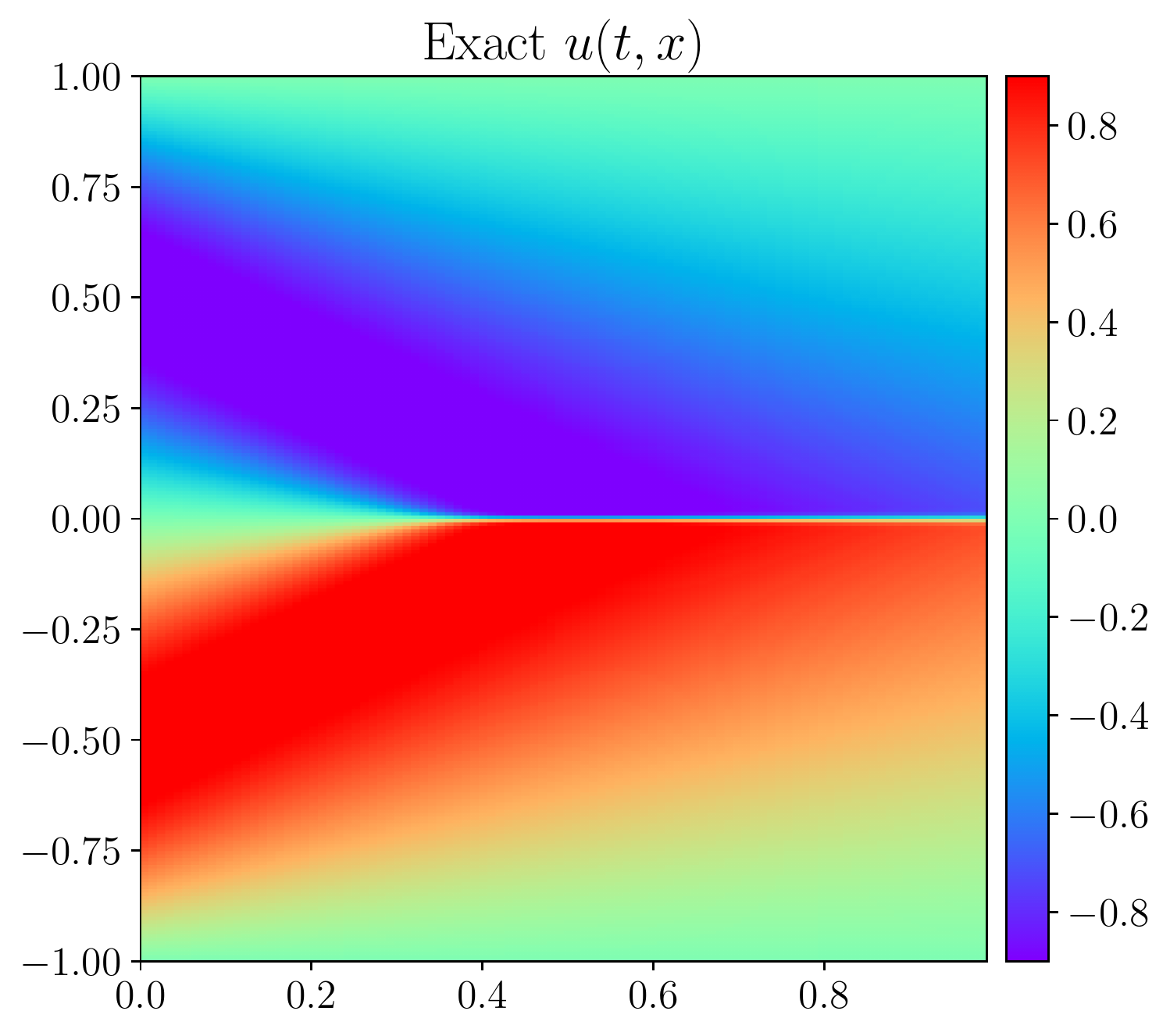}} 
\subfigure[PIG-GAN Absolute Error]{\label{fig:B_err_pig} \includegraphics[scale=0.22]{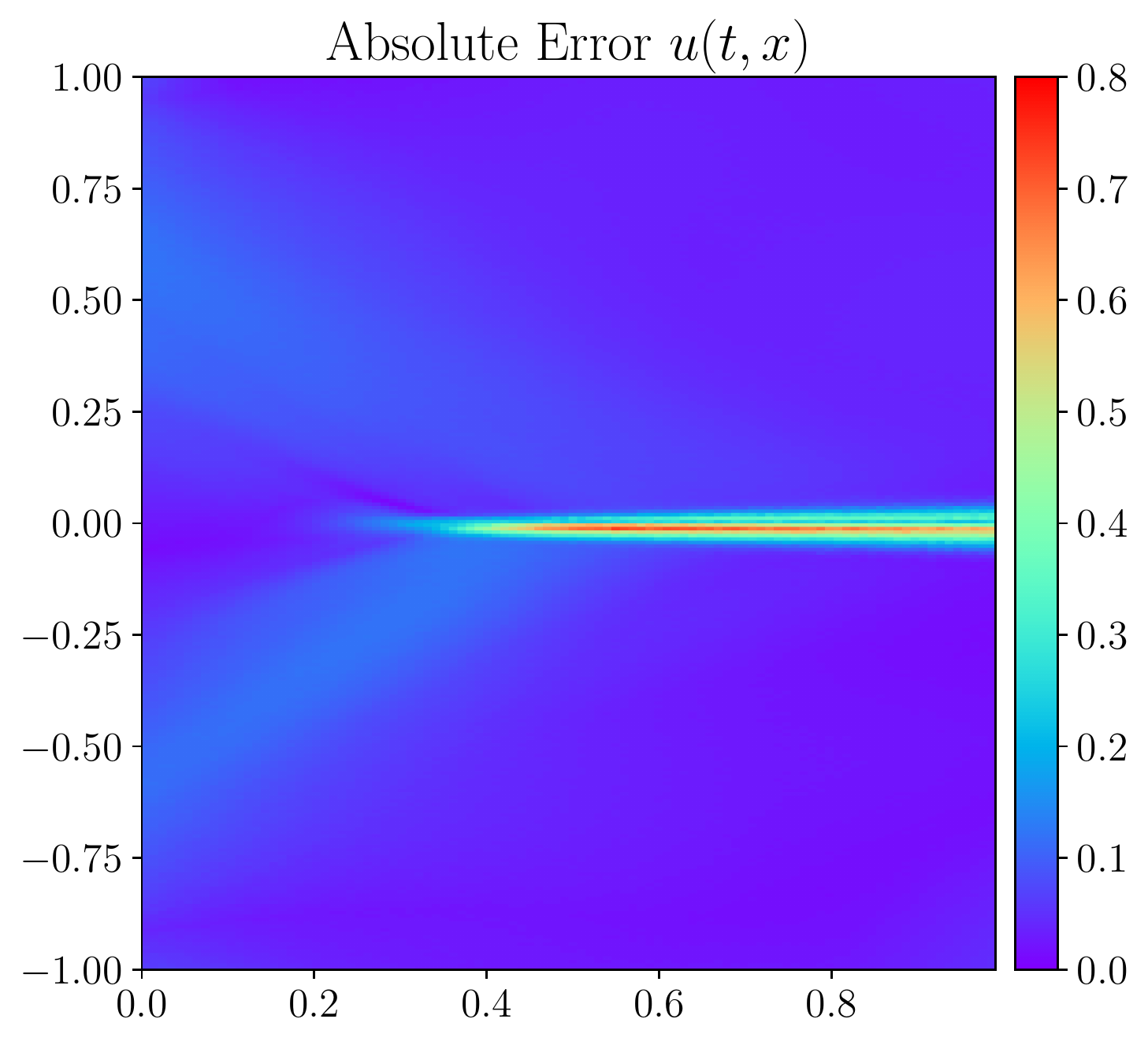}}
\subfigure[PID-GAN Absolute Error]{\label{fig:B_err_pid} \includegraphics[scale=0.22]{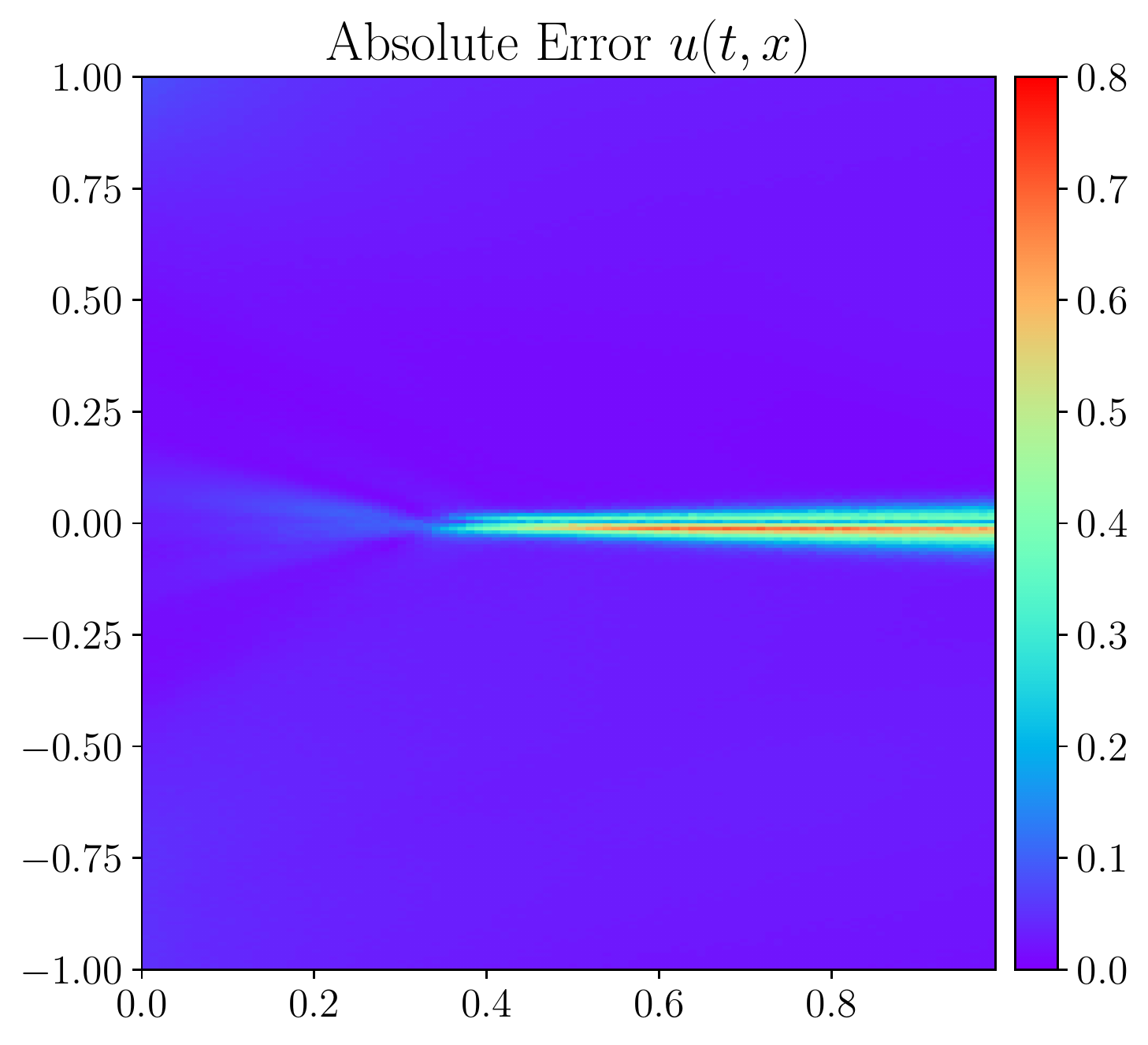}}
\subfigure[PIG-GAN Variance]{\label{fig:B_var_pig} \includegraphics[scale=0.22]{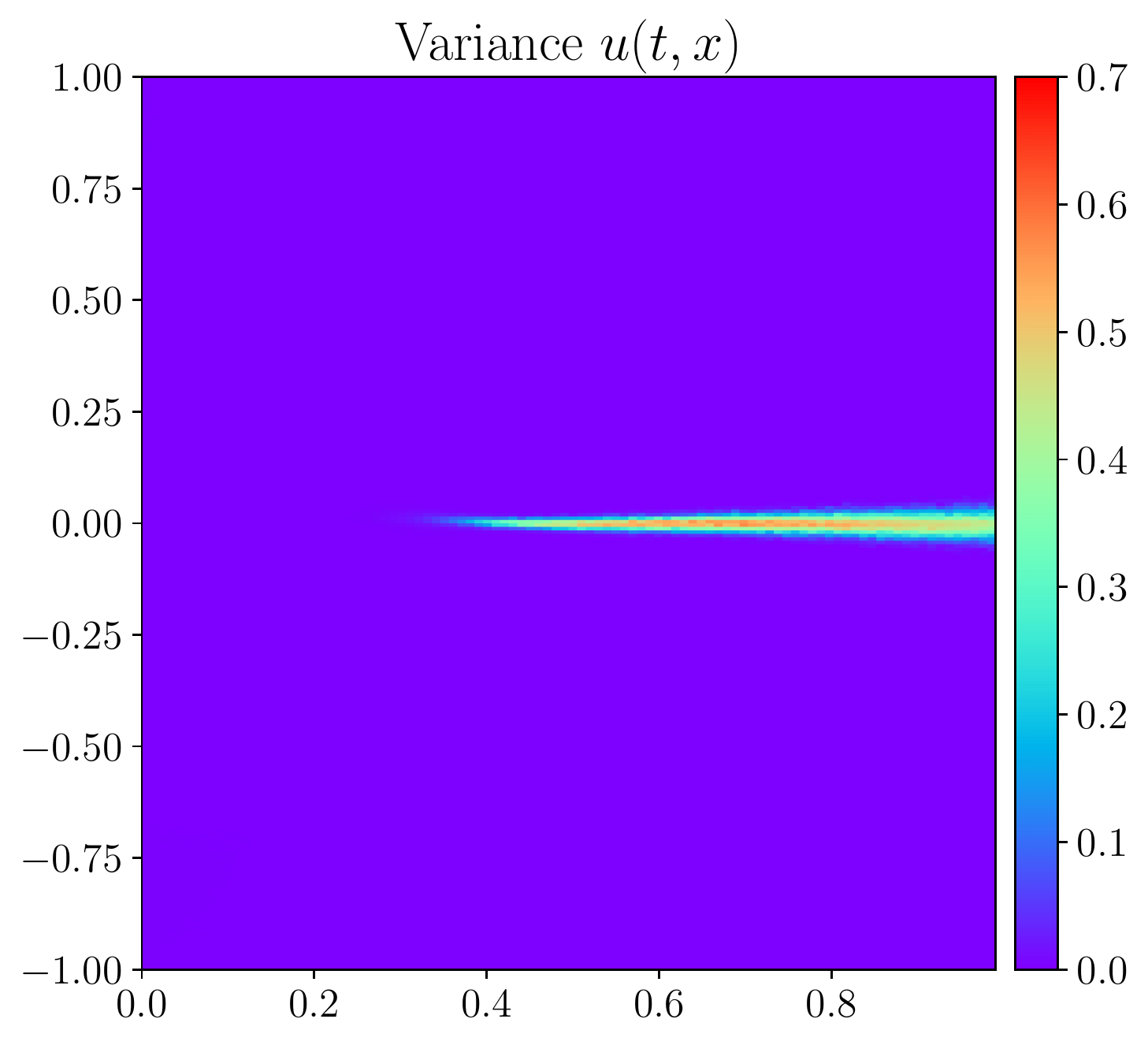}}
\subfigure[PID-GAN Variance]{\label{fig:B_var_pid} \includegraphics[scale=0.22]{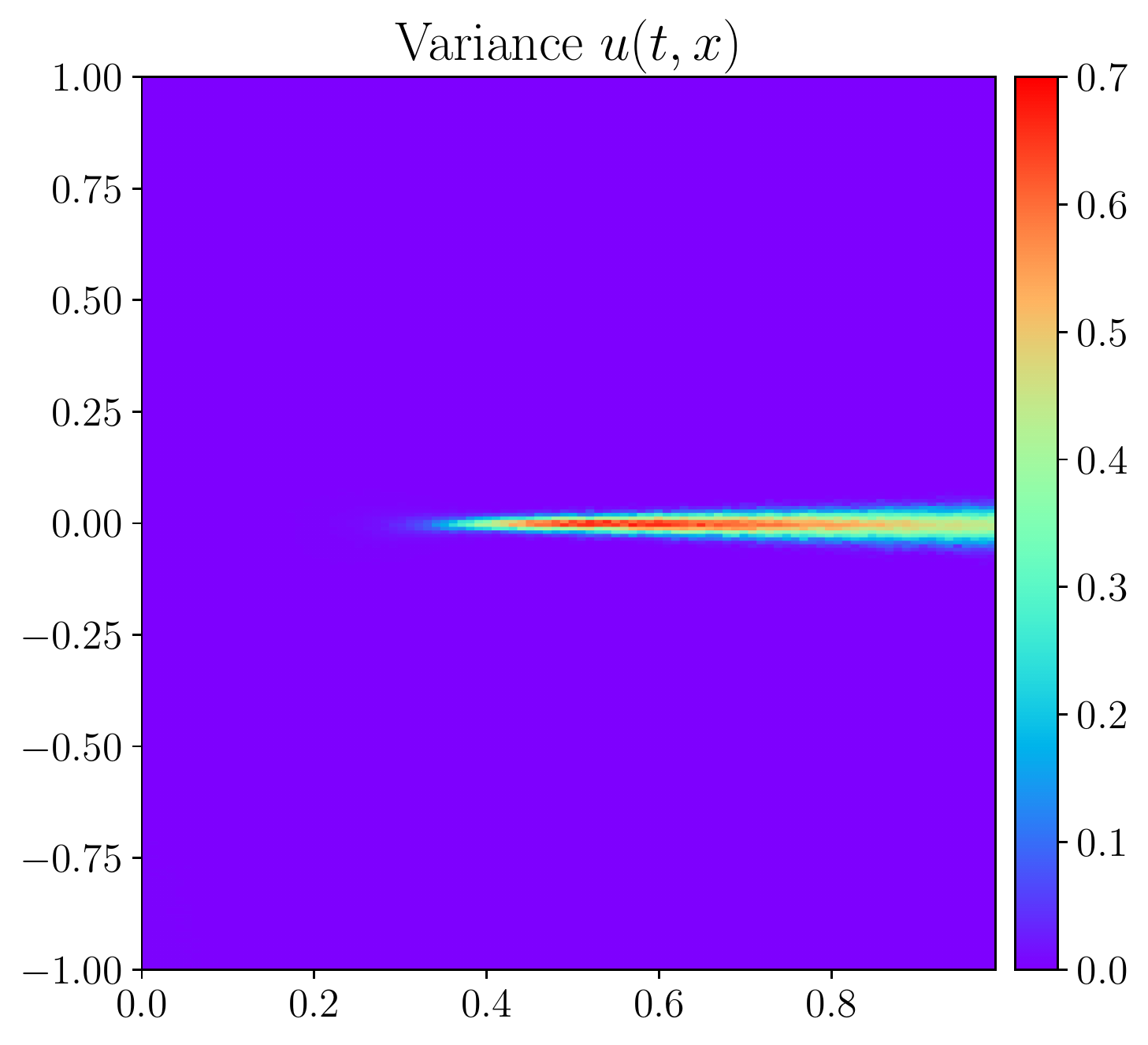}}
\vspace{-3ex}
\caption{Visualization of Absolute Error and Variance for Burger's Equation.}
\label{fig:B_viz}
\end{figure*}

\begin{figure*}[ht]
\centering
\subfigure[Exact]{\label{fig:S_Exact} \includegraphics[scale=0.22]{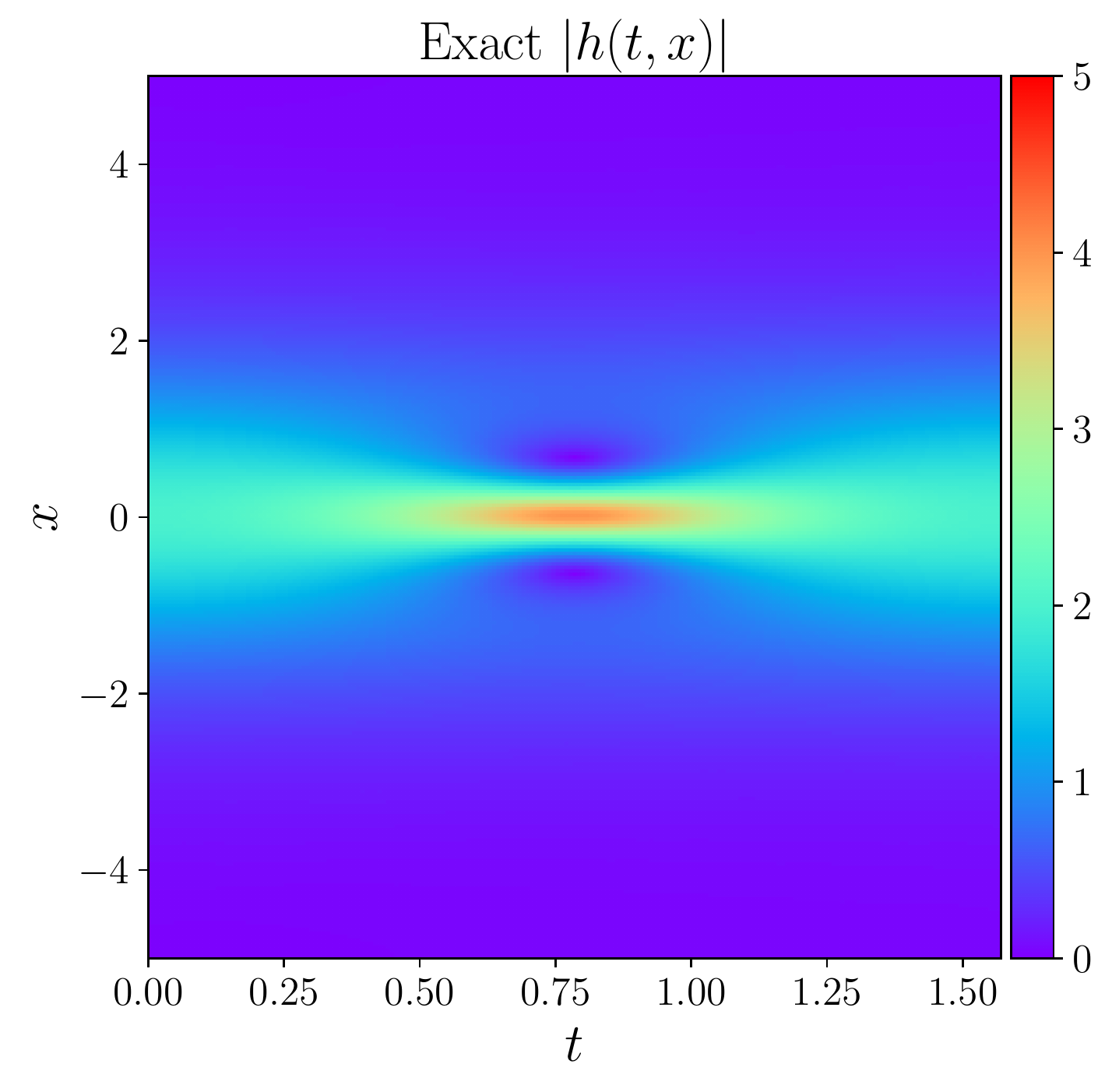}} 
\subfigure[PIG-GAN Absolute Error]{\label{fig:S_err_pig} \includegraphics[scale=0.22]{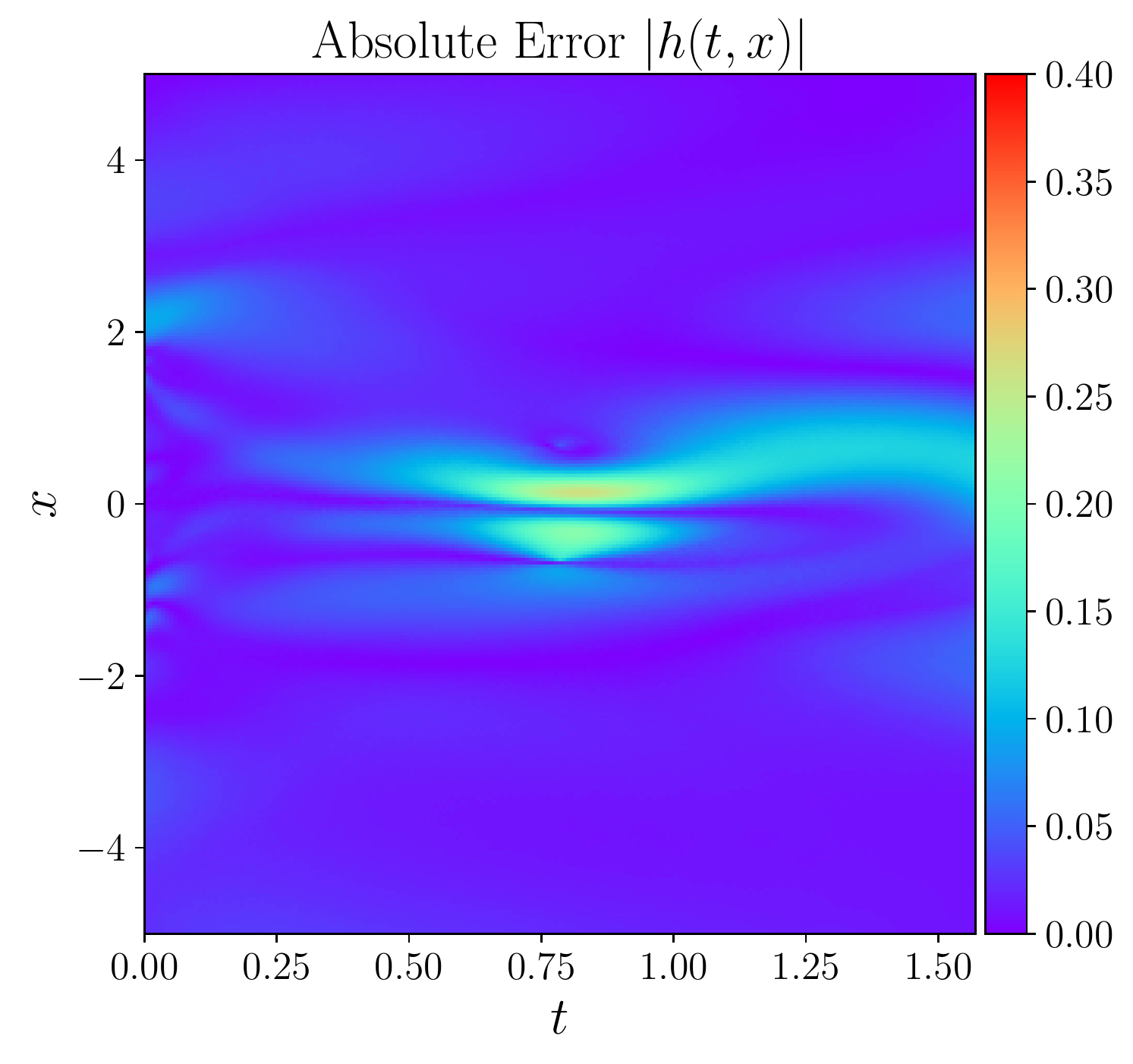}}
\subfigure[PID-GAN Absolute Error]{\label{fig:S_err_pid} \includegraphics[scale=0.22]{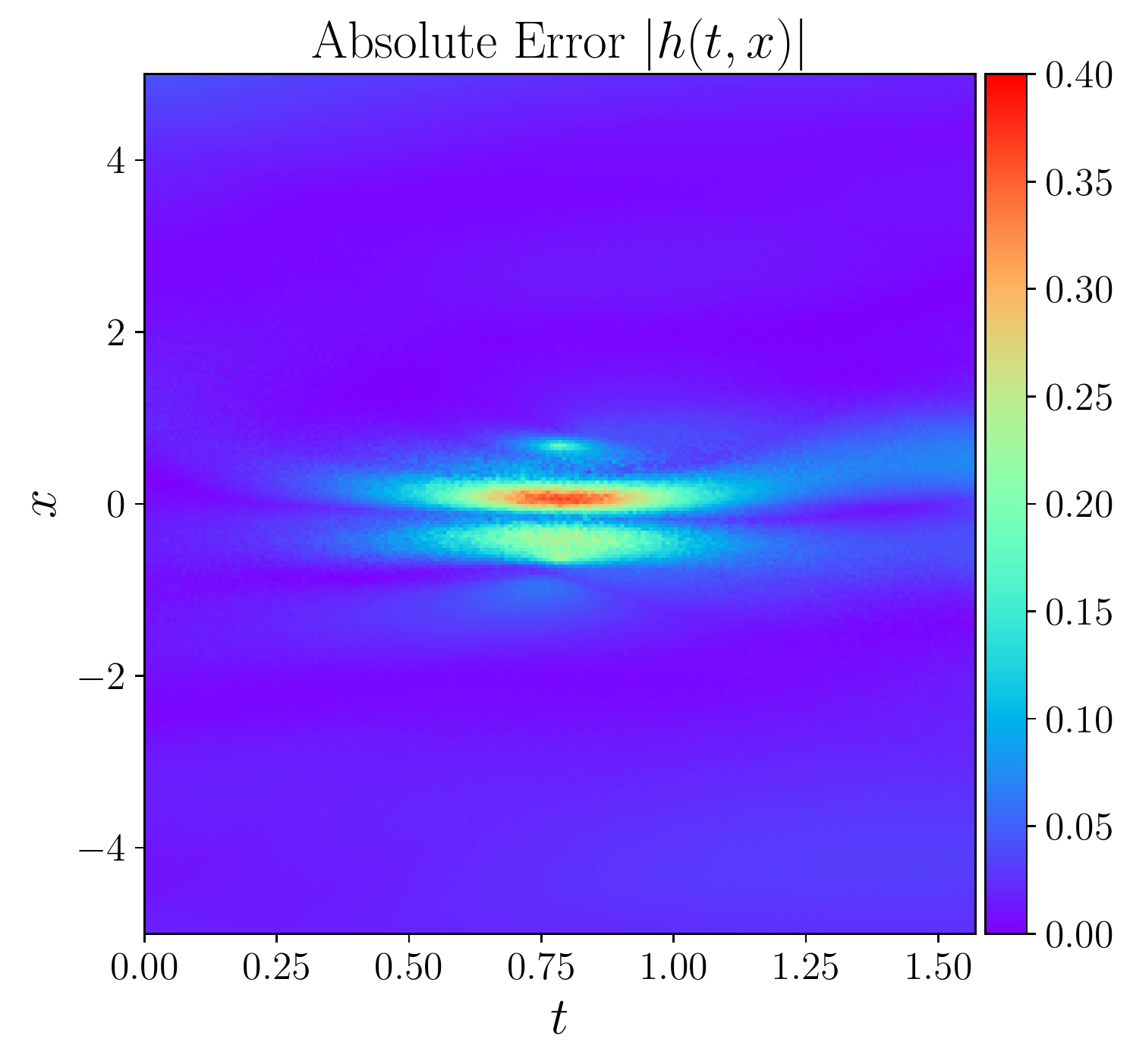}}
\subfigure[PIG-GAN Variance]{\label{fig:S_var_pig} \includegraphics[scale=0.22]{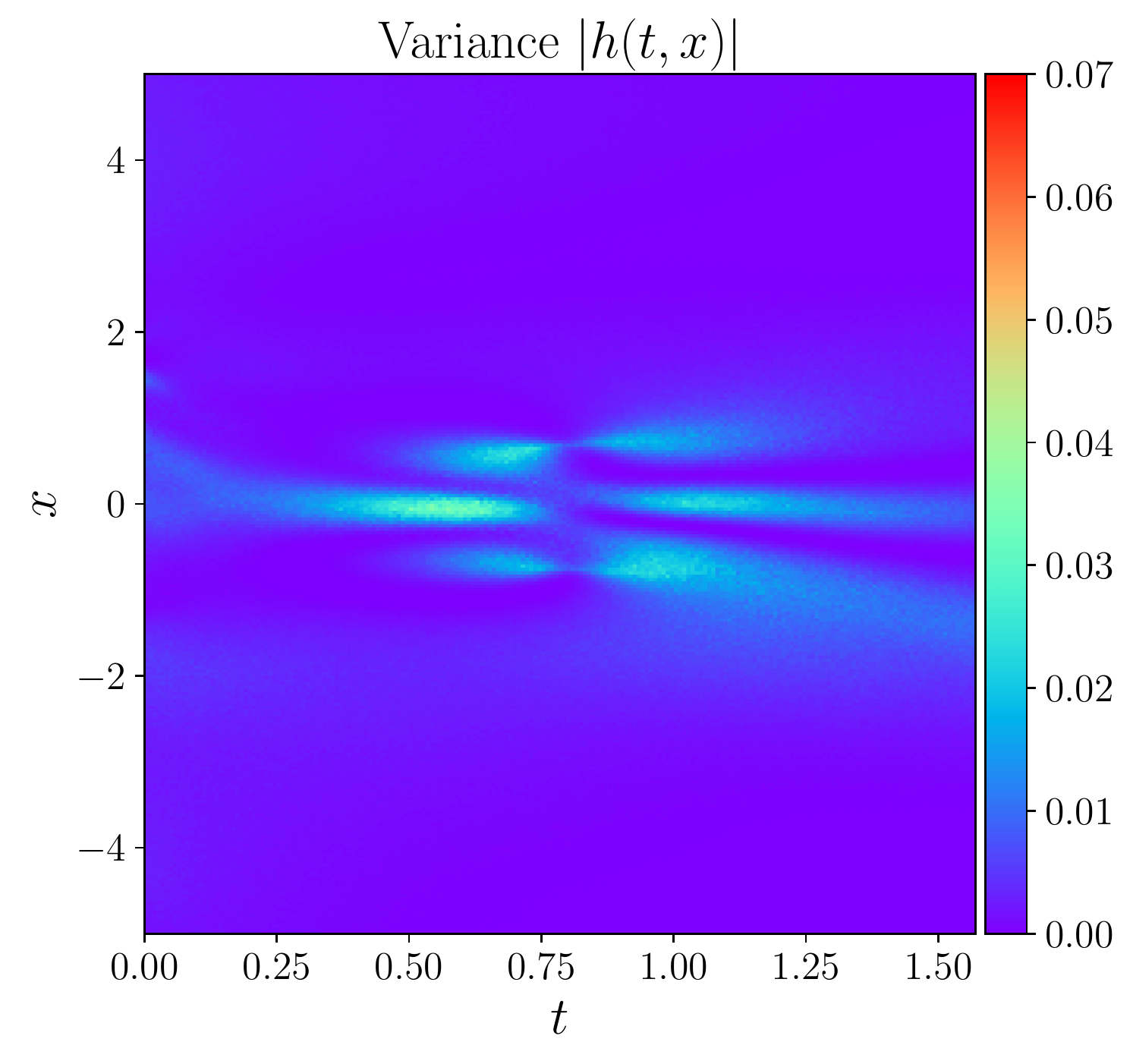}}
\subfigure[PID-GAN Variance]{\label{fig:S_var_pid} \includegraphics[scale=0.22]{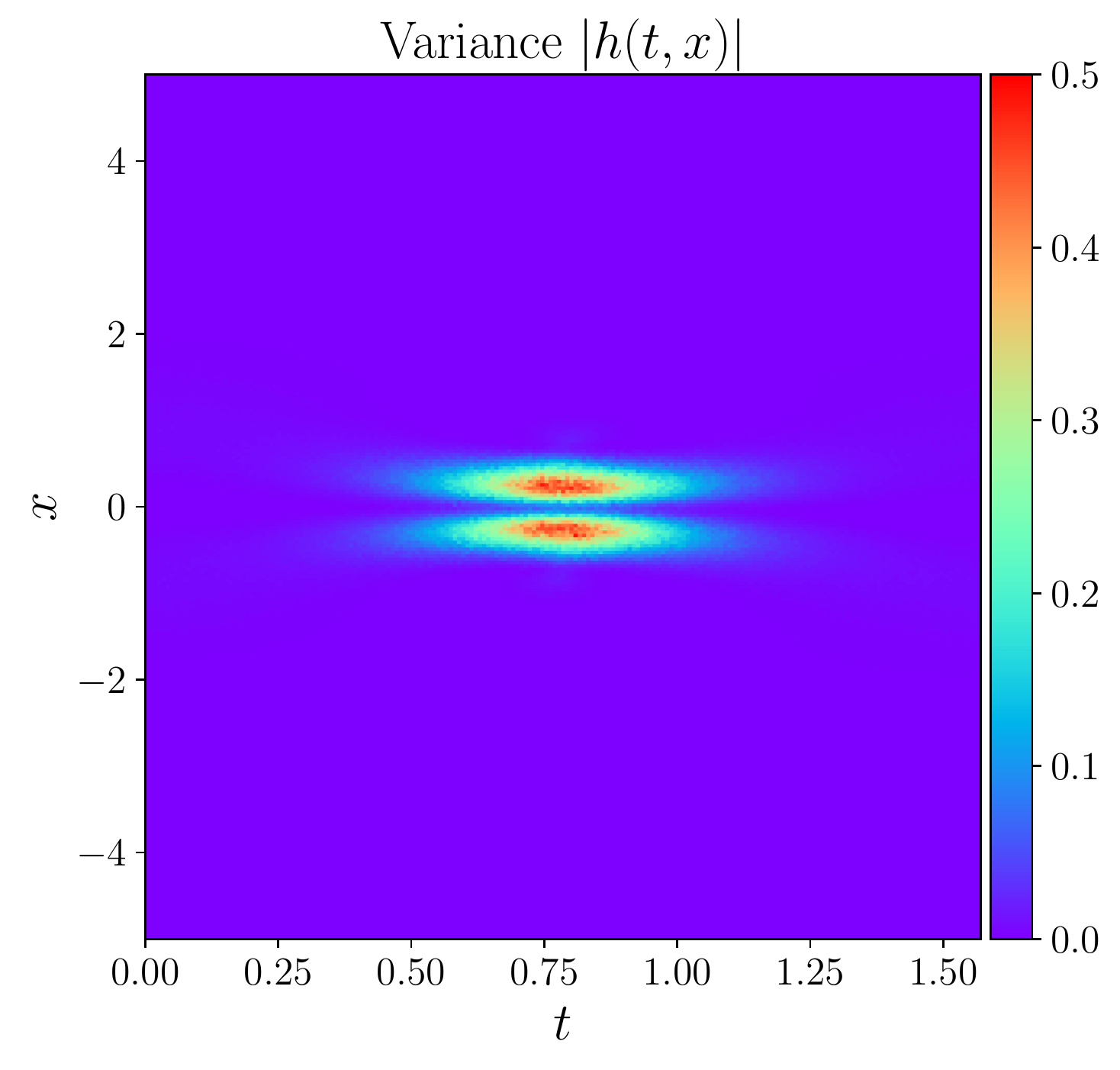}}
\vspace{-3ex}
\caption{Visualization of Absolute Error and Variance for Schr\"odinger Equation.}
\label{fig:S_viz}
\end{figure*}

 \noindent \textbf{Visualization of PDE Solutions:}
Figure \ref{fig:B_viz} shows the exact solution of the Burgers' equation along with the absolute errors and variances of PIG-GAN and PID-GAN. Burgers' equation has a nonlinear steep response at x=0, where both the predictive models show a strong correlation between regions with higher variances and higher absolute errors, which is a desirable behavior. 
For example, if the variance of a model is high in a certain region, it suggests that the model is less confident in its predictions and thus can incur larger errors. On the contrary, a model with low variance and a higher value of errors indicates poor and over-confident predictions. Similar to Burgers', Figure \ref{fig:S_viz} shows the exact solutions of the Schr\"{o}dinger equation along with the absolute errors and variances of PID-GAN and PIG-GAN. Again, we observe two steep responses centered around x = -1 and 1 for t=0.79. Similar to the Burgers', we see PID-GAN has higher variance with a higher absolute error at the steep region. However, the absolute errors and the variances of PIG-GAN are not only concentrated over the steep region but also spread all over the spatio-temporal domain. This means that PIG-GAN is both less confident and more prone to errors in its prediction even in regions with smooth responses. 
Additional analyses of the results comparing the predicted and exact solutions of Burgers' and Schr\"{o}dinger equation have been provided in Appendix \ref{sec-additionalresult}.

% \begin{figure*}[ht]
% \centering
% \subfigure[Exact u]{\label{fig:D_u_Exact} \includegraphics[scale=0.25]{figures/Darcy/Exact_u.jpg}} 
% \subfigure[PIG-GAN Absolute Error u]{\label{fig:D_u_err_pig} \includegraphics[scale=0.25]{figures/Darcy/Absolute_error_u_pig.jpg}}
% \subfigure[PID-GAN Absolute Error u]{\label{fig:D_u_err_pid} \includegraphics[scale=0.25]{figures/Darcy/Absolute_error_u_pid.jpg}}
% \subfigure[PIG-GAN Variance u]{\label{fig:D_u_var_pig} \includegraphics[scale=0.25]{figures/Darcy/u_var_pig.jpg}}
% \subfigure[PID-GAN Variance u]{\label{fig:D_u_var_pid} \includegraphics[scale=0.25]{figures/Darcy/u_var_pid.jpg}}

% \subfigure[Exact k]{\label{fig:D_k_Exact} \includegraphics[scale=0.25]{figures/Darcy/Exact_k.jpg}} 
% \subfigure[PIG-GAN Absolute Error k]{\label{fig:D_k_err_pig} \includegraphics[scale=0.25]{figures/Darcy/Absolute_error_k_pig.jpg}}
% \subfigure[PID-GAN Absolute Error k]{\label{fig:D_k_err_pid} \includegraphics[scale=0.25]{figures/Darcy/Absolute_error_k_pid.jpg}}
% \subfigure[PIG-GAN Variance k]{\label{fig:D_k_var_pig} \includegraphics[scale=0.25]{figures/Darcy/k_var_pig.jpg}}
% \subfigure[PID-GAN Variance k]{\label{fig:D_k_var_pid} \includegraphics[scale=0.25]{figures/Darcy/k_var_pid.jpg}}
% \vspace{-3ex}
% \caption{Darcy Visualization}
% \label{fig:D_viz}
% \end{figure*}

\section{Conclusions and Future Work}
We presented a GAN based framework for UQ with physics-supervision, referred to as PID-GAN.  
A unique aspect of PID-GAN is its ability to exploit the full potential of adversarial optimization in-built in GAN frameworks for training both the discriminator and generator on labeled and unlabeled points. Future directions of research can explore the use of PID-GANs for inference tasks such as rejection sampling.

\section{Acknowledgement}
This work was supported by NSF grant $\#2026710$.

\bibliographystyle{ACM-Reference-Format}
\bibliography{sample-base}

\appendix
\appendixpage
\addappheadtotoc
% \newpage

\section{Implementation Details}
\label{sec-implementation}
In this section, we describe the model, data, and training details for reproducing the experiments.

% \subsection*{Baseline models}
% We evaluate the performance of our proposed PID-GAN with its GAN-based counterpart PIG-GAN, with conventional conditional GAN (cGAN), and with physics-informed neural network-based PINN-Drop and APINN-Drop. In section 2, we discuss these models in detail.

\subsection{Procedure for selecting hyparparameters}
For the baselines, we choose the exact hyperparameters and model architectures as described in their respective papers. 

\par \noindent \textbf{Architecture selection:}
In the GAN-based formulation, the discriminator is much easier to train than the generator. To infer the PDE solutions, for every discriminator update, the generator is updated 5 times for all of our GAN-based models. Since we are providing additional information (physics consistency score) to the discriminator, it can easily overfit, which further leads to the vanishing gradient problem.  To address this problem, we choose a simpler architecture for the discriminator in the PID-GAN model when compared to its PIG counterpart. 

\par \noindent \textbf{Procedure for selecting $\lambda$:}
We tune the trade-off hyperparameter $\lambda$ using random search, where we set a heuristic to select the initial $\lambda = 1/ (\mathcal{R}_{\theta_0})$ where $\theta_0$ are the generator weights after random initialization. For the baselines shown in PDE problems, we choose the exact trade-off $\lambda$ as reported in the respective papers, while for other datasets, we perform a random hyperparameter search to select the optimal $\lambda$.

We use the Adam optimizer with a learning rate of $10^{-4}$ for PDE problems and $10^{-3}$ for other datasets. 

\subsection{Experimental Setup}
% \subsubsection*{Synthetic Dataset}
% For the synthetic dataset, we randomly select 30 labeled points, which are uniformly distributed across the data. We further optimize our models using 270 unlabeled points.
In this section, we provide additional information regarding each of the datasets used in the main text. For each of our experiments, we normalized the inputs to follow zero mean and unit standard deviations. 
\par \noindent \textbf{Real-world Dataset:}
To evaluate our model performance on imperfect physics, we use two real-world datasets: collision prediction dataset, and tossing trajectory prediction dataset. 
We select 108 labeled points and 436 unlabeled points to train our models on the collision dataset. Meanwhile, we use 217 labeled points with 327 unlabeled points to train on the tossing dataset. All of the comparative baselines are trained for 5,000 epochs and 10,000 epochs on collision and tossing dataset respectively. 

\par \noindent \textbf{Partial Differential Eqations (PDE):}

\par \noindent \textbf{Burgers Equation:}
Burgers' equation is a fundamental partial differential equation that has multiple applications in areas ranging from applied mathematics like fluid mechanics, nonlinear acoustics, to traffic flow problems. For Burgers' equation, we train our model on 150 labeled points, which are uniformly distributed across the initial condition data $\{(x,t)| t = 0\}$ and the boundary points $\{(x, t)| x \in \{-1, 1\}\}$. We further use 10,000 randomly chosen unlabeled points (collocation points for PDEs) to optimize our models. We train the baseline models for 30,000 epochs, which is common practice in the existing literature. 

\par \noindent \textbf{Schr\"{o}dinger Equation:}
We use 100 labeled points for Schrodinger equation, where we select 50 uniformly distributed data points across the initial condition $\{(x,t)| t = 0\}$ and 50 uniformly distributed data points across the boundary points $\{(x, t)| x \in \{5, -5\}\}$. Moreover, we select 20,000 unlabeled points (collocation points for PDEs) that are randomly chosen from the input space using the Latin Hypercube Sampling strategy. Similar to the previous works, we train our baseline models for 50,000 epochs.

\par \noindent \textbf{Darcy's Equation:}
For Darcy's equation, we use 200 scattered labeled points uniformly distributed over the data space. Additionally, we use 400 boundary points uniformly distributed over the four boundaries. Moreover, we select 10,000 unlabeled points (collocation points for PDEs) that we randomly choose from the input space. All of the baselines are trained over 30,000 epochs.

\begin{figure*}[ht]
\centering
\subfigure[Exact k]{
\label{fig:D_u_Exact} \includegraphics[scale=0.25]{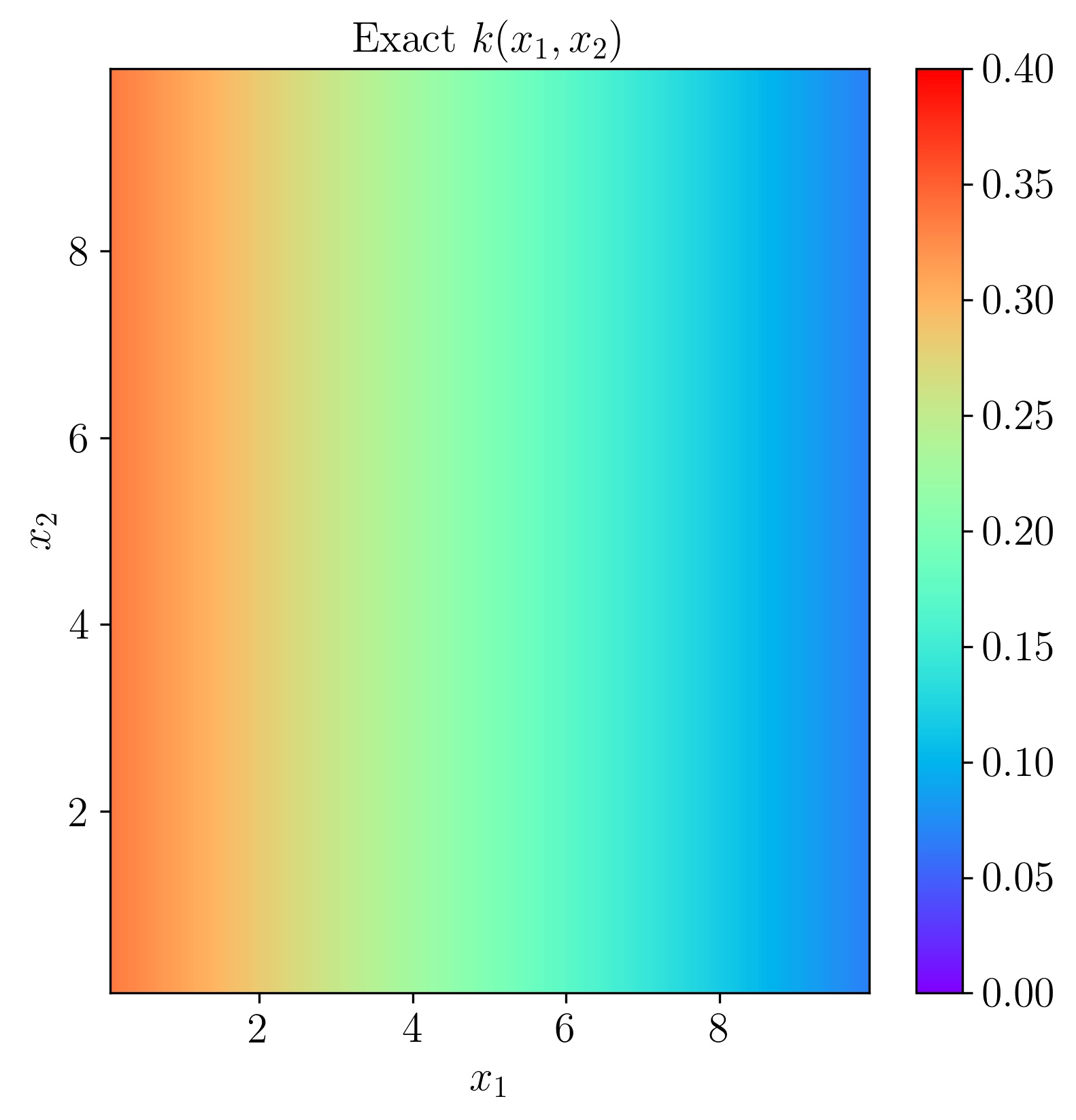}} 
\subfigure[PID-GAN Absolute Error k]{
\label{fig:D_u_err_pig} \includegraphics[scale=0.25]{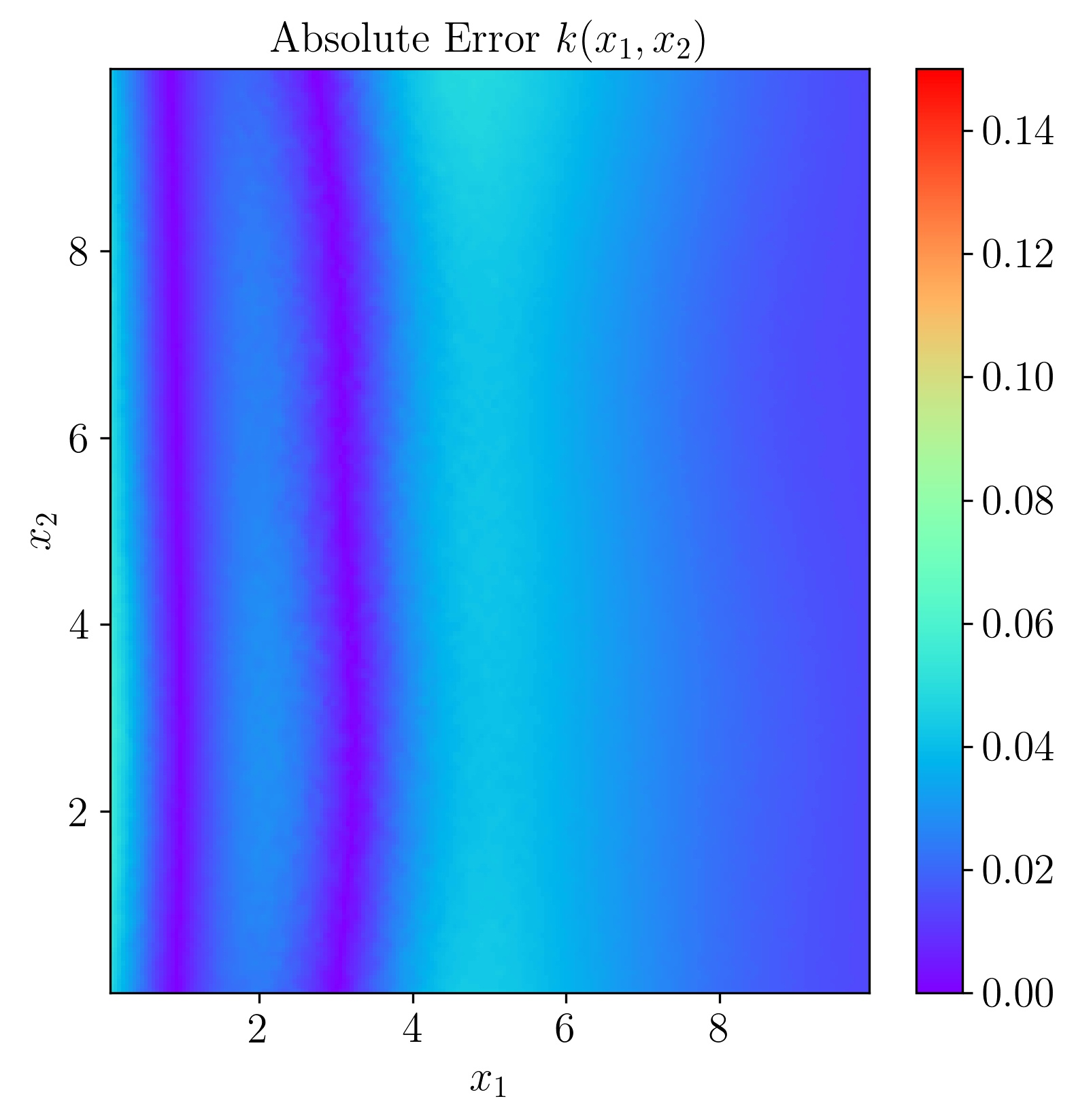}}
\subfigure[APINN-Drop Absolute Error k]{
\label{fig:D_u_err_pid} \includegraphics[scale=0.25]{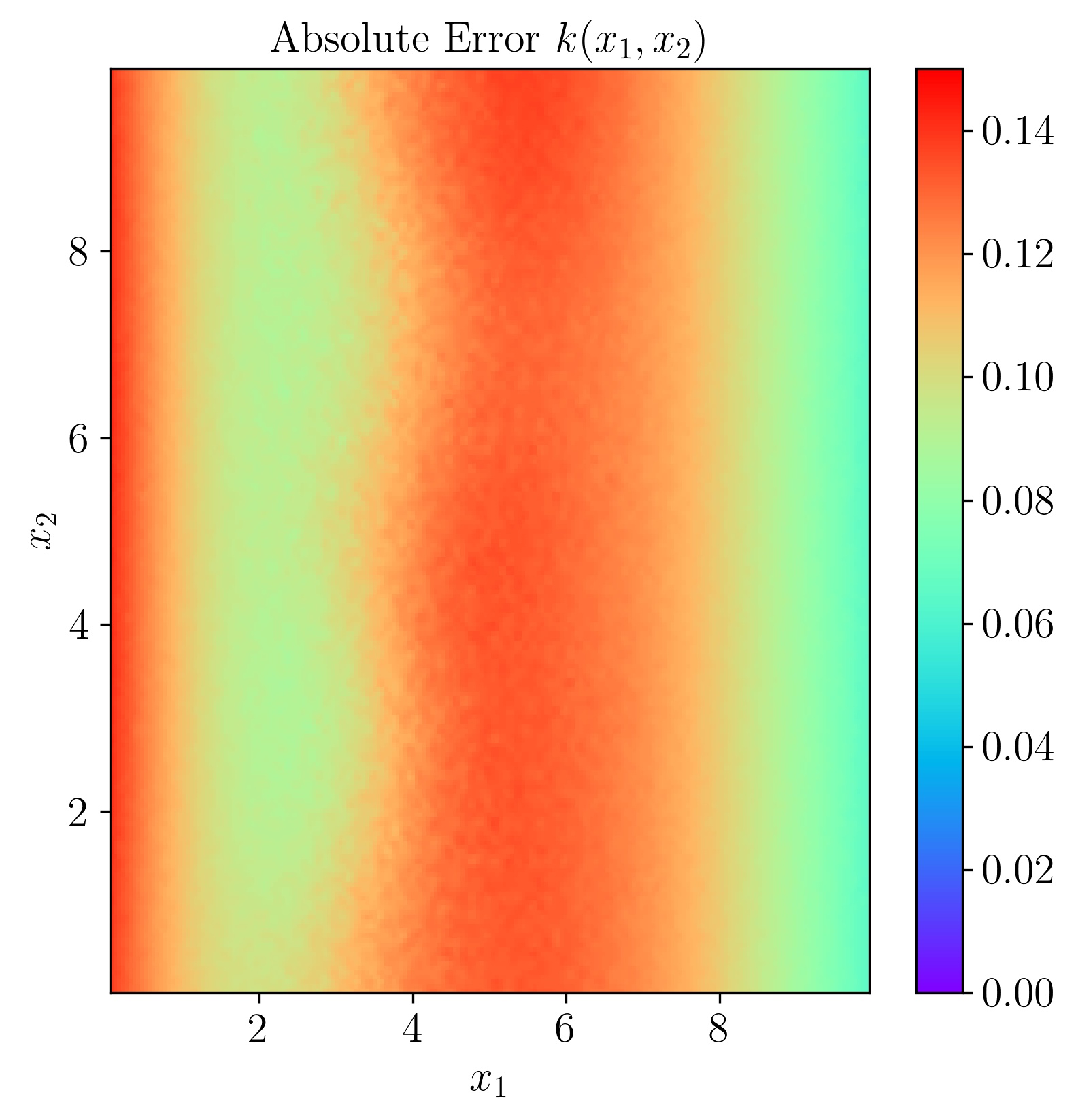}}
\subfigure[PID-GAN Variance k]{
\label{fig:D_u_var_pig} \includegraphics[scale=0.25]{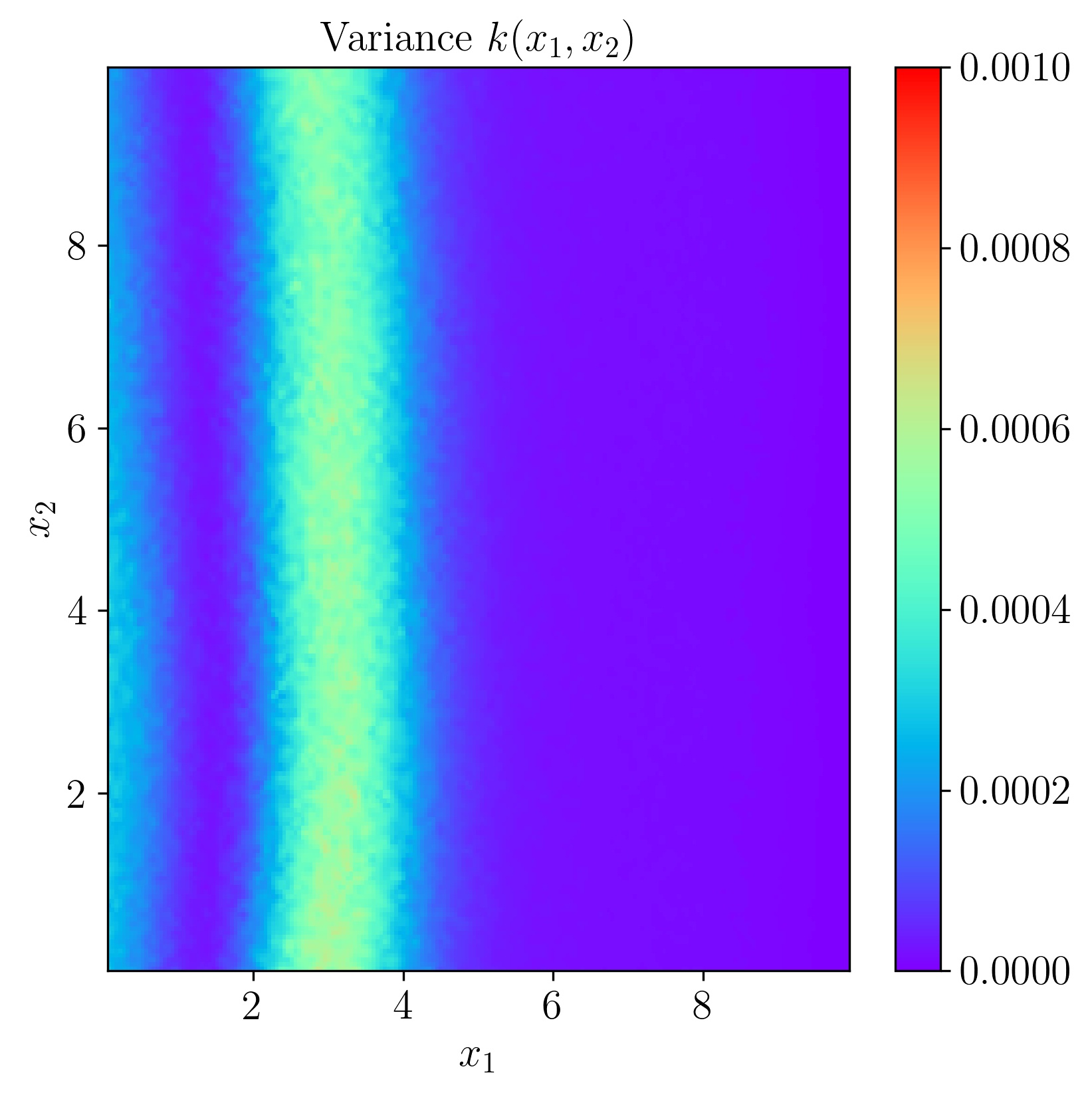}}
\subfigure[APINN-Drop Variance k]{
\label{fig:D_u_var_pid} \includegraphics[scale=0.25]{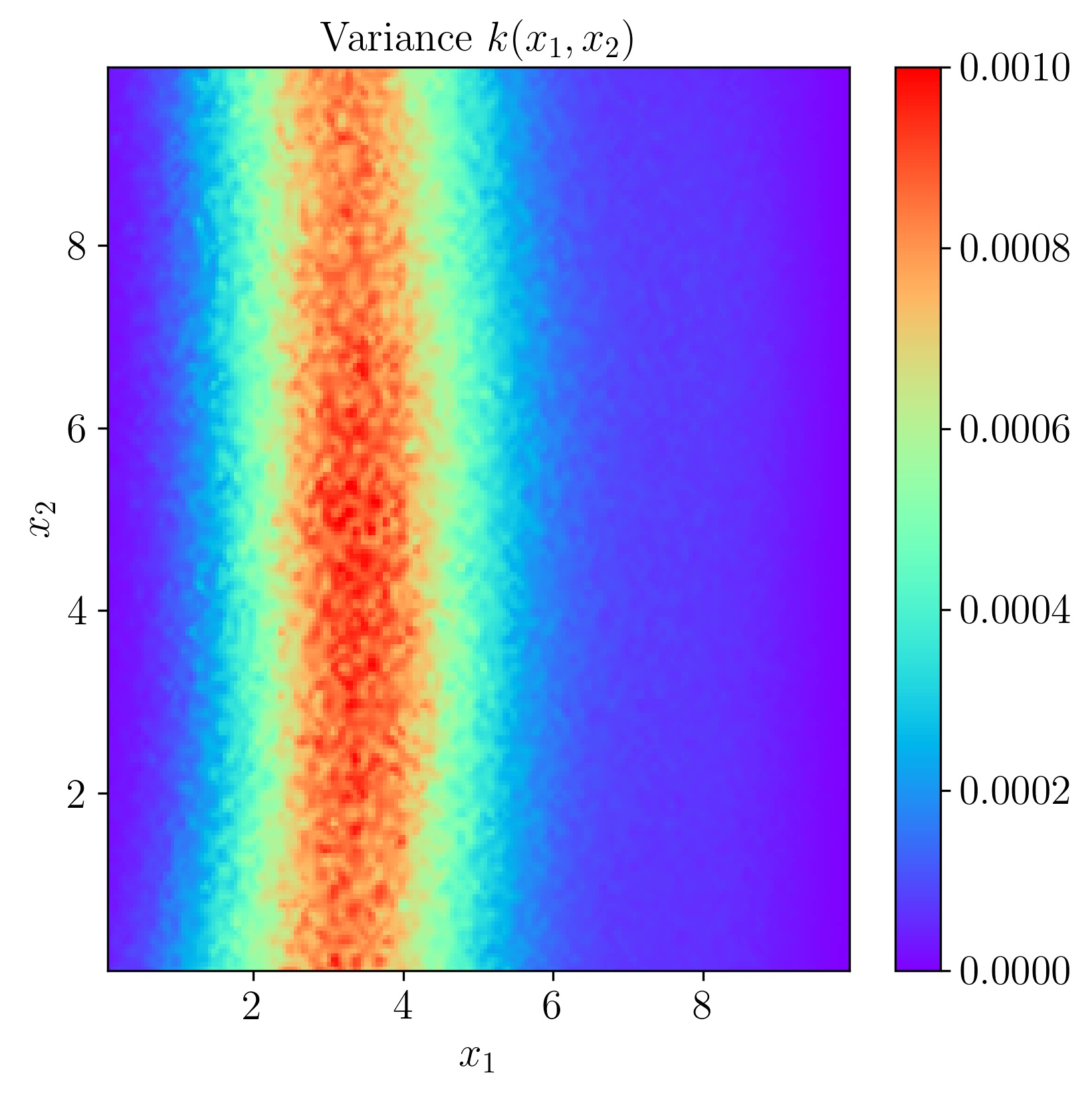}}
\vspace{-3ex}
\caption{Comparison of PID-GAN and Adaptive PINN-Drop in terms of absolute error and variance on Darcy's equation.}
\label{fig:D_viz}
\end{figure*}

\begin{figure*}[ht]
\centering
\subfigure[PID-GAN]{
\label{fig:pid_burgers} \includegraphics[scale=0.20]{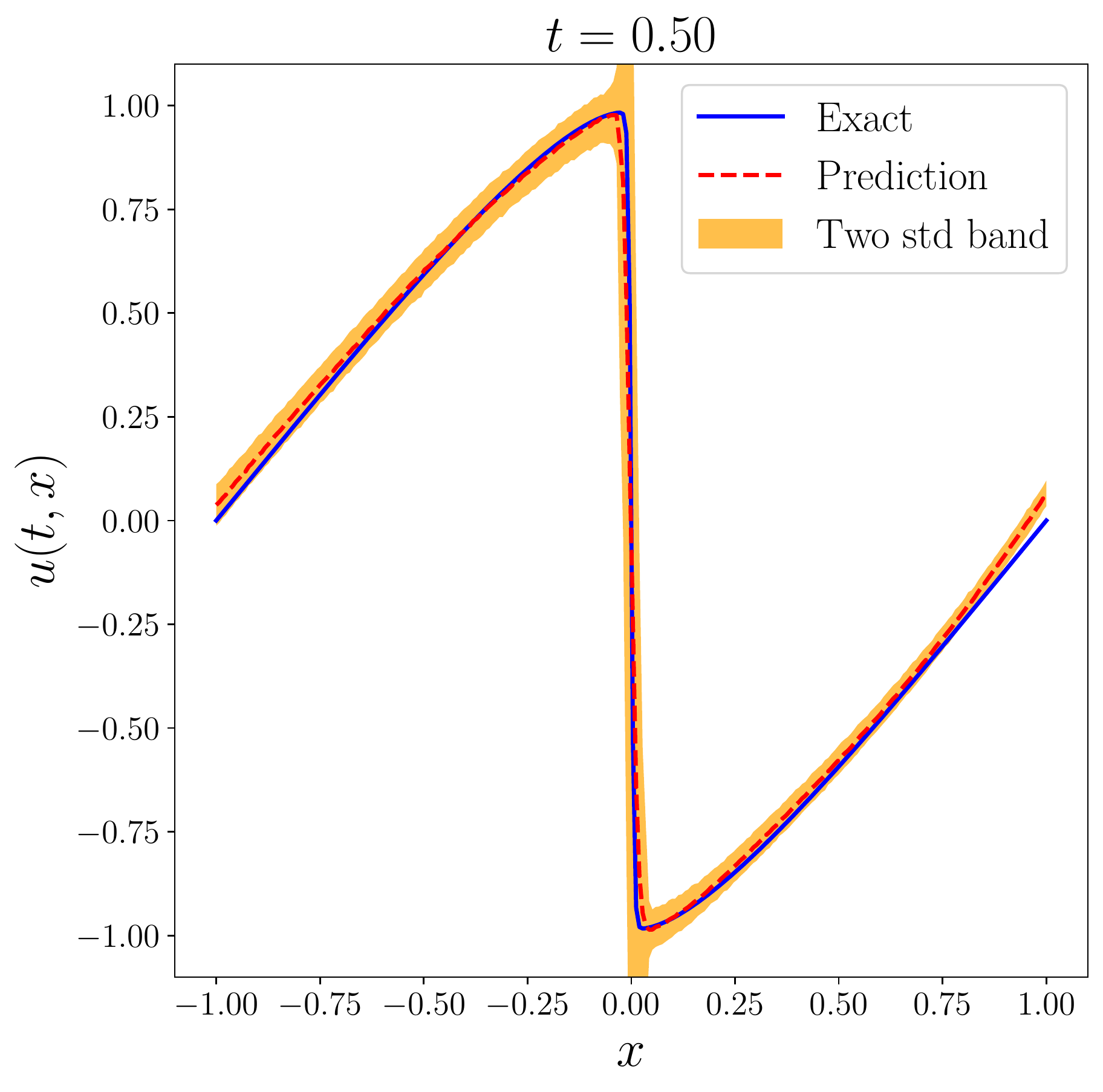}} 
\subfigure[PIG-GAN]{
\label{fig:pig_burgers} \includegraphics[scale=0.20]{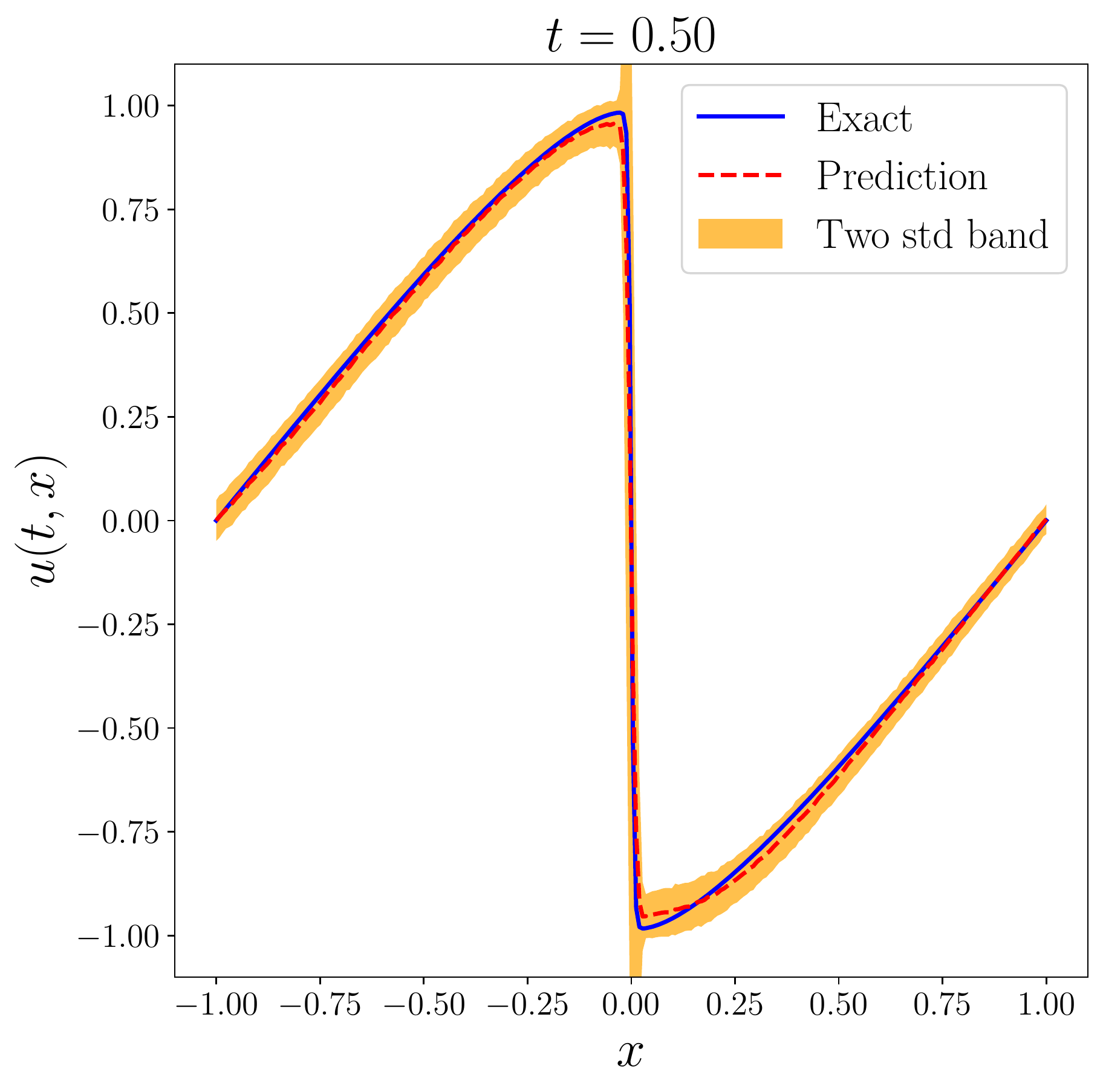}}
\subfigure[PINN-Drop]{
\label{fig:pinn_burgers} \includegraphics[scale=0.20]{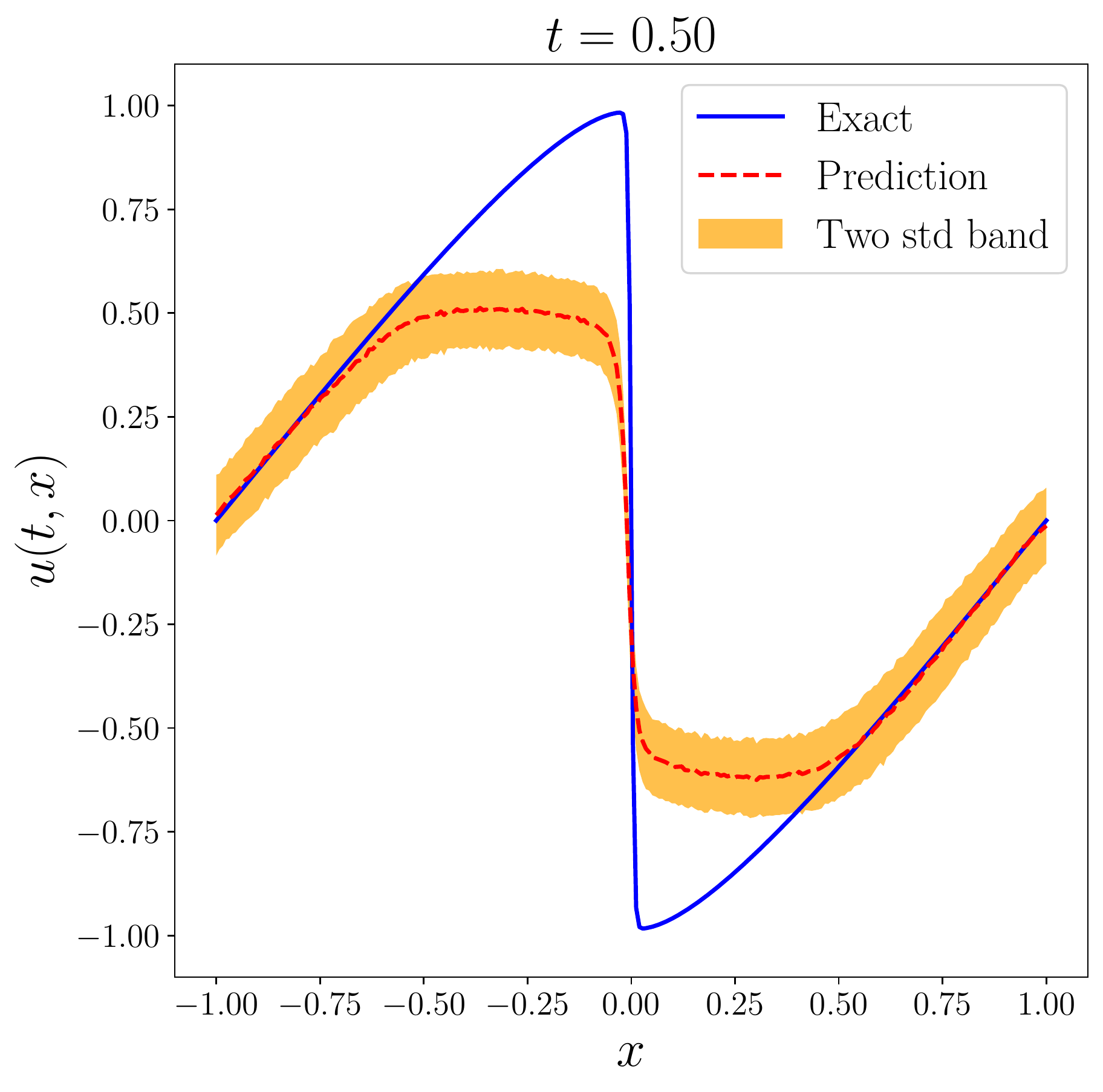}}
\subfigure[APINN-Drop]{
\label{fig:apinn_burgers} \includegraphics[scale=0.20]{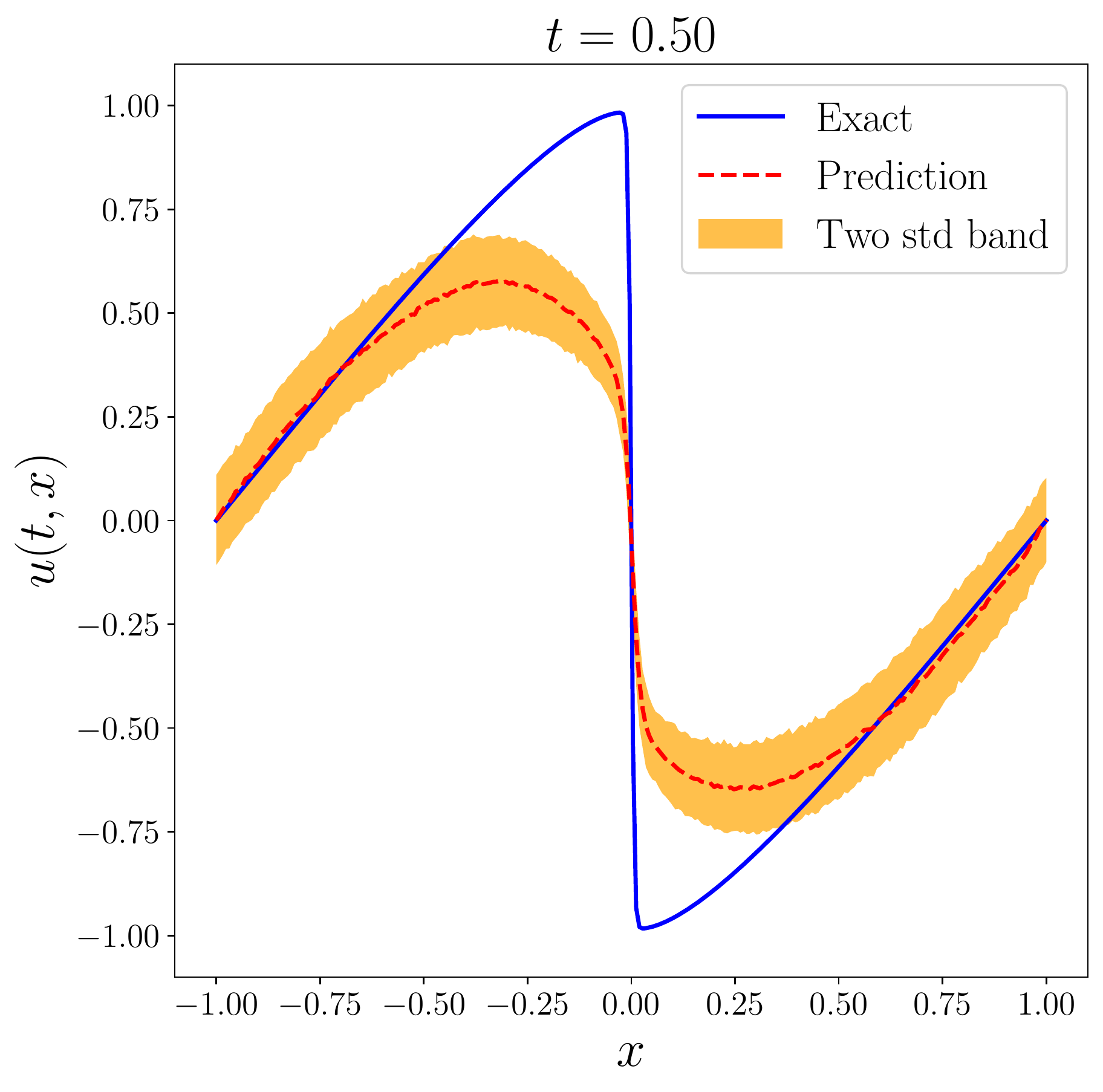}}
\vspace{-3ex}
\caption{Comparison of the predicted and exact solutions of Burgers' equation corresponding to $t=0.50$ snapshot.}
\label{fig:Burgers_predictions}
\end{figure*}

\begin{figure*}[ht]
\centering
\subfigure[PID-GAN]{
\label{fig:pid_schro} \includegraphics[scale=0.12]{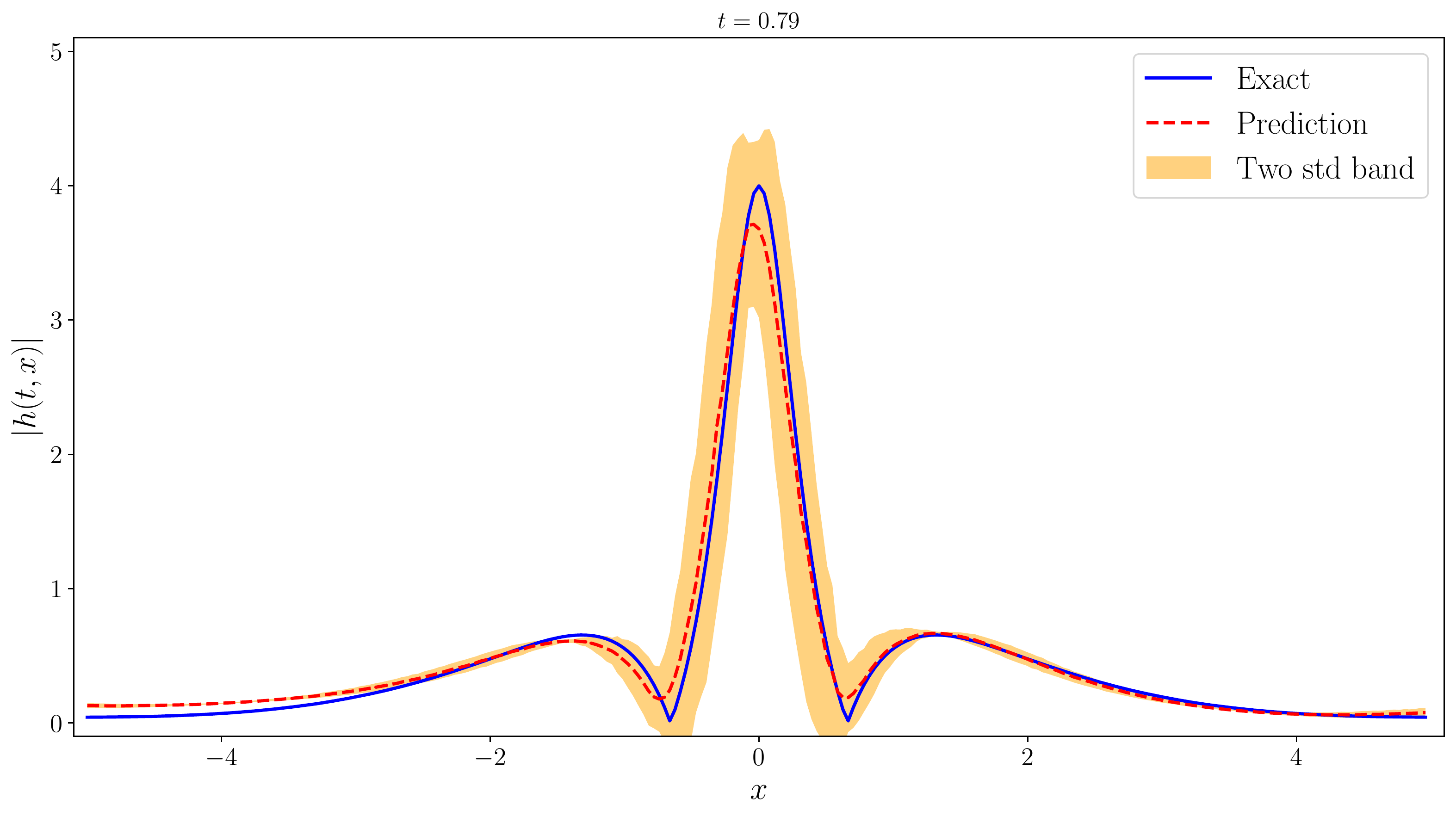}} 
\subfigure[PIG-GAN]{
\label{fig:pig_schro} \includegraphics[scale=0.12]{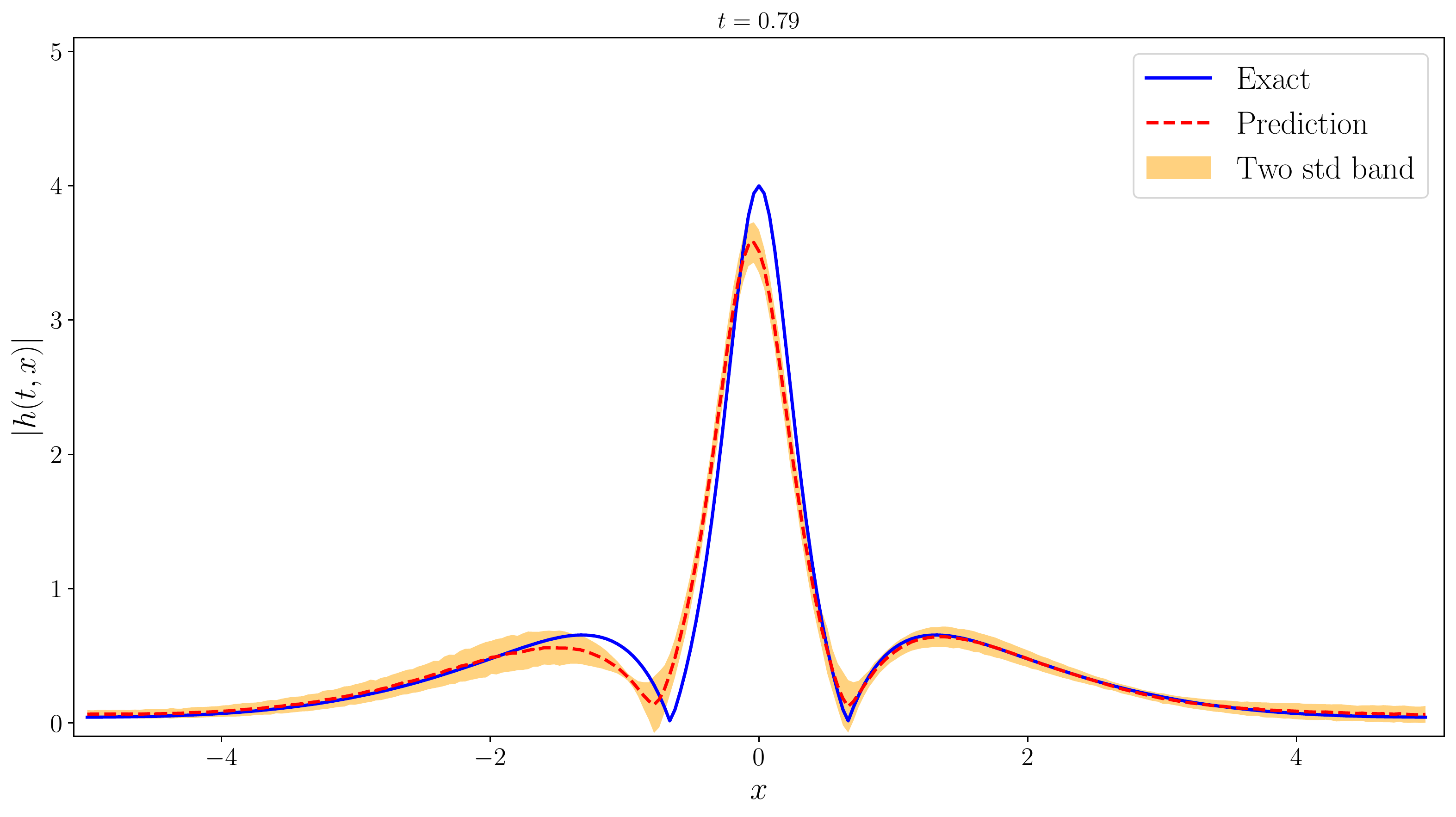}}
\subfigure[PINN-Drop]{
\label{fig:pinn_schro} \includegraphics[scale=0.12]{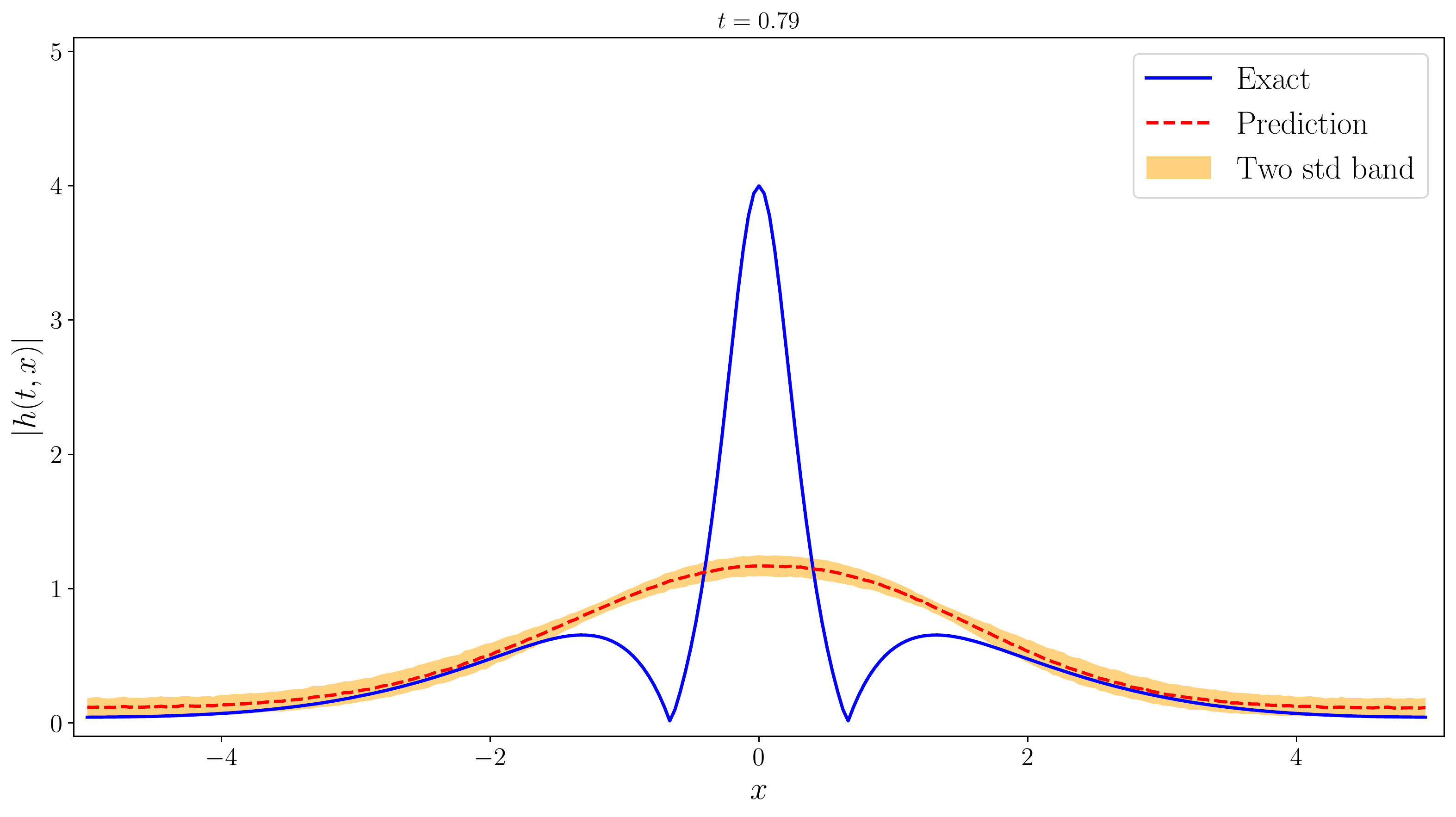}}
\subfigure[APINN-Drop]{
\label{fig:apinn_schro} \includegraphics[scale=0.12]{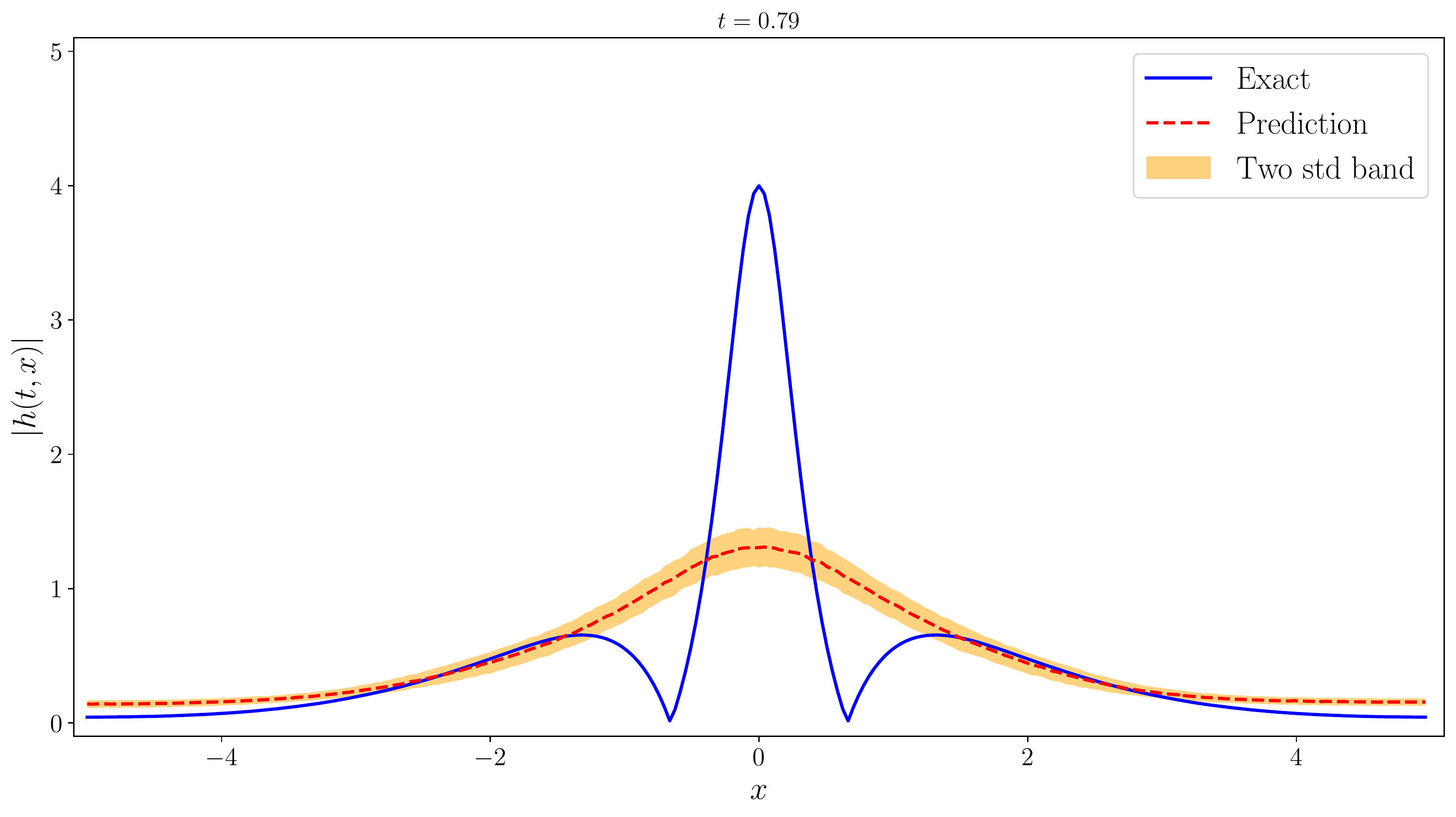}}
\vspace{-3ex}
\caption{Comparison of the predicted and exact solutions of Schr\"{o}dinger equation corresponding to $t=0.79$ snapshot.}
\label{fig:schro_predictions}
\end{figure*}

\section{Additional Analysis of Results}
\label{sec-additionalresult}
\subsection{Comparing PID-GAN and APINN-Drop on Darcy's Equation}
We visualize the performance of the PID-GAN and APINN-Drop for noisy conditions to gain more insights into each of these models. From Table 2, we observed that PID-GAN and APINN-Drop had similar relative $L^2$ error - u. However, the relative $L^2$ error - k of APINN-Drop is significantly worse than that of PID-GAN. This is also evident from the plots of Absolute error of $k$.  Also, the variances of the PID-GAN are much lower than those of the APINN-Drop. We observe the trend that PINN variants usually have much larger variances in their predictions, thus achieving higher values of 95\% C.I. It can be inferred that APINN-Drop is usually under-confident in its predictions.  However, this behavior might not be desirable. Ideally, we would want to have a higher value of 95\% C.I. with lower values of standard deviations. PID-GAN on the other hand, generates significantly lower errors in $k$ while having lower variances with a relatively high value of 95\% C.I.  
% \subsection*{Effect of Training Fraction}
% For collision dataset, we perform a comparative study to illustrate the effect of training fractions on non-ideal physics scenario. 

% We observe the PID-GAN outperforms the PIG-GAN on every training fractions. At a lower training fraction 

% As discussed in the experiment section, we observe PID-GAN outperforms all other baselines although the ground truth predictions do not follow the ideal physics. The conventional physics informed approaches such as PIG-GAN,  
% For synthetic dataset, we perform additional ...
% \begin{figure}
%     \centering
%     \includegraphics[scale=0.31]{supplementary_figures/training_frac.pdf}
%     \caption{Effect of training fractions for collision dataset}
%     \label{fig:my_label}
% \end{figure}

\subsection{Prediction comparison for Burgers' equation}
Figure \ref{fig:Burgers_predictions} shows the comparison of the predicted and the exact solutions of Burgers' equation for different baselines at t=0.5 snapshot. It is evident from the figure that at x=0, there is a steep slope, which is hard to predict by conventional neural networks. The original PINN paper, which does point estimates, shows better performance on finding the exact solution of Burgers' equation than our dropout based PINN-Drop method. This justifies our observation on MC-Dropout that its minor perturbation can easily throw-off a model to become physically inconsistent, which motivates us to estimate the uncertainty using GAN-based models.

\subsection{Prediction comparison for Schr\"{o}dinger equation}
Figure \ref{fig:schro_predictions} illustrates a similar observation for MC-Dropout based models. PINN-Drop and APINN-Drop perform well on the smooth response region, whereas, for the steep response region, the predictions and the uncertainty of these models are inconsistent. GAN-based model predictions are close to the exact solutions, and from the results, the PID-GAN performs much better than the PIG-GAN in terms of residual error and uncertainty estimate values.

\end{document}